%% file: main.tex
\documentclass[10pt,twocolumn,letterpaper]{article}

\usepackage[pagenumbers]{cvpr} % To force page numbers, e.g. for an arXiv version

\input{preamble}

\definecolor{cvprblue}{rgb}{0.21,0.49,0.74}
\usepackage[pagebackref,breaklinks,colorlinks,allcolors=cvprblue]{hyperref}

\def\confName{CVPR}
\def\confYear{2027}

\title{Semantic Detour Flow: Pixel Diffusion that Decides Semantics First}

\author{%
Quan Dao\textsuperscript{1}\quad
Tuan Pham\textsuperscript{2}\quad
Anh Nguyen\textsuperscript{3}\quad
Duc Vu\textsuperscript{1}
\\[0.25em]
Quyen Tran\textsuperscript{1}\quad
Stephan Mandt\textsuperscript{2}\quad
Dimitris Metaxas\textsuperscript{1}
\\[0.4em]
\normalsize\textsuperscript{1}Rutgers University\quad
\normalsize\textsuperscript{2}University of California, Irvine\quad
\normalsize\textsuperscript{3}Johns Hopkins University
\\[0.3em]
{\tt\small quan.dao@rutgers.edu}
}

\begin{document}
\maketitle
\input{sec/0_abstract}
\input{sec/1_intro}
\input{sec/2_related}
\input{sec/3_method}
\input{sec/4_guidance}
\input{sec/5_experiments}
\input{sec/6_extensions}
\input{sec/7_conclusion}
% References: enable once \cite commands replace the red [REF: ...] markers (BibTeX errors on a paper with no citations).
{
    \small
    \bibliographystyle{ieeenat_fullname}
    \bibliography{main}
}

% Appendix: compiled with the paper; split the PDF at submission if the venue wants a separate file.
\input{sec/X_suppl}

\end{document}

%% file: preamble.tex
\newcommand{\appendixitem}[1]{%
    \item \hyperref[#1]{Section~\ref*{#1}\hspace{1em}\nameref*{#1}}%
}
\usepackage{xspace}
\newcommand{\method}{Detour\xspace}
\usepackage{adjustbox}
\newcommand{\gcheck}{\textcolor{ForestGreen}{\checkmark}}

%% file: sec/0_abstract.tex
\begin{abstract}
Diffusion models trained in semantic latent spaces converge much faster than pixel-space models, but rely on decoders that limit fine details.
We attribute this gap to the denoising target at high noise: the posterior mean of pixels is blurry, whereas that of self-supervised features forms a semantic prototype that preserves content.
Motivated by this observation, we propose \emph{\method}, a pixel-space flow that follows frozen DINOv2 features at high noise before transitioning to pixels at low noise.
A single network trained with a unified velocity loss learns semantics near noise and fine details near data, while sampling follows a pixel-space ODE without an encoder or decoder.
On ImageNet, with the same network and 20 training epochs, \method reduces FID-10k from 92.8 to 38.8 compared to standard pixel flow, achieving FID-50k scores of 1.43 at $256{\times}256$ and 1.59 at $512{\times}512$ with longer training.
The path requires no class labels and extends naturally to text-to-image generation: with matched networks, \method achieves comparable text alignment with $1.5\times$ fewer training iterations.
A 3.3B model achieves GenEval / DPG-Bench scores of 0.88 / 87.4 at $1024{\times}1024$ and 0.90 / 87.8 at $512{\times}512$, outperforming the compared pixel-space models at both resolutions while preserving fine textures and rendered text that semantic-latent decoders may lose.
Code are available at \url{https://github.com/quandao10/detour-flow}.
\end{abstract}

%% file: sec/1_intro.tex
\section{Introduction}
\label{sec:intro}

\begin{figure*}[t]
  \centering
  \includegraphics[width=\linewidth]{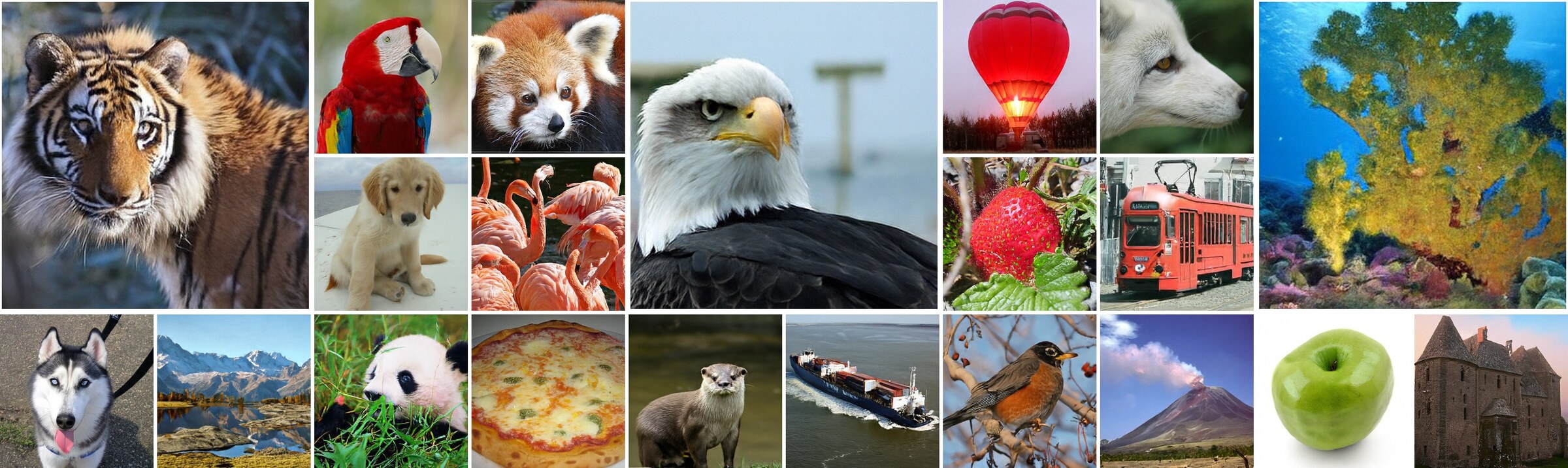}
  \caption{\textbf{\method samples} in pixel space: $512{\times}512$ (large) and $256{\times}256$ (small).}
  \label{fig:teaser}
\end{figure*}

Diffusion and flow models \cite{ho2020denoising, song2020score, lipman2022flow, liu2022flow} have become the dominant approach to image generation. Early methods operated directly in pixel space~\cite{ho2020denoising, dhariwal2021diffusion}. Latent diffusion~\cite{rombach2022high} enabled efficient high-resolution generation by operating in the compressed space of a pretrained autoencoder, while transformer backbones~\cite{peebles2023scalable, ma2024sit} further scaled this approach.

Latent diffusion, however, inherits the limitations of its autoencoder. The decoder limits reconstruction quality, often losing fine details such as textures, small text, and thin structures. The latent representation must also balance reconstruction quality and generative performance~\cite{yao2025reconstruction, leng2025repa}, while training requires two stages.
Pixel-space diffusion avoids these limitations and can be trained end to end~\cite{hoogeboom2023simple, hoogeboom2025simpler, li2026back, chen2025pixelflow, wang2026pixnerd, ma2026pixelgen}. Its main drawback is training cost: pixel-space models require substantially more training epochs or larger networks to match the FID of latent models.

Recent work shows that diffusion in \emph{semantic} latent spaces converges much faster. Aligning latents with self-supervised features~\cite{yu2024representation, yao2025reconstruction} or performing diffusion directly in DINOv2 feature space~\cite{zheng2026diffusion} achieves strong FID with substantially less training.
However, these methods still require a decoder to reconstruct pixels from semantic features, which do not fully preserve fine image details. On held-out text-to-image data, the RAE decoder~\cite{zheng2026diffusion} achieves only 18\,dB PSNR, losing small text, facial details, and fine textures. Its text-to-image extension~\cite{singh2026improved} also relies on a feature decoder.
This raises a natural question: \textit{Can a pixel-space model converge \textbf{as fast as} a model trained in semantic latent space?}

\paragraph{Why semantic latents converge fast.}
We argue that the key lies in what the denoiser predicts at high noise. Under a squared loss, the optimal denoiser predicts the posterior mean. At high noise, many different images are consistent with the noisy input, so their posterior mean in pixel space becomes a blur that carries little information about the object. In DINOv2 space, however, the posterior mean forms a semantic prototype that retains class information.
At low noise, the noisy input strongly constrains the possible clean images, making the pixel-space posterior mean sharper and more informative. Pixels therefore become a better prediction target, as they preserve the fine details absent from semantic features.

\paragraph{Our approach.}
We propose \method (Semantic Detour Flow), a pixel-space flow whose noisy state follows the frozen DINOv2 features of the image at high noise and transitions to the image at low noise (\cref{fig:path}).
A fixed pixel shuffle places the features in pixel space, and a standard pixel transformer predicts both the image and the features under a single velocity loss.
The encoder is used only to construct training targets, so sampling is a standard pixel-space ODE from noise to RGB, with neither an encoder nor a decoder.
With the same network and training budget, \method reduces FID-10k from 92.8 to 38.8 relative to plain pixel flow, and reaches FID-50k of 1.43 on ImageNet $256{\times}256$ and 1.59 on $512{\times}512$ within 200 epochs (\cref{fig:teaser}).
The path requires no class labels and extends directly to text conditioning. With a matched network, \method matches the prompt-following performance of pixel flow with $1.5\times$ fewer iterations, with the largest gains on counting, position, and attribute binding.
A 3.3B model trained in stages of increasing resolution across multiple aspect ratios scores above the pixel-space models we compare with on both GenEval and DPG-Bench (\cref{tab:t2i}). Because Detour generates pixels directly, it preserves the fine texture and rendered text that a feature decoder loses (\cref{fig:t2i_gallery}).

% \paragraph{What makes the path work.}
% Two properties decide how well the hand-over trains: it should proceed at a steady rate, and the
% signal should keep the same norm at every $t$, so the noise level alone sets the signal-to-noise
% ratio. A trigonometric schedule has both. Guidance differs too: the code settles the class early, so
% classifier-free guidance saturates near noise while AutoGuidance works along the whole path, and the
% two compose.

\paragraph{Contributions.}
\begin{itemize}[leftmargin=*,itemsep=1pt,topsep=2pt]
  \item We explain the fast convergence of semantic latent diffusion through the high-noise posterior mean, supported by a linear probe analysis.
  \item We propose \method, a pixel-space flow that captures semantics at high noise and fine details at low noise, trained with a single velocity loss and sampled without a decoder. We further identify the key design choices behind its effectiveness.
  \item On ImageNet, \method achieves FID-50k of \textbf{1.43} ($256{\times}256$) and \textbf{1.59} ($512{\times}512$) within 200 epochs, without an autoencoder.
  \item For text-to-image generation, \method learns compositional alignment $1.5\times$ faster than pixel flow with the same network. A 3.3B model \textbf{outperforms the compared pixel-space models} on both prompt-following benchmarks while preserving pixel-level details.
\end{itemize}

%% file: sec/2_related.tex
\section{Related Work}
\label{sec:related}

\paragraph{Pixel-space diffusion.}
Although pixel-space diffusion avoids the reconstruction limits of autoencoders~\cite{dhariwal2021diffusion, hoogeboom2023simple, hoogeboom2025simpler, kingma2023understanding, jabri2022scalable}, they require more computational budget and often lags behind latent-space diffusions. Recent work~\cite{li2026back} narrows the gap with latent-based models through clean-image prediction, patch decoding with neural fields or frequency-specific heads~\cite{wang2026pixnerd, ma2026deco}, multi-resolution generation~\cite{chen2025pixelflow}, and self-supervised pretraining~\cite{lei2025advancing}. Other pixel transformers~\cite{ma2026pixelgen, yu2026pixeldit, chen2026dip} further improve architectures and training strategies. In contrast, \method retains a standard pixel transformer and instead changes \emph{what the noisy state contains}.

\paragraph{Semantic latents and representation alignment.}
Latent diffusion~\cite{rombach2022high, peebles2023scalable, ma2024sit} converges faster when latents are aligned with self-supervised features~\cite{yu2024representation, yao2025reconstruction, leng2025repa} or directly use the encoder's feature space~\cite{zheng2026diffusion}. However, these methods still require a learned decoder to reconstruct pixels.
In pixel space, representation alignment is more challenging: naively applying REPA to JiT-B worsens FID at 200 epochs (4.37 to 5.14), while PixelREPA~\cite{shin2026representation} requires an adapter and token masking to achieve improvements. Other pixel-space methods predict DINOv2 features as auxiliary outputs~\cite{ghosh2026dino, baade2026latent}.
In contrast, \method incorporates DINOv2 features directly into the \emph{diffusion path}, requiring neither an alignment loss nor a decoder, with the encoder used only during training.

\paragraph{Diffusion paths.}
Flow matching and stochastic interpolants~\cite{lipman2022flow, liu2022flow, albergo2025stochastic} allow flexible paths between noise and data. Prior work explores alternative noise schedules~\cite{karras2022elucidating, esser2024scaling, nichol2021improved}, introduces blurring or other corruptions~\cite{hoogeboom2022blurring, bansal2023cold, rissanen2022generative}, or generates progressively across resolutions or modalities~\cite{ho2022cascaded, gu2022f, chen2025pixelflow}.
In contrast, \method defines a path on clean signal that interpolates between semantic features and RGB images in pixel coordinates.

% \paragraph{Guidance.}
% Classifier-free guidance (CFG) \cite{ho2022classifier} trades diversity for fidelity, and guidance intervals \cite{kynkaanniemi2024applying} and APG \cite{sadat2024eliminating} reduce its artifacts. AutoGuidance \cite{karras2024guiding} instead contrasts the model with a weaker version of itself; we use the early checkpoint of the same model as the weaker version.

%% file: sec/3_method.tex
\section{Semantic Detour Flow}
\label{sec:method}

\subsection{Motivation: what denoiser learns at high noise}
\label{sec:prelim}
\begin{figure*}[t]
  \centering
  \begin{minipage}[b]{0.395\textwidth}
    \centering
    \includegraphics[width=\linewidth]{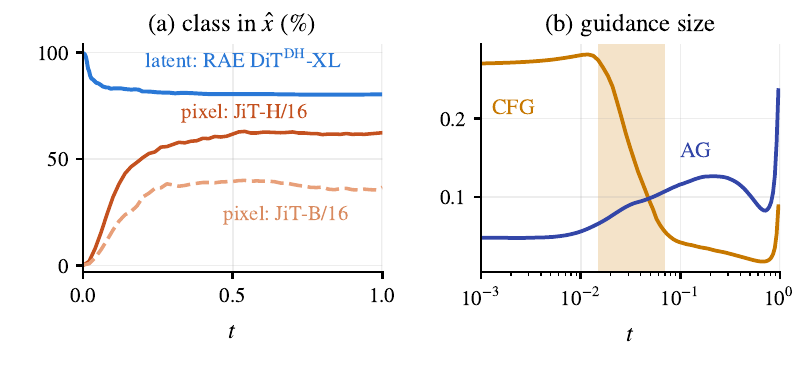}
    \caption{\textbf{Where the semantics are decided.} (a)~the class is readable in the latent
    prediction at once, in the pixel one only after $t\approx0.1$; (b)~CFG acts before
    $t\approx0.05$, AutoGuidance (AG) along the whole path.}
    \label{fig:probe}
    \label{fig:directions}
  \end{minipage}\hfill
  \begin{minipage}[b]{0.585\textwidth}
    \centering
    \includegraphics[width=\linewidth]{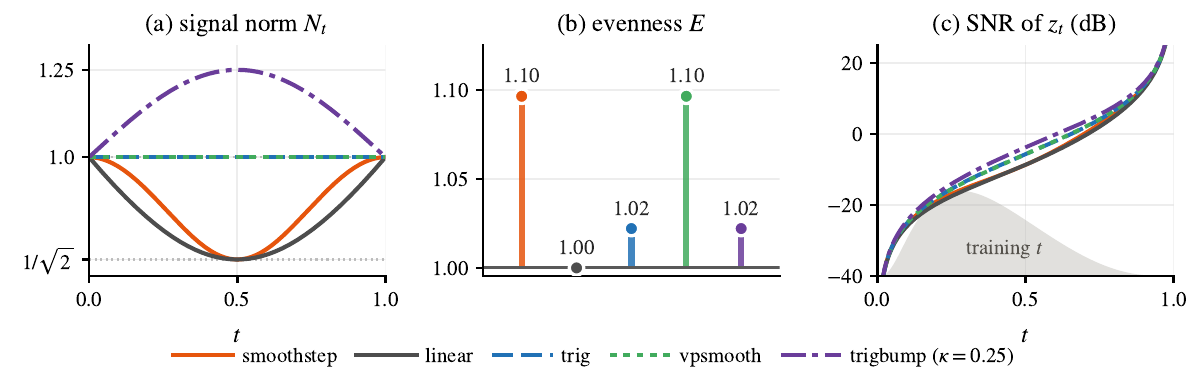}
    \caption{\textbf{Hand-over schedules}, convex (solid) against norm-preserving (dashed). The signal
    norm sags to $1/\sqrt2$ when convex, holds $1$ when trigonometric, and is raised to $1.25$ by the
    bump. Coinciding curves are dashed on top of each other.}
    \label{fig:supp_schedules}
  \end{minipage}
\end{figure*}

In flow-matching, where $t\in[0,1]$ interpolates from noise ($t=0$) to data ($t=1$), the noisy state is $z_t = t\,x + (1-t)\,\varepsilon$, with $\varepsilon\sim\mathcal N(0,I)$, and a network $\hat x_\theta(z_t,t,y)$ predicts the clean image where y is class condition. Under a squared loss, the optimal prediction is the posterior mean, $\hat x_\theta = \mathbb E[x\mid z_t,y]$.
As $t\to0$, the noisy state carries little information about the image, and the posterior mean approaches the \emph{class-mean image}, which appears blurry. Yet this is where the model begins to determine the sample's content, including object identity, location, and pose.
For a self-supervised feature map $c(x)$, the corresponding limit is $\mathbb E[c(x)\mid y]$, a \emph{class prototype}. Since classes are nearly linearly separable in DINOv2 space, this prototype retains meaningful semantic information.
\Cref{sec:supp_probe} provides the probe protocol and per-step accuracies for \cref{fig:probe}.

\paragraph{Probe test.}
We measure how much class information the clean prediction $\hat x$ carries during sampling using a linear probe on DINOv2 features (\cref{fig:probe}). Pixel-model predictions are first encoded by DINOv2, while latent-model predictions already lie in this feature space, allowing both to be evaluated with the same probe. We additionally test class-mean images using a pretrained ResNet-50 classifier (\cref{sec:supp_probe}).
At the first step from pure noise, the prediction of a latent model trained on DINOv2-B features (RAE DiT$^{\rm DH}$-XL) is correctly classified for 99.8\% of samples, compared to only 0.1\% for the pixel model JiT-H/16, near chance level.
The class means show a similar gap: the mean DINOv2 feature is correctly classified for all 1{,}000 classes, whereas ResNet-50 recognizes the mean pixel image for only 1.0\% of classes.
This suggests that pixel-model predictions carry little class information at high noise and must recover it later from a weak signal.
We attribute the faster convergence of semantic-latent diffusion~\cite{zheng2026diffusion} to this difference. With the same 296M network and 20 training epochs, JiT achieves FID-10k of 12.8 in DINOv2-B space, compared to 92.8 in pixel space (\cref{tab:recipe}).
At low noise, the noisy state strongly constrains the clean image, producing a sharper pixel-space prediction that preserves fine details without a decoder.
\method hence learns semantics from features at high noise and fine details from pixels at low noise.

\begin{figure*}[t]
  \centering
  \includegraphics[width=0.9\linewidth]{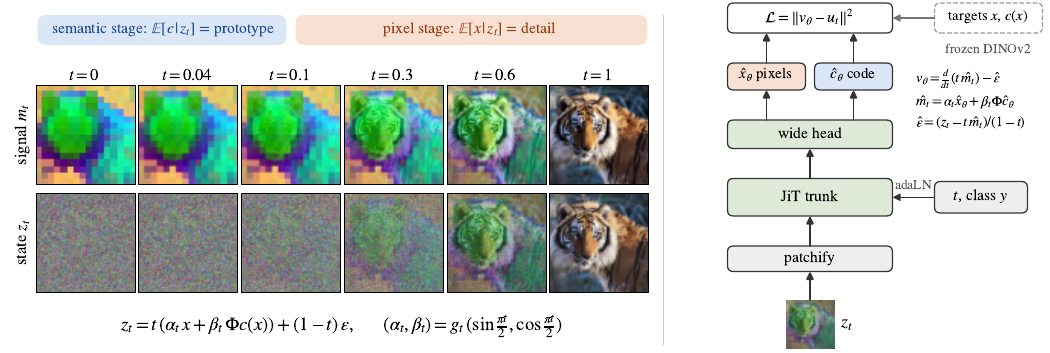}
  \caption{\textbf{\method.} Left: signal $m_t$ and state $z_t$ on a real image (code shown by PCA colours). Right: network predicts image \& code.}
  \label{fig:path}
\end{figure*}

\subsection{The detour path}
\label{sec:path}
Let $c(x)\in\mathbb R^{h\times w\times 768}$ denote the frozen DINOv2-B features of $x$ on the $16{\times}16$ patch grid, centred and scaled to a standard deviation of $A{=}0.5$, the scale of the image itself, so that neither term of the mix dominates the other. We call $c(x)$ the \emph{code}. Let $\Phi$ be the pixel shuffle that maps each 768-dimensional token into the $3{\times}16{\times}16$ pixels of its corresponging patch, so that $\Phi c(x)$ is an image-shaped tensor aligned with $x$.
The noisy state mixes the image and the code and then adds noise:
\begin{equation}
  z_t = t\,m_t + (1-t)\,\varepsilon,\qquad m_t = \alpha_t\,x + \beta_t\,\Phi c(x).
  \label{eq:path}
\end{equation}
The mixing weights hand the signal over from the code ($\alpha_0{=}0$, $\beta_0{=}1$) to the image ($\alpha_1{=}1$, $\beta_1{=}0$). We use a trigonometric hand-over with a mid-path norm bump $g_t$,
\begin{equation}
  (\alpha_t,\beta_t) = g_t\,(\sin\theta_t,\cos\theta_t),\quad g_t = 1+\kappa\sin2\theta_t,
  \label{eq:trig}
\end{equation}
with $\theta_t=\pi t/2$ and $\kappa{=}0.25$; \cref{sec:design} explains design choice of detour path.
Differentiating \cref{eq:path} gives the target velocity
\begin{equation}
  u_t = (\alpha_t+t\alpha_t')\,x + (\beta_t+t\beta_t')\,\Phi c(x) - \varepsilon .
  \label{eq:vel}
\end{equation}
Plain pixel flow is the special case $\alpha\equiv1$, $\beta\equiv0$.
\Cref{fig:path} (left) shows the path on a real image with its code. Near noise, the state contains mainly the code, whose posterior mean is a class prototype; near data, it contains the image, whose posterior mean carries the detail. For display, the code is shown by its first three principal components, whereas in the model $\Phi$ places all 768 channels in each patch.
DINOv2-B's 768 channels are exactly one $16{\times}16{\times}3$ patch, so $\Phi$ is an isometry onto the image itself. \Cref{sec:supp_codedim} breaks that coincidence: it compares encoders (DINOv2-L, MAE, LeJEPA, CLIP) and code widths $C\neq 3p^2$, which must be projected down or carried in a padded state.

\subsection{Training objective}
\label{sec:objective}
\label{sec:loss}
The network (\cref{fig:path}, right) predicts both endpoints of the hand-over for each patch: the image $\hat x_\theta$ and the semantic code $\hat c_\theta$. These predictions are combined to obtain the predicted signal $\hat m_t$ and the noise estimate $\hat\varepsilon$ implied by $z_t$:
\begin{equation}
  \hat m_t = \alpha_t\,\hat x_\theta + \beta_t\,\Phi\hat c_\theta,\qquad \hat\varepsilon = \frac{z_t - t\,\hat m_t}{1-t},
  \label{eq:param}
\end{equation}
and the predicted velocity follows \cref{eq:vel} with predictions in place of targets:
\begin{equation}
  v_\theta(z_t,t,y) = \tfrac{d}{dt}\bigl(t\,\hat m_t\bigr) - \hat\varepsilon .
  \label{eq:vtheta}
\end{equation}
Our velocity objective follows:
\begin{equation}
  \mathcal L(\theta) = \mathbb E_{x,y,\varepsilon,t}\,\bigl\| v_\theta(z_t,t,\tilde y) - u_t \bigr\|^2 ,
  \label{eq:loss}
\end{equation}
where $\tilde y$ is the class label, replaced by a null label with probability 0.1 to enable guidance. Time $t$ is drawn from a logit-normal distribution $(-0.63,0.85)$ and, with probability 0.1, from $\mathcal U[0,0.1]$. That second draw is \emph{interval sampling}: the short interval near noise where the class is settled (\cref{sec:window}) is otherwise barely trained, since the logit-normal draw alone places only 0.28\% of the training times in it and the FID lost there does not fall with further training (\cref{sec:supp_stage}).
No alignment loss or other auxiliary objective is used.

\paragraph{What the loss learns.}
\Cref{eq:vtheta} is linear in the two predictions, $v_\theta = a_t\,\hat x_\theta + b_t\,\Phi\hat c_\theta - z_t/(1-t)$, with $a_t = \alpha_t/(1-t)+t\alpha_t'$ and $b_t=\beta_t/(1-t)+t\beta_t'$. The target satisfies the same identity,
\begin{equation}
  u_t = a_t\,x + b_t\,\Phi c(x) - \frac{z_t}{1-t},
  \label{eq:identity}
\end{equation}
so \cref{eq:loss} equals $\mathbb E\,\|a_t(\hat x_\theta-x) + b_t\,\Phi(\hat c_\theta-c)\|^2$, and its minimiser predicts the posterior means $\mathbb E[x|z_t]$ and $\mathbb E[c|z_t]$.
Near noise, $a_t\to0$ and $b_t\to1$: the loss is the error on the code, and the model learns semantic prototypes as a latent model would. Near data, $a_t\approx 1/(1-t)$ and $b_t\to0$: the loss becomes the JiT pixel loss, and the model learns detail.
A single network and a single loss thus cover both regimes, without switching between stages. In appendix, \Cref{sec:supp_theory} derives the velocity, the loss and the closed-form score on this path, and
\cref{sec:supp_code} is ablation about the architecture head.

\begin{figure*}[t]
  \centering
  \includegraphics[width=0.85\linewidth]{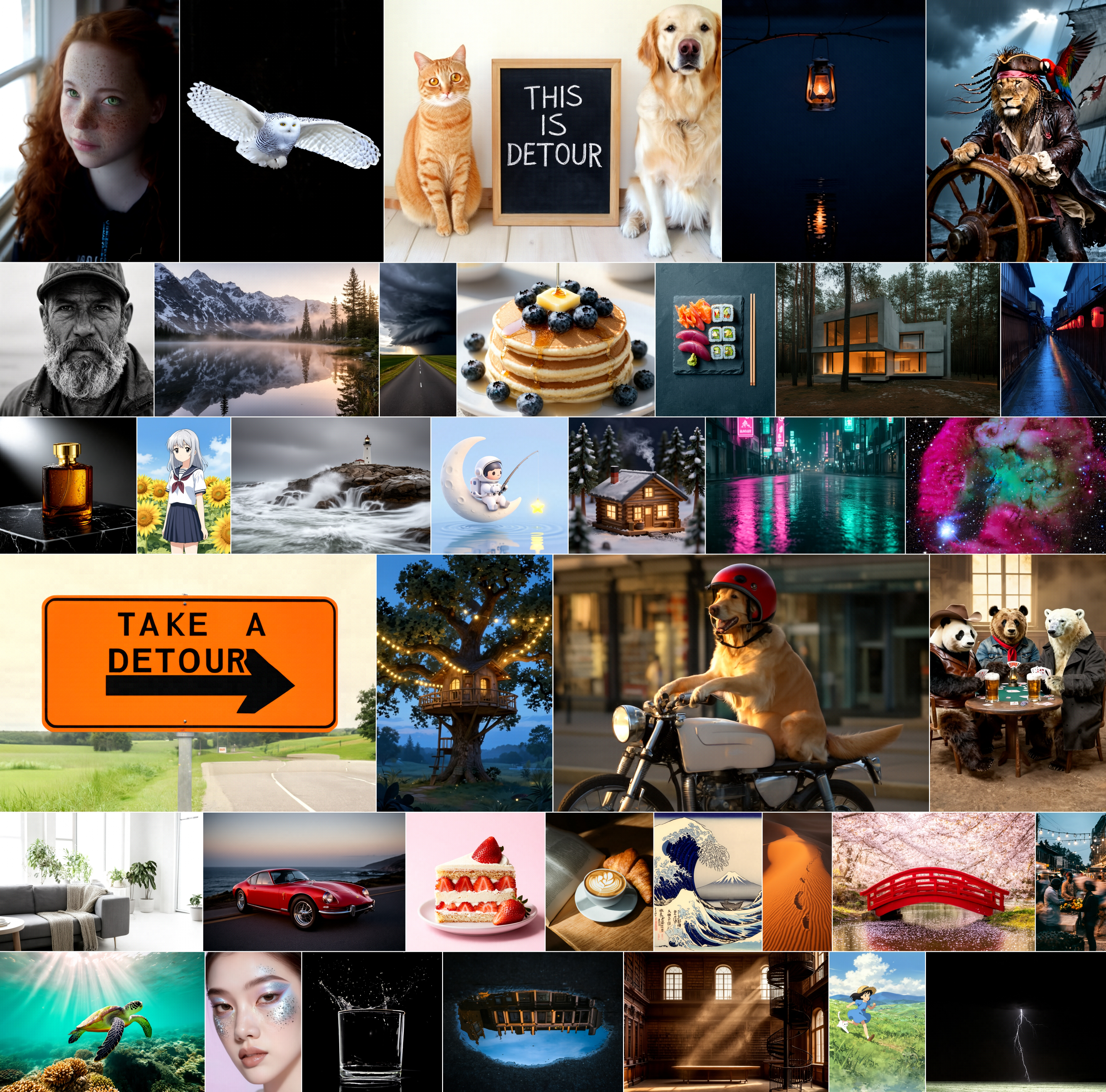}
  \caption{\textbf{Text-to-image samples} from the 3.3B \method model at about $1024{\times}1024$ pixels and several aspect ratios.}
  \label{fig:t2i_gallery}
\end{figure*}

\subsection{Design of the detour}
\label{sec:design}

\paragraph{Hand-over schedule.}
The mixing weights $(\alpha_t,\beta_t)$ control the transition from semantic features to pixels. We consider two desirable properties: a steady hand-over and preservation of signal magnitude (\cref{fig:supp_schedules}).

\textbf{Evenness.}
We measure the smoothness of the hand-over using the relative pixel weight $p_t=\alpha_t/(\alpha_t+\beta_t)$, which increases from $0$ to $1$. Since its average rate of change is $1$, the RMS rate $E$ satisfies $E^2=1+\operatorname{Var}(\mathrm dp_t/\mathrm dt)\ge1$. The minimum $E=1$ corresponds to a constant rate, ensuring a steady transition from features to pixels (\cref{fig:supp_schedules}b).

\textbf{Signal norm preservation.}
In $z_t=t\,m_t+(1-t)\varepsilon$, changes in $\|m_t\|$ alter the signal-to-noise ratio beyond what is controlled by $t$. Since $x$ and $\Phi c(x)$ are nearly uncorrelated and have similar scales, the relative signal magnitude is approximated by $N_t=\sqrt{\alpha_t^2+\beta_t^2}$. A trigonometric schedule maintains $N_t=1$, whereas a convex mixture ($\alpha_t+\beta_t=1$) reduces it to $1/\sqrt2$ at the midpoint, weakening the signal during the transition (\cref{fig:supp_schedules}a,c).

\begin{table}[t]
\centering
\scriptsize
\setlength{\tabcolsep}{5pt}
\begin{tabular}{lccc}
\toprule
Schedule & $E$ & $N_{0.5}$ & FID-10k $\downarrow$ \\
\midrule
Smoothstep & 1.10 & 0.71 & 66.6 \\
Linear & 1.00 & 0.71 & 60.7 \\
\textsc{vpsmooth} & 1.10 & 1.00 & 58.1 \\
Trigonometric & 1.02 & 1.00 & 52.7 \\
Trig + bump & 1.02 & 1.25 & \textbf{51.0} \\
\bottomrule
\end{tabular}
\caption{Hand-over schedules (JiT-B, 20 epochs, unguided FID-10k). $E$: evenness; $N_{0.5}$: midpoint signal magnitude.}
\label{tab:handover}
\vspace{-6mm}
\end{table}

\Cref{tab:handover} shows that norm preservation improves FID by about 8 points, while a more even hand-over gains another 5--6 points. These improvements are approximately additive, making the trigonometric schedule strongest baseline.

\textbf{Mid-path norm bump.}
We further increase the signal magnitude near $t=0.5$, where pixels and features contribute equally and the transition may be hardest to learn. The bump in \cref{eq:trig} scales both weights equally, preserving the hand-over rate while raising $N_t$ to $1+\kappa\sin2\theta_t$ (\cref{fig:supp_schedules}a). With $\kappa=0.25$, the midpoint magnitude reaches $1.25$ without affecting either endpoint. See appendix \cref{sec:supp_sched} for further ablation.

\paragraph{Multi-layer code.}
\label{sec:k4}
The final DINOv2 layer emphasizes high-level semantics, while intermediate layers preserve more spatial layout and object-part information. Following~\cite{singh2026improved}, we average the layer-normalized tokens from blocks 6, 8, 10, and 12, then add the spatial mean of the final-layer tokens. This retains 768 channels without additional encoder passes.

\paragraph{Architecture.}
We adopt JiT~\cite{li2026back} without its patch bottleneck, using a wide two-block output head of width 2048~\cite{zheng2026diffusion}. The head predicts 768 pixel values and 768 semantic code values per $16{\times}16$ patch. JiT-B and JiT-XL have 296M and 849M parameters, respectively. At $512{\times}512$, we increase the patch size to $32{\times}32$ (JiT-XL/32) to maintain the same token count as at $256{\times}256$, and bilinearly upsample the $16{\times}16$ code before applying $\Phi$.

%% file: sec/4_guidance.tex
\subsection{Sampling}
\label{sec:guidance}

\begin{table*}[t]
\centering
\caption{\textbf{Class-conditional ImageNet} (FID-50k). Gray: latent space. $^\dagger$: frozen pretrained encoder during training.}
\label{tab:system}
\small
\setlength{\tabcolsep}{3pt}
\begin{adjustbox}{max width=0.495\textwidth}
\begin{tabular}{@{}lrrcccc@{}}
\toprule
\multicolumn{7}{c}{ImageNet $256{\times}256$} \\
Method & Params & Ep. & FID & IS & Prec. & Rec. \\
\midrule
\color{gray}DiT-XL/2 \cite{peebles2023scalable} & \color{gray}675M & \color{gray}1400 & \color{gray}2.27 & \color{gray}278 & \color{gray}0.83 & \color{gray}0.57 \\
\color{gray}REPA$^\dagger$\cite{yu2024representation} & \color{gray}675M & \color{gray}800 & \color{gray}1.42 & \color{gray}306 & \color{gray}0.80 & \color{gray}0.65 \\
\color{gray}LightningDiT \cite{yao2025reconstruction} & \color{gray}675M & \color{gray}800 & \color{gray}1.35 & \color{gray}295 & \color{gray}0.79 & \color{gray}0.65 \\
\color{gray}RAE$^\dagger$\cite{zheng2026diffusion} & \color{gray}839M & \color{gray}800 & \color{gray}1.13 & \color{gray}263 & \color{gray}0.78 & \color{gray}0.67 \\
\midrule
ADM-G \cite{dhariwal2021diffusion} & 554M & 396 & 4.59 & 187 & 0.82 & 0.52 \\
SiD2 \cite{hoogeboom2025simpler} & -- & 1280 & 1.38 & -- & -- & -- \\
PixNerd$^\dagger$ \cite{wang2026pixnerd} & 700M & 320 & 1.93 & 298 & 0.80 & 0.60 \\
DeCo$^\dagger$\cite{ma2026deco} & 682M & 800 & 1.62 & 301 & 0.80 & 0.62 \\
PixelDiT\cite{yu2026pixeldit} & 797M & 320 & 1.61 & 293 & 0.78 & 0.64 \\
JiT-G/16\cite{li2026back} & 2B & 600 & 1.82 & 293 & 0.79 & 0.62 \\
PixelREPA-H$^\dagger$\cite{shin2026representation} & 953M & 600 & 1.81 & 317 & -- & -- \\
PixelDiT2-H$^\dagger$\cite{yu2026pixeldit} & 1.0B & 600 & 1.46 & 302 & -- & -- \\
\textbf{\method-XL/16}$^\dagger$ & 849M & 320 & \textbf{1.35} & 267 & 0.76 & 0.67 \\
\midrule
\multicolumn{7}{@{}l}{\emph{$\le$200 epochs}} \\
JiT-XL/16\cite{li2026back} & 676M & 200 & 2.89 & 272 & 0.85 & 0.66 \\
JiT-G/16\cite{li2026back} & 2B & 200 & 2.15 & -- & -- & -- \\
LatentForcing-L$^\dagger$\cite{baade2026latent} & 465M & 200 & 2.48 & -- & -- & -- \\
% Dino Forcing-L$^\dagger$\cite{ghosh2026dino} & 468M & 200 & 2.26 & 310 & 0.84 & 0.70 \\
Dino Forcing-XL$^\dagger$\cite{ghosh2026dino} & 686M & 200 & 2.13 & 310 & 0.83 & 0.71 \\
PixelGen\cite{ma2026pixelgen} & 676M & 160 & 1.83 & 294 & 0.79 & 0.63 \\
\textbf{\method-XL/16}$^\dagger$ & 849M & 200 & \textbf{1.43} & 267 & 0.77 & 0.65 \\
\bottomrule
\end{tabular}
\end{adjustbox}
\hfill
\begin{adjustbox}{max width=0.495\textwidth}
\begin{tabular}{@{}lrrcccc@{}}
\toprule
\multicolumn{7}{c}{ImageNet $512{\times}512$} \\
Method & Params & Ep. & FID & IS & Prec. & Rec. \\
\midrule
\color{gray}DiT-XL/2 \cite{peebles2023scalable} & \color{gray}675M & \color{gray}600 & \color{gray}3.04 & \color{gray}241 & \color{gray}0.84 & \color{gray}0.54 \\
\color{gray}REPA$^\dagger$\cite{yu2024representation} & \color{gray}675M & \color{gray}200 & \color{gray}2.08 & \color{gray}275 & \color{gray}0.83 & \color{gray}0.58 \\
\color{gray}EDM2-XXL\cite{karras2024analyzing} & \color{gray}1.5B & \color{gray}733 & \color{gray}1.25 & \color{gray}-- & \color{gray}-- & \color{gray}-- \\
\color{gray}RAE$^\dagger$\cite{zheng2026diffusion} & \color{gray}839M & \color{gray}400 & \color{gray}1.13 & \color{gray}260 & \color{gray}0.80 & \color{gray}0.63 \\
\midrule
ADM-G\cite{dhariwal2021diffusion} & 559M & 388 & 7.72 & 173 & 0.87 & 0.42 \\
SiD2\cite{hoogeboom2025simpler} & -- & 1280 & 1.48 & -- & -- & -- \\
PixNerd$^\dagger$\cite{wang2026pixnerd} & 700M & 320+ & 2.45 & 276 & 0.79 & 0.63 \\
EPG-L/32\cite{lei2025advancing} & 540M & 800 & 2.35 & 295 & -- & -- \\
DiP-XL/32\cite{chen2026dip} & 631M & -- & 2.31 & 292 & -- & -- \\
DeCo$^\dagger$\cite{ma2026deco} & 682M & 340 & 2.22 & 290 & 0.80 & 0.60 \\
PixelDiT\cite{yu2026pixeldit} & 797M & 850 & 1.81 & 279 & 0.78 & 0.67 \\
JiT-G/32\cite{li2026back} & 2B & 600 & 1.78 & 307 & -- & -- \\
WaiT-G/32\cite{pavasovic2026wait} & 2B & 600 & 1.67 & -- & -- & -- \\
% PixelMoE-H\cite{dong2026pixelmoe} & 3.7B & 320 & 1.53 & 322 & -- & -- \\
PixelDiT2-H/16$^\dagger$\cite{yu2026pixeldit2} & 1.1B & 680 & 1.48 & 296 & -- & -- \\
WaiT-G/16\cite{pavasovic2026wait} & 2B & 600 & \textbf{1.30} & -- & -- & -- \\
\textbf{\method-XL/32}$^\dagger$ & 849M & 280 & 1.35 & 297 & 0.79 & 0.68 \\
\midrule
\multicolumn{7}{@{}l}{\emph{$\le$200 epochs}} \\
JiT-H/32\cite{li2026back} & 956M & 200 & 2.51 & -- & -- & -- \\
JiT-G/32\cite{li2026back} & 2B & 200 & 2.11 & -- & -- & -- \\
\textbf{\method-XL/32}$^\dagger$ & 849M & 160 & \textbf{1.59} & 286 & 0.79 & 0.65 \\
\bottomrule
\end{tabular}
\end{adjustbox}
\end{table*}
\paragraph{The sampler and the class window.}
\label{sec:window}
We sample by integrating $\dot z = v_\theta(z,t)$ from $z_0=\varepsilon$ to $z_1$, which is returned as the RGB image. Sampling  requires neither an encoder nor a decoder.
Class identity is largely determined near noise, before $t\approx0.07$, when conditional and unconditional code predictions converge (\cref{tab:supp_window}). We therefore use Euler steps on the grid $t_i=(i/N)^p$ with $p{=}4$, concentrating steps near noise (\cref{sec:supp_solver}).
This early class determination also affects classifier-free guidance (CFG)~\cite{ho2022classifier}, which uses $v_u+s(v_c-v_u)$, where $v_c$ and $v_u$ denote conditional and null-label velocities.

\paragraph{CFG vanishes once the class is decided.}
By \cref{eq:identity}, the CFG direction $v_c-v_u$ depends on differences between conditional and unconditional posterior means, $\Delta f=\mathbb E[f\mid z_t,y]-\mathbb E[f\mid z_t]$ for $f\in\{x,c\}$. These differences are bounded in proportion to the remaining class uncertainty $1-p(y\mid z_t)$ (\cref{sec:supp_guidance}).
Near noise, $\Delta c\approx\mathbb E[c\mid y]-\mathbb E[c]$, so CFG pushes samples toward their class prototypes, reducing diversity. Once the code in $z_t$ reveals the class, $p(y\mid z_t)\to1$, and the CFG direction vanishes regardless of model quality.
\Cref{fig:directions} confirms this: $\|v_c-v_u\|$ drops by $5$--$10\times$ after $t\approx0.05$, with similar magnitudes at epochs 100, 160, and 200 (\cref{sec:supp_guidance}). Further training thus provides little additional CFG signal. Nevertheless, applying CFG only within $t\in[0,0.05)$ already corrects $60\%$ of the samples that unguided sampling gets wrong (\cref{tab:supp_window}).

\paragraph{AutoGuidance complements CFG.}
AutoGuidance~\cite{karras2024guiding} contrasts the model with a weaker version of itself through $v_c-v_g$. Since both use the same class label, this direction captures model-quality differences rather than class information and remains active after the class is determined.
We use the epoch-20 checkpoint of the same model as $v_g$, requiring no additional training. The AutoGuidance direction remains active throughout the path and is nearly orthogonal to CFG ($\cos\approx0.2$), allowing them to be combined:
\begin{equation}
  v = v_u + s\,(v_c-v_u) + (a-1)\,(v_c-v_g), t\in[0.05,0.9],
  \label{eq:combined}
\end{equation}
with $v=v_c$ outside this interval, following~\cite{kynkaanniemi2024applying}. CFG guides class selection near noise, while AutoGuidance improves sample quality along the remaining trajectory.

%% file: sec/5_experiments.tex
\section{Experiments}
\label{sec:exp}

\begin{table}[t]
\centering
\caption{\textbf{From pixel flow to \method} (JiT-B, 296M, unguided FID, Heun 50 uniform sampling). $\dagger$: with warmup lr}
\label{tab:recipe}
\small
\setlength{\tabcolsep}{4pt}
\begin{adjustbox}{max width=\linewidth}
\begin{tabular}{@{}lccc@{}}
\toprule
 & \multicolumn{2}{c}{20 ep., FID-10k} & 80 ep. \\
Model & FID & $\Delta$ & FID-50k \\
\midrule
Pixel flow (JiT, same network) & 92.8 & -- & 48.3 \\
+ detour path (\cref{sec:path}) & 66.6 & $-26.2$ & -- \\
+ trig.\ mix (\cref{sec:design}) & 52.7 & $-13.9$ & -- \\
+ norm bump $\dagger$ (\cref{sec:design}) & 49.1 & $-3.6$ & -- \\
+ fitted logit-normal $t$ (\cref{sec:objective}) & 47.9 & $-1.2$ & 23.4 \\
+ interval sampling (\cref{sec:objective}) & 45.5 & $-2.4$ & 20.4 \\
+ multi-layer code (\cref{sec:k4}) & \textbf{38.8} & $-6.7$ & \textbf{15.8} \\
% \midrule
% \color{gray}JiT on RAE latent (21 ep., decoder) & \color{gray}12.8 & & \\
\bottomrule
\end{tabular}
\end{adjustbox}
\vspace{-5mm}
\end{table}

\begin{figure*}[t]
  \centering
  \includegraphics[width=0.95\linewidth]{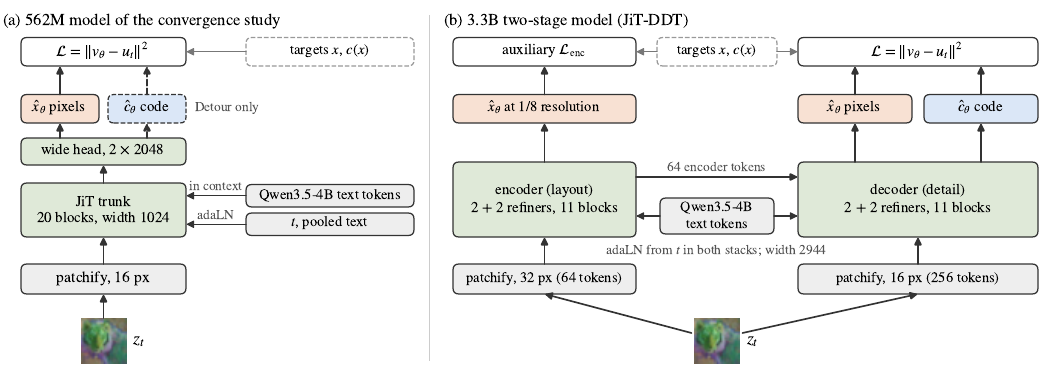}
  \caption{\textbf{Text-to-image architectures.} (a)~The 562M model of convergence study, trained on detour path or with plain pixel flow (no code output). (b)~The 3.3B model: a 32-pixel encoder lays out image and passes its tokens to a 16-pixel decoder that predicts  image \&  code.}
  \label{fig:t2i_arch}
  % \vspace{-5mm}
\end{figure*}

\begin{table*}[t]
\centering
\caption{\textbf{Text-to-image comparison} on GenEval (no prompt rewriting) and DPG-Bench. Gray: latent models; bold: best pixel-space scores; --: unreported. Baselines from original papers or Tuna-2~\cite{liu2026tuna}, SANA~\cite{xie2410sana}, Janus-Pro~\cite{chen2025janus}, and Qwen-Image~\cite{wu2025qwen}.}
\label{tab:t2i}
\footnotesize
\setlength{\tabcolsep}{3.2pt}
\begin{adjustbox}{max width=\linewidth}
\begin{tabular}{@{}lcc ccccccc cccccc@{}}
\toprule
 & & & \multicolumn{7}{c}{GenEval} & \multicolumn{6}{c}{DPG-Bench} \\
\cmidrule(lr){4-10}\cmidrule(lr){11-16}
Method & Params & Res & Single & Two & Count & Colors & Position & Attr. & Overall & Global & Entity & Attribute & Relation & Other & Overall \\
\midrule
% \color{gray}Janus-Pro\cite{chen2025janus} & \color{gray}7B & \color{gray}384 & \color{gray}0.99 & \color{gray}0.89 & \color{gray}0.59 & \color{gray}0.90 & \color{gray}0.79 & \color{gray}0.66 & \color{gray}0.80 & \color{gray}86.9 & \color{gray}88.9 & \color{gray}89.4 & \color{gray}89.3 & \color{gray}89.5 & \color{gray}84.2 \\
% \color{gray}SDXL\cite{podell2024sdxl} & \color{gray}2.6B & \color{gray}1024 & \color{gray}0.98 & \color{gray}0.74 & \color{gray}0.39 & \color{gray}0.85 & \color{gray}0.15 & \color{gray}0.23 & \color{gray}0.55 & \color{gray}83.3 & \color{gray}82.4 & \color{gray}80.9 & \color{gray}86.8 & \color{gray}80.4 & \color{gray}74.7 \\
% \color{gray}PixArt-$\Sigma$\cite{chen2024pixart} & \color{gray}0.6B & \color{gray}1024 & \color{gray}-- & \color{gray}-- & \color{gray}-- & \color{gray}-- & \color{gray}-- & \color{gray}-- & \color{gray}0.54 & \color{gray}86.9 & \color{gray}82.9 & \color{gray}88.9 & \color{gray}86.6 & \color{gray}87.7 & \color{gray}80.5 \\
\color{gray}SD3-Medium\cite{esser2024scaling} & \color{gray}2B & \color{gray}1024 & \color{gray}0.98 & \color{gray}0.74 & \color{gray}0.63 & \color{gray}0.67 & \color{gray}0.34 & \color{gray}0.36 & \color{gray}0.62 & \color{gray}87.9 & \color{gray}91.0 & \color{gray}88.8 & \color{gray}80.7 & \color{gray}88.7 & \color{gray}84.1 \\
\color{gray}FLUX.1-dev\cite{labs2025flux1kontextflowmatching} & \color{gray}12B & \color{gray}1024 & \color{gray}0.98 & \color{gray}0.81 & \color{gray}0.74 & \color{gray}0.79 & \color{gray}0.22 & \color{gray}0.45 & \color{gray}0.66 & \color{gray}74.4 & \color{gray}90.0 & \color{gray}89.0 & \color{gray}90.9 & \color{gray}88.3 & \color{gray}83.8 \\
\color{gray}SANA-1.6B\cite{xie2410sana} & \color{gray}1.6B & \color{gray}1024 & \color{gray}0.99 & \color{gray}0.77 & \color{gray}0.62 & \color{gray}0.88 & \color{gray}0.21 & \color{gray}0.47 & \color{gray}0.66 & \color{gray}86.0 & \color{gray}91.5 & \color{gray}88.9 & \color{gray}91.9 & \color{gray}90.7 & \color{gray}84.8 \\
\color{gray}Lumina-Image 2.0\cite{qin2025lumina} & \color{gray}2.6B & \color{gray}1024 & \color{gray}-- & \color{gray}0.87 & \color{gray}0.67 & \color{gray}-- & \color{gray}-- & \color{gray}0.62 & \color{gray}0.73 & \color{gray}-- & \color{gray}92.0 & \color{gray}90.2 & \color{gray}94.9 & \color{gray}-- & \color{gray}87.2 \\
\color{gray}Z-Image-Turbo\cite{cai2025z} & \color{gray}6B & \color{gray}1024 & \color{gray}1.00 & \color{gray}0.95 & \color{gray}0.77 & \color{gray}0.89 & \color{gray}0.65 & \color{gray}0.68 & \color{gray}0.82 & \color{gray}91.3 & \color{gray}89.6 & \color{gray}90.1 & \color{gray}92.2 & \color{gray}88.7 & \color{gray}84.9 \\
\color{gray}Qwen-Image\cite{wu2025qwen} & \color{gray}20B & \color{gray}1024 & \color{gray}0.99 & \color{gray}0.92 & \color{gray}0.89 & \color{gray}0.88 & \color{gray}0.76 & \color{gray}0.77 & \color{gray}0.87 & \color{gray}91.3 & \color{gray}91.6 & \color{gray}92.0 & \color{gray}94.3 & \color{gray}92.7 & \color{gray}88.3 \\
\midrule
% HiDream-O1-Image\refme{HiDream-O1} & 8B & GenEval 0.90 & DPG 89.8 (excluded)
PixNerd-XXL/16\cite{wang2026pixnerd} & 1.2B & 512 & 0.97 & 0.86 & 0.44 & 0.83 & 0.71 & 0.53 & 0.73 & 80.5 & 87.9 & 87.2 & 91.3 & 72.8 & 80.9 \\
PixelDiT\cite{yu2026pixeldit} & 1.3B & 512 & \textbf{1.00} & 0.94 & 0.70 & 0.90 & 0.53 & 0.65 & 0.78 & 88.0 & 90.9 & 87.6 & 89.8 & 88.5 & 83.7 \\
PixelGen-XXL/16\cite{ma2026pixelgen} & 1.1B & 512 & 0.99 & 0.88 & 0.59 & 0.90 & 0.70 & 0.70 & 0.79 & -- & -- & -- & -- & -- & -- \\
DeCo-XXL/16\cite{ma2026deco} & 1.1B & 512 & \textbf{1.00} & 0.92 & 0.72 & 0.91 & 0.80 & 0.79 & 0.86 & -- & -- & -- & -- & -- & 81.4 \\
RG-T2I-L/16\cite{pan2026learning} & 0.9B & 512 & -- & -- & -- & -- & -- & -- & 0.89 & -- & -- & -- & -- & -- & 87.1 \\
MiniT2I-L/16\cite{wangminit2i} & 0.9B & 512 & -- & -- & -- & -- & -- & -- & 0.88 & -- & -- & -- & -- & -- & 85.9 \\
PixelFlow-XL/4\cite{chen2025pixelflow} & 0.9B & 512 & -- & -- & -- & -- & -- & -- & 0.60 & -- & -- & -- & -- & -- & 77.9 \\
Iris\cite{cai2026iris} & 3B & 1024 & -- & -- & -- & -- & -- & -- & 0.80 & -- & -- & -- & -- & -- & 86.5 \\
Tuna-2\cite{liu2026tuna} & 7B & 1024 & 0.99 & 0.96 & 0.80 & 0.91 & 0.84 & 0.76 & 0.87 & \textbf{89.5} & 91.4 & \textbf{92.1} & 91.9 & \textbf{88.8} & 86.5 \\
\midrule
% Both rows are the SAME stage C model (iteration 200000), evaluated at the two resolutions it was
% trained on. The previous single row (0.91 / 88.0) was the stage B 180000 model at 512 -- a different
% checkpoint from the one the gallery and the 1024 row come from.
\textbf{\method} & 3.3B & 512 & \textbf{1.00} & 0.96 & \textbf{0.83} & \textbf{0.93} & \textbf{0.86} & 0.81 & \textbf{0.90} & 85.3 & \textbf{92.5} & 90.7 & \textbf{94.9} & 85.7 & \textbf{87.8} \\
\textbf{\method} & 3.3B & 1024 & 0.98 & \textbf{0.98} & 0.78 & 0.89 & 0.83 & \textbf{0.82} & 0.88 & 85.3 & 92.3 & 90.4 & \textbf{94.9} & 85.9 & 87.4 \\
\bottomrule
\end{tabular}
\end{adjustbox}
\vspace{-4mm}
\end{table*}

\begin{figure*}[t]
  \centering
  \begin{minipage}[t]{0.60\textwidth}
    \vspace{0pt}
    \centering
    \includegraphics[width=\linewidth]{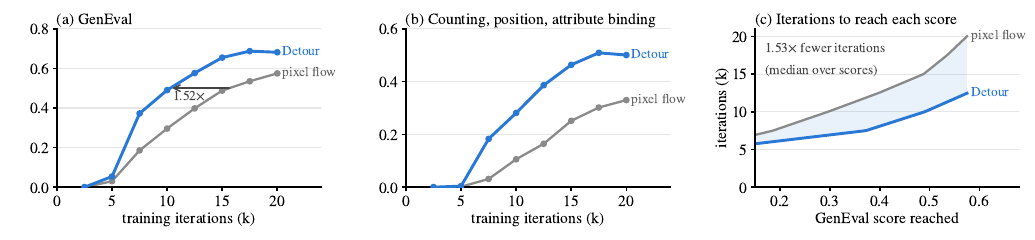}
    \caption{\textbf{\method learns semantics faster than pixel flow}, same network. (a)~GenEval, (b)~counting, position and attribute binding, (c)~iterations to first reach each GenEval score.}
    \label{fig:t2i_conv}
  \end{minipage}\hfill
  \begin{minipage}[t]{0.34\textwidth}
    \vspace{0pt}
    \centering
    \captionof{table}{\textbf{Guidance} (\method-XL/16, epoch 200).}
    \label{tab:guidance}
    \scriptsize
    \setlength{\tabcolsep}{4pt}
\begin{adjustbox}{max width=\linewidth}
\begin{tabular}{@{}lcc@{}}
\toprule
Sampling & FID-50k & IS \\
\midrule
Unguided & 3.69 & 182 \\
CFG 2.25 on $[0.05,0.9]$ & 2.27 & 283 \\
AutoGuidance 2.0 & 1.81 & 210 \\
\textbf{CFG 1.75 + AutoGuidance 2.0} & \textbf{1.43} & 267 \\
\bottomrule
\end{tabular}
\end{adjustbox}
  \end{minipage}
\vspace{-5mm}
\end{figure*}

\paragraph{Setup.}
We train class-conditional models on ImageNet \cite{russakovsky2015imagenet} at $256{\times}256$ (JiT-B/16 with 296M parameters and JiT-XL/16 with 849M) and at $512{\times}512$ (JiT-XL/32 with 849M, with the noise scaled by 2 \cite{hoogeboom2023simple}). All models use a batch size of 1024, AdamW, a constant learning rate of $2{\times}10^{-4}$ and an EMA decay of 0.9996 (\cref{sec:supp_impl}).
We sample with 100 Euler steps on the grid $t=u^4$. Following JiT \cite{li2026back}, we compute FID \cite{heusel2017gans} and IS \cite{salimans2016improved} with torch-fidelity against the ImageNet training statistics; precision and recall \cite{kynkaanniemi2019improved} use the ADM evaluator \cite{dhariwal2021diffusion}. Ablations report FID-10k.

%-------------------------------------------------------------------------

%-------------------------------------------------------------------------

\subsection{Main results}
\label{sec:exp_main}
\Cref{tab:system} compares \method with pixel-space and latent-space models.
At $256{\times}256$, \method-XL/16 achieves FID-50k of \textbf{1.43} in 200 epochs, outperforming JiT-G (2.15), Dino Forcing-XL~\cite{ghosh2026dino} (2.13), and PixelGen~\cite{ma2026pixelgen} (1.83), all trained for at most 200 epochs. SiD2 reaches 1.38 but requires $6\times$ more epochs.
At $512{\times}512$, \method-XL/32 achieves \textbf{1.59} in 160 epochs, compared to JiT-G/32 (1.78, 600 epochs). Models with 16-pixel patches achieve lower FID but use $4\times$ more tokens and longer training.
Without cfg, \method achieves FID-50k of 3.69 and 4.19 at resolution $256$ and $512$.
%-------------------------------------------------------------------------

\subsection{Does our Detour speed up pixel diffusion?}
\label{sec:exp_recipe}
\Cref{tab:recipe} progressively adds improvements to plain pixel flow using the \emph{same} network. Incorporating DINOv2 features into the path yields the largest FID gain ($92.8\to66.6$), while an even, norm-preserving hand-over further reduces FID by 13.9. Subsequent improvements retain their gains at 80 epochs. The full recipe achieves FID-10k of 38.8 after 20 epochs and unguided FID-50k of 15.8 after 80 epochs. Although semantic latent diffusion still converges faster, \method recovers much of this advantage while generating directly in pixel space.

%-------------------------------------------------------------------------

\subsection{Guidance}
\label{sec:exp_guidance}
The guidance results (\cref{tab:guidance}) support the analysis of \cref{sec:guidance}. CFG alone only reaches FID-50k 2.27: variants such as APG~\cite{sadat2024eliminating}, separate scales for code and pixels, or rescaling change FID-10k by less than 0.03, and its FID-50k stops improving after epoch 160 while the unguided FID keeps decreasing.
AutoGuidance alone already outperforms CFG (1.81) at a much lower IS, and the combination reaches 1.43.
The same sampler transfers to $512{\times}512$; only the CFG scale changes ($1.75\to2.0$). There, CFG alone reaches FID-50k 2.54 and the combination 1.59. \Cref{sec:supp_guidance} gives the sweeps, the per-direction measurements and the variants that did not help.
The appendix adds code feedback (\cref{sec:supp_feedback}), one-step distillation (\cref{sec:distill,sec:supp_dmd}), other encoders and code widths (\cref{sec:supp_codedim}).

%% file: sec/6_extensions.tex
\section{Text-to-Image Generation}
\label{sec:t2i}
The detour path does not depend on class labels, so it applies directly to text conditioning. We train a 3.3B-parameter pixel transformer with the ImageNet recipe. It follows the two-stage JiT-DDT design~\cite{chopra2026pyramidjit} (\cref{fig:t2i_arch}b; \cref{sec:supp_t2i_arch}): an encoder on 32-pixel patches lays out the image, and a decoder on 16-pixel patches predicts the image and the code. Text is encoded by a frozen Qwen3.5-4B \cite{team2026qwen3}.

\paragraph{Data and training.}
We assemble about 31M image--text pairs from public sources, 18M real photographs and 13M synthetic.
Training goes up in resolution in four stages: 50k iterations at $256{\times}256$ on 2M synthetic
images~\cite{naeem_khoshnevis_2026}, 80k more at $256$ on the full corpus, 50k at about $512$, and
20k at about $1024$, the last in 17 aspect-ratio buckets so non-square images are not cropped. Details on the mixes, the captions, the resolution handling and the cost are in \Cref{sec:supp_t2i_arch}.

\paragraph{Results.}
At $512{\times}512$ the model reaches $0.90$ on GenEval~\cite{ghosh2023geneval} and $87.8$ on
DPG-Bench~\cite{hu2024ella}, and $0.88$ / $87.4$ at $1024{\times}1024$ (\cref{tab:t2i}); samples are
in \cref{fig:t2i_gallery} and \cref{sec:supp_t2i_gallery}. The $1024$ stage trains shortest and
last, and scores below the $512$ one on both benchmarks. Both scores are above the pixel-space
models of \cref{tab:t2i}, by most on counting and on DPG-Bench relations, though some of them score
higher on the DPG-Bench global and other categories; the model is also above the 20B Qwen-Image on
GenEval ($0.90$ against $0.87$) and $0.5$ below it on DPG-Bench, with six times fewer parameters.
Unlike RAE \cite{zheng2026diffusion} and its text-to-image extension~\cite{singh2026improved}, \method samples pixels directly, so its detail is not bounded by a feature decoder: the published RAE decoder reconstructs held-out images of our data at only 18.2\,dB PSNR \cite{singh2026improved, zheng2026diffusion}, and with the 562M network below, the high-frequency gap to real images is 0.149 on the RAE latent but 0.058 on our detour  (\cref{sec:supp_t2i_conv}).

\paragraph{Convergence against pixel flow.}
To isolate the effect of the diffusion path, we train the same 562M network (\cref{fig:t2i_arch}a) on 2M synthetic images at $256{\times}256$, using either \method or plain pixel flow with identical data and optimization.
\method reaches the same GenEval scores with a median of $1.53\times$ fewer iterations (\cref{fig:t2i_conv}c), reaching 0.5 in 10.3k versus 15.6k iterations.
The largest gains are in composition (\cref{fig:t2i_conv}b): at 10k iterations, \method achieves 39\% versus 16\% on counting and 21\% versus 4.5\% on attribute binding, with higher DPG-Bench scores (72.0 vs.\ 66.0).
With identical noise, pixel flow still merges objects that \method correctly separates (\cref{fig:supp_t2i_samples}). The advantage persists at 30k iterations (GenEval 0.76 vs.\ 0.70; \cref{tab:supp_t2i_conv}).

%% file: sec/7_conclusion.tex
\section{Conclusion}
\label{sec:conclusion}
At high noise, the posterior mean is blurry in pixels but retains semantic information in DINOv2 features. Motivated by this, \method routes the diffusion path through semantic features near noise and transitions to pixels near data, without requiring a decoder. On ImageNet, it achieves FID-50k of 1.43 at $256{\times}256$ and 1.59 at $512{\times}512$. A 3.3B text-to-image model outperforms the compared pixel-space models on GenEval and DPG-Bench.

\paragraph{LLM Usage Statement.}
We used Claude to assist with coding and implementation, and ChatGPT to improve the clarity and readability of the manuscript. All technical contributions, experimental results, and scientific conclusions were developed and verified by the authors.

%% file: sec/X_suppl.tex
\clearpage
\setcounter{page}{1}
\maketitlesupplementary
\appendix

% =====================================================================

\noindent\textbf{Contents of the appendix.}
\begin{itemize}\itemsep0pt \parskip0pt \small
    \appendixitem{sec:supp_theory}
    \appendixitem{sec:supp_impl}
    \appendixitem{sec:supp_probe}
    \appendixitem{sec:supp_design}
    \appendixitem{sec:supp_stage}
    \appendixitem{sec:supp_code}
    \appendixitem{sec:supp_codedim}
    \appendixitem{sec:supp_solver}
    \appendixitem{sec:supp_guidance}
    \appendixitem{sec:supp_feedback}
    \appendixitem{sec:supp_dmd}
    \appendixitem{sec:supp_t2i_arch}
    \appendixitem{sec:supp_t2i_conv}
    \appendixitem{sec:supp_samples}
    \appendixitem{sec:supp_limits}
    \appendixitem{sec:supp_t2i_gallery}
\end{itemize}

\section{Derivations}
% \tuan{hay lam 1 cai paragraph o day de list ra la derive nhung cai gi va de lam gi lol}
\label{sec:supp_theory}

\paragraph{Velocity identity and endpoint behavior.}
From \cref{eq:path}, $\varepsilon=(z_t-t\,m_t)/(1-t)$. Substituting into \cref{eq:vel} yields \cref{eq:identity}, where
\begin{equation}
a_t=\frac{\alpha_t}{1-t}+t\alpha_t',
\qquad
b_t=\frac{\beta_t}{1-t}+t\beta_t'.
\end{equation}
We derive the endpoint limits for continuously differentiable mixing weights satisfying $\alpha_0=0$, $\beta_0=1$, $\alpha_1=1$, and $\beta_1=0$.

As $t\to0$, continuity of the weights and their derivatives gives $a_t\to0$ and $b_t\to1$.

As $t\to1$, we use the first-order Taylor expansion of a function $f$ around $t=1$:
\begin{equation}
f(t)=f(1)+f'(1)(t-1)+o(1-t),
\end{equation}
where $o(1-t)$ denotes a remainder that vanishes faster than $1-t$. Applying this to $\alpha_t$ and $\beta_t$, with $\alpha_1=1$ and $\beta_1=0$, gives
\begin{equation}
\begin{aligned}
\alpha_t &= 1-(1-t)\alpha_1'+o(1-t),\\
\beta_t &= -(1-t)\beta_1'+o(1-t).
\end{aligned}
\end{equation}
Substituting into the coefficients and using $\alpha_t'\to\alpha_1'$ and $\beta_t'\to\beta_1'$ yields
\begin{equation}
\begin{aligned}
a_t &= \frac{1}{1-t}-\alpha_1'+t\alpha_t'+o(1)
     =\frac{1}{1-t}+o(1),\\
b_t &= -\beta_1'+t\beta_t'+o(1)=o(1).
\end{aligned}
\end{equation}
Thus, $a_t\sim1/(1-t)$ and $b_t\to0$. These limits hold for any continuously differentiable hand-over schedule satisfying the endpoint conditions.

By the conditional flow-matching theorem~\cite{lipman2022flow}, the optimal velocity is
\begin{equation}
v^\star(z_t,t,y)
=a_t\,\mathbb E[x\mid z_t,t,y]
+b_t\,\Phi\mathbb E[c(x)\mid z_t,t,y]
-\frac{z_t}{1-t}.
\end{equation}
Its ODE transports Gaussian noise to the conditional data distribution, enabling sampling without an encoder.

\paragraph{Score derivation.}
From \cref{eq:path}, the noisy state conditioned on a clean image $x$ follows
\begin{equation}
p_t(z\mid x)=\mathcal N\bigl(z;\,t\,m_t(x),(1-t)^2I\bigr),
\end{equation}
where $m_t(x)=\alpha_t x+\beta_t\Phi c(x)$. The score of this Gaussian is
\begin{equation}
\nabla_z\log p_t(z\mid x)
=-\frac{z-t\,m_t(x)}{(1-t)^2}
=-\frac{\varepsilon}{1-t}.
\end{equation}
To obtain the marginal score, we integrate over the unknown clean image. Since
$p_t(z)=\int p_t(z\mid x)p(x)\,dx$, differentiating under the integral gives
\begin{equation}
\begin{aligned}
\nabla_z\log p_t(z)
&=\frac{\int \nabla_z p_t(z\mid x)p(x)\,dx}{p_t(z)}\\
&=\int \nabla_z\log p_t(z\mid x)\,p(x\mid z)\,dx\\
&=\mathbb E[\nabla_z\log p_t(z\mid x)\mid z].
\end{aligned}
\end{equation}
Substituting the Gaussian score yields
\begin{equation}
\boxed{
\nabla_z\log p_t(z)
=-\frac{\mathbb E[\varepsilon\mid z]}{1-t}
=\frac{t\,\mathbb E[m_t(x)\mid z]-z}{(1-t)^2}.
}
\end{equation}
By linearity of conditional expectation,
\begin{equation}
\mathbb E[m_t(x)\mid z]
=\alpha_t\mathbb E[x\mid z]
+\beta_t\Phi\mathbb E[c(x)\mid z].
\end{equation}
We estimate this quantity using the network's two predictions,
$\hat m_t=\alpha_t\hat x_\theta+\beta_t\Phi\hat c_\theta$, giving the plug-in score estimate
\begin{equation}
\boxed{
\hat s_\theta(z,t)=\frac{t\hat m_t-z}{(1-t)^2}.
}
\end{equation}
This estimate is exact when both heads recover their respective posterior means. However, the velocity loss identifies only $a_t\hat x_\theta+b_t\Phi\hat c_\theta$, not the two heads separately. Thus, an optimal velocity prediction does not necessarily imply an exact score estimate. We use this estimate for distillation (\cref{sec:supp_dmd}).

\paragraph{Only the combination is identified.}
The velocity loss $\mathbb E\|a_t\delta x+b_t\Phi\delta c\|^2$, where $\delta x=\hat x-x$ and $\delta c=\hat c-c$, constrains only the weighted sum $a_t\hat x+b_t\Phi\hat c$, not the individual predictions. Indeed, errors satisfying $a_t\delta x=-b_t\Phi\delta c$ cancel exactly. This ambiguity does not affect the optimal sampling velocity, which depends only on the sum, but may affect the score estimate derived from $\hat m_t=\alpha_t\hat x+\beta_t\Phi\hat c$. A decoupled loss,
\begin{equation}
\mathbb E\left[a_t^2\|\delta x\|^2+b_t^2\|\Phi\delta c\|^2\right],
\end{equation}
has the same optimal velocity while separately identifying the posterior means of both targets wherever their weights are nonzero, assuming sufficient model capacity.

\paragraph{Where the schedule enters the loss.}
Decomposing the predicted velocity into the sum and difference of its two predictions gives
\begin{equation}
v_\theta =
\frac{a_t+b_t}{2}(\hat x_\theta+\Phi\hat c_\theta)
+\frac{a_t-b_t}{2}(\hat x_\theta-\Phi\hat c_\theta)
-\frac{z_t}{1-t}.
\end{equation}
For a convex hand-over $(\alpha_t,\beta_t)=(w_t,1-w_t)$, the coefficient of the difference becomes
\begin{equation}
\frac{a_t-b_t}{2}
=\frac{2w_t-1}{2(1-t)}+t\,w_t'.
\end{equation}
Thus, a rapid hand-over (large $w_t'$) amplifies errors in $\hat x_\theta-\Phi\hat c_\theta$ over a narrow range of $t$. This motivates the evenness criterion in \cref{sec:design}, although other factors may also affect performance.

\paragraph{Trajectory geometry.}
Near noise, the path expands as
\begin{equation}
z_t=\varepsilon+t(\Phi c(x)-\varepsilon)+O(t^2),
\end{equation}
since $\alpha_0=0$ and $\beta_0=1$. The image contribution enters only at order $t\alpha_t=O(t^2)$. Thus, early sampling steps primarily follow semantic features, while pixel details emerge later. This introduces curvature relative to straight pixel flow, potentially requiring more steps for accurate numerical integration.

\paragraph{Bound on the CFG direction.}
For $f\in\{x,c\}$, the law of total expectation gives
\begin{equation}
\mathbb E[f\mid z_t]
=\sum_{y'}p(y'\mid z_t)\mathbb E[f\mid z_t,y'].
\end{equation}
Hence, the conditional-unconditional difference is
\begin{equation}
\begin{aligned}
\Delta f
&=\mathbb E[f\mid z_t,y]-\mathbb E[f\mid z_t]\\
&=\sum_{y'\neq y}p(y'\mid z_t)
\bigl(\mathbb E[f\mid z_t,y]-\mathbb E[f\mid z_t,y']\bigr).
\end{aligned}
\end{equation}
Taking norms and bounding the class-conditional differences gives
\begin{equation}
\|\Delta f\|
\leq (1-p(y\mid z_t))
\max_{y'\neq y}
\|\mathbb E[f\mid z_t,y]-\mathbb E[f\mid z_t,y']\|.
\end{equation}
Thus, the CFG direction is bounded by the remaining class uncertainty, yielding \cref{eq:cfg_vanish}.

% =====================================================================
\section{Implementation details}
\label{sec:supp_impl}

\Cref{tab:supp_models} summarizes the model configurations, while \cref{tab:supp_impl} details the training and sampling settings. These settings apply unless otherwise noted.

\begin{table}[htbp]
\centering
\caption{\textbf{Model configurations.}}
\label{tab:supp_models}
\small
\setlength{\tabcolsep}{4pt}
\begin{adjustbox}{max width=\linewidth}
\begin{tabular}{@{}lcccccc@{}}
\toprule
Model & Res. & Patch & Depth & Width & Heads & Params \\
\midrule
JiT-B/16 & 256 & 16 & 12 & 768 & 12 & 296M \\
JiT-XL/16 & 256 & 16 & 28 & 1152 & 16 & 849M \\
JiT-XL/32 & 512 & 32 & 28 & 1152 & 16 & 849M \\
\midrule
Output head & \multicolumn{6}{l}{2 blocks, width 2048, 16 heads; re-reads input patches} \\
\bottomrule
\end{tabular}
\end{adjustbox}
\end{table}

\begin{table}[htbp]
\centering
\caption{\textbf{Training and sampling settings.}}
\label{tab:supp_impl}
\small
\setlength{\tabcolsep}{3pt}
\begin{adjustbox}{max width=\linewidth}
\begin{tabular}{@{}lp{5.6cm}@{}}
\toprule
Data & ImageNet-1k training set; center crop, horizontal flip \\
Code & DINOv2-B with registers at $224{\times}224$ ($16{\times}16$ tokens); average of layer-normalized tokens from blocks 6/8/10/12 plus the mean final-layer token; mean-centered, globally scaled to std $A{=}0.5$; PixelShuffle(16) \\
Path & Trigonometric with norm bump, $\kappa{=}0.25$ \\
$t$ sampler & Logit-normal$(-0.63,0.85)$; 10\% sampled from $\mathcal U[0,0.1]$ \\
Optimizer & AdamW $(0.9,0.95)$, constant lr $2{\times}10^{-4}$, 1-epoch warmup \\
Batch / labels & 1024 / 10\% null labels \\
Precision & bf16, \texttt{torch.compile} \\
EMA & 0.9996 for all reported results \\
Epochs & 200 (XL/16), 160 (XL/32); 20--80 (JiT-B ablations) \\
Noise scale & 1 at $256{\times}256$; 2 at $512{\times}512$ \\
Sampler & Euler, 100 steps, $t=u^4$ \\
Guidance & CFG $s$ + AutoGuidance $a{=}2$ using the model's epoch-20 EMA checkpoint; $s{=}1.75$ on $[0.05,0.9]$ (256), $s{=}2.0$ on $[0.05,0.75]$ (512) \\
Cost & JiT-B: $\sim$23 min/epoch on $4\times$A100-40GB. JiT-XL/16: 11.3 min/epoch on $8\times$H200, totaling 38 h ($\sim$13 GPU-days) for 200 epochs. JiT-XL/32 uses 32-pixel patches to retain 256 tokens/image; 160 epochs take $\sim$30 h (10 GPU-days) on $8\times$H200. \\
\bottomrule
\end{tabular}
\end{adjustbox}
\end{table}

\paragraph{Evaluation.}
FID and IS are computed with torch-fidelity against the ImageNet training statistics released with JiT~\cite{li2026back}. FID-10k is used to compare settings and FID-50k to confirm them. Precision and recall use the ADM evaluator and reference batch~\cite{dhariwal2021diffusion}. EMA weights are used throughout: at a constant learning rate, the FID of the raw weights fluctuates between checkpoints for \method, while the EMA gives a smooth trend.

% =====================================================================

\section{Probe test}
\label{sec:supp_probe}

We examine the class information in clean predictions
during sampling and distinguish its presence from its
recognizability by classifiers trained on real images.

\paragraph{Evaluation protocol.}
We train a linear probe on mean-pooled DINOv2-B
features from 50 real images per ImageNet class.
The probe reads latent predictions directly and
encodes pixel predictions with DINOv2-B.
We also evaluate RGB predictions using a pretrained
ResNet-50 (\texttt{IMAGENET1K\_V2}), decoding latent
predictions with the RAE decoder.
As references, the two classifiers achieve 76.2\%
and 89.9\% accuracy on real images, respectively,
and ResNet-50 achieves 87.3\% on RAE reconstructions.
\Cref{tab:supp_probe_models} evaluates class-conditioned
predictions over 100 classes with eight noise seeds
per class and 50-step Euler sampling.

\begin{table}[t]
\centering
\caption{\textbf{Class accuracy of clean predictions}
(\%). First (pure-noise) and final sampling steps,
evaluated by a DINOv2 linear probe and an RGB ResNet-50.}
\label{tab:supp_probe_models}
\setlength{\tabcolsep}{3pt}
\small
\begin{tabular}{llrrrr}
\toprule
 & & \multicolumn{2}{c}{Linear probe}
   & \multicolumn{2}{c}{ResNet-50} \\
\cmidrule(lr){3-4}\cmidrule(lr){5-6}
Space & Model & First & Last & First & Last \\
\midrule
Latent & RAE DiT$^{\mathrm{DH}}$-XL, final
    & \textbf{99.8} & 80.4 & 0.7 & 83.6 \\
Latent & RAE DiT$^{\mathrm{DH}}$-XL, ep.\ 80
    & 99.3 & 78.4 & 0.6 & 79.7 \\
Latent & RAE DiT$^{\mathrm{DH}}$-XL, ep.\ 20
    & 97.0 & 80.9 & 1.0 & 78.4 \\
Latent & RAE DiT$^{\mathrm{DH}}$-S, ep.\ 20
    & 95.2 & 74.5 & 0.7 & 61.9 \\
\midrule
Pixel & JiT-B/16 & 0.2 & 36.7 & 0.2 & 53.7 \\
Pixel & JiT-H/16 & 0.1 & 62.4 & 0.5 & 74.3 \\
Pixel & JiT-H/16, CFG & 0.1 & 82.9 & 0.5 & 91.7 \\
% \midrule
% \multicolumn{2}{l}{Class-mean DINOv2 feature}
%     & 100.0 & -- & 4.4$^{\dagger}$ & -- \\
% \multicolumn{2}{l}{Class-mean pixel image}
%     & -- & -- & 1.0 & -- \\
\bottomrule
\end{tabular}
% \vspace{1pt}

% {\footnotesize $^{\dagger}$After RAE decoding.}
\end{table}

\begin{figure}[t]
\centering
\includegraphics[width=\linewidth]{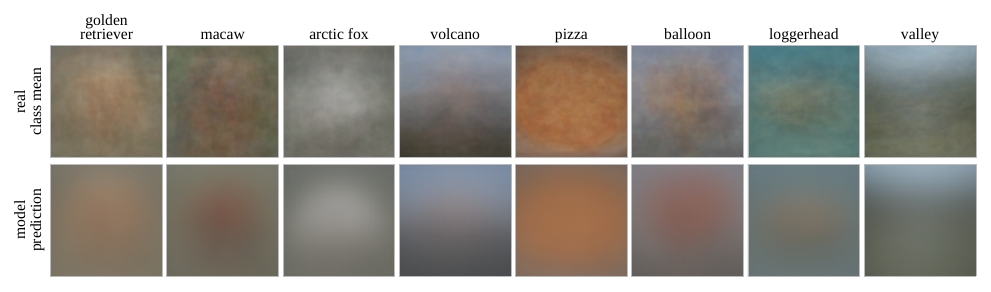}
\caption{\textbf{First-step pixel predictions resemble
class-mean images.} Top: mean of 100 real ImageNet
images per class. Bottom: JiT-H/16 predictions from
pure noise, averaged over 32 seeds. The predictions
correlate with the corresponding class means (0.85).}
\label{fig:supp_pixel_classmean}
\end{figure}

\paragraph{High-noise predictions.}
At $t=0$, both latent and pixel models approximate
their respective class-conditional means.
The latent prediction closely matches the DINOv2
class prototype (cosine similarity 0.98), achieving
99.8\% probe accuracy. This accuracy exceeds that
on individual real images because the prediction
approximates a class prototype rather than a
specific image; the DINOv2 class means themselves
achieve 100\% accuracy.
In contrast, JiT-H/16 approximates the class-mean
image (correlation 0.85, \cref{fig:supp_pixel_classmean}),
which is poorly recognized by real-image classifiers.

\paragraph{Class information versus recognizability.}
To account for the distribution shift between real
images and high-noise predictions, we retrain the
same probe on first-step model predictions using
24 training and eight held-out noise seeds per
class across all 1,000 classes, with seeds shared
across classes.
As shown in \cref{tab:supp_probe_retrain}, pixel
accuracy increases from 0.4\% to 53.8\%.
Thus, high-noise pixel predictions contain class
information, but this information is less readily
recognizable by classifiers trained on real images.

\begin{table}[t]
\centering
\caption{\textbf{First-step class accuracy} (\%) with
the same mean-pooled DINOv2 linear-probe architecture.
Only the probe's training data change; evaluation
uses eight held-out seeds per class over 1,000 classes.}
\label{tab:supp_probe_retrain}
\small
\begin{tabular}{lcc}
\toprule
Probe trained on & Pixel & Latent \\
\midrule
Real images & 0.4 & 99.5 \\
Model predictions & \textbf{53.8} & 100.0 \\
\bottomrule
\end{tabular}
\end{table}

\subsection{What drives faster convergence?}
\label{sec:supp_classmean}

We investigate whether class-conditional prototypes
alone explain Detour's convergence advantage.

\paragraph{Removing class prototypes.}
Let $\mu_y=\mathbb{E}[c(x)\mid y]$ and
$\mu=\mathbb{E}[c(x)]$ denote the class-conditional
and global feature means, computed at each spatial
position and channel. We compare two variants:
\begin{equation}
\begin{aligned}
c_{\mathrm{only}}(x)&=\mu_y,\\
c_{\mathrm{remove}}(x)&=c(x)-\mu_y+\mu.
\end{aligned}
\label{eq:supp_classmean}
\end{equation}
The first retains only the class prototype, while
the second removes the class-dependent mean and
preserves within-class variation up to global
rescaling.
We replace $c(x)$ in the Detour path and its
corresponding velocity target with each code
variant, leaving the training objective unchanged.
Both variants use JiT-B at $256\times256$ for
20 epochs with matched architecture, data,
hand-over schedule, sampler, and code variance.

\begin{table}[t]
\centering
\caption{\textbf{Contribution of class-conditional
prototypes.} FID-10k at 20 epochs (lower is better).
Gain retained is the fraction of Detour's observed
improvement over pixel flow.}
\label{tab:supp_classmean}
\small
\setlength{\tabcolsep}{4pt}
\begin{tabular}{llcc}
\toprule
Method & Code & FID $\downarrow$ & Gain \\
\midrule
Pixel flow & -- & 92.80 & 0\% \\
Prototype only & $\mu_y$ & 86.41 & 13.5\% \\
Prototype removed & $c-\mu_y+\mu$ & 54.22 & 81.6\% \\
Detour & $c$ & \textbf{45.50} & 100\% \\
\bottomrule
\end{tabular}
\end{table}

As shown in \cref{tab:supp_classmean}, retaining
only class prototypes recovers 13.5\% of the
observed FID improvement over pixel flow, whereas
removing them retains 81.6\%.
Thus, class prototypes alone do not explain
Detour's convergence advantage.

\paragraph{Semantic information beyond class means.}
For DINOv2 patch features
$C\in\mathbb{R}^{P\times D}$, we compare linear
probes on mean-pooled and channel Gram features:
\begin{equation}
\bar c=\frac{1}{P}\sum_{p=1}^{P}C_p,
\qquad
G(C)=\frac{1}{P}C^\top C.
\label{eq:supp_gram}
\end{equation}

\begin{table}[t]
\centering
\caption{\textbf{Class information in real DINOv2
codes.} Top-1 accuracy (\%) over 1,000 ImageNet
classes; chance is 0.1\%.}
\label{tab:supp_classmean_probe}
\small
\begin{tabular}{lcc}
\toprule
Readout & Original $c$ & $c-\mu_y+\mu$ \\
\midrule
Pooled features & 75.33 & 4.69 \\
Gram features & 72.16 & \textbf{73.23} \\
\bottomrule
\end{tabular}
\end{table}

Removing class means substantially reduces
pooled-feature accuracy but preserves class
information accessible through Gram features
(\cref{tab:supp_classmean_probe}).
For the prototype-removed model, Gram accuracy
increases from chance at $t=0$ to 58.22\% at
$t=0.1$, while its pixel prediction still
achieves only 0.67\%.

\paragraph{Takeaway:} Our analysis in \cref{sec:prelim} shows that
semantic posterior means are readily recognizable
at high noise, providing the initial motivation
for Detour. However, these results suggest that Detour benefits from
semantic structure beyond class prototypes.
Even after removing class-dependent means,
semantic information emerges early in the
code predictions, potentially facilitating
subsequent pixel generation.

% =====================================================================

\section{Path design}
\label{sec:supp_design}

\paragraph{Experimental setup.}
Unless otherwise specified, the ablations in this
section use Heun sampling with 50 uniform steps,
without interval sampling or multi-layer codes.

\subsection{Hand-over schedules}
\label{sec:supp_sched}

We compare schedules with different hand-over rates
and signal magnitudes (\cref{fig:supp_schedules}).
Let $u=\mathrm{clip}((t-t_1)/(t_2-t_1),0,1)$
denote the position within the hand-over window,
with $[t_1,t_2]=[0,1]$ by default.

\paragraph{Schedule families.}
Convex schedules use
$(\alpha_t,\beta_t)=(w_t,1-w_t)$, where
\begin{equation}
\begin{aligned}
\text{Linear:}\quad &w_t=u,\\
\text{Smoothstep:}\quad &w_t=3u^2-2u^3.
\end{aligned}
\end{equation}
\textsc{vpsmooth} normalizes the smoothstep weights
to preserve the coefficient norm:
\begin{equation}
(\alpha_t,\beta_t)=
\frac{(w_t,1-w_t)}{\sqrt{w_t^2+(1-w_t)^2}}.
\end{equation}
The trigonometric family uses
\begin{equation}
(\alpha_t,\beta_t)=g_t(\sin\theta_t,\cos\theta_t),
\quad \theta_t=\frac{\pi t}{2},
\end{equation}
where $g_t=1$ for \textsc{trig} and
$g_t=1+\kappa\sin(2\theta_t)$ for \textsc{trigbump}.
The gain $g_t$ changes the signal magnitude without
affecting the relative pixel weight
$\alpha_t/(\alpha_t+\beta_t)$.

As shown in \cref{tab:handover}, both a more uniform
hand-over and norm preservation improve FID, with
trigonometric mixing providing the strongest baseline.

\paragraph{Norm bump ablation.}
We vary $\kappa$ to adjust the midpoint signal
magnitude while keeping the hand-over timing fixed.
\Cref{tab:supp_bump} shows that $\kappa=0.25$
performs best.

\begin{table}[htbp]
\centering
\caption{\textbf{Norm bump ablation}
(JiT-B, 20 epochs, unguided FID-10k).
All runs use fitted logit-normal time sampling
and learning-rate warmup.}
\label{tab:supp_bump}
\small
\setlength{\tabcolsep}{8pt}
\begin{tabular}{@{}ccc@{}}
\toprule
$\kappa$ & Midpoint norm & FID-10k $\downarrow$ \\
\midrule
0 & 1.00 & 50.2 \\
0.25 & 1.25 & \textbf{47.9} \\
0.50 & 1.50 & 48.3 \\
\bottomrule
\end{tabular}
\end{table}

\section{The class window \& interval sampling}
\label{sec:supp_stage}

We investigate when class identity is determined during
sampling and how this observation informs classifier-free
guidance and training-time sampling.

\paragraph{Effect of guidance intervals.}
We generate 1,000 samples (one per class) using the
80-epoch trigonometric model, with fixed noise across
settings and CFG scale 3 restricted to different time
intervals. To measure the dependence of code predictions
on class conditioning, we compute
\begin{equation}
G(t)=\mathbb E\left[
\frac{\|\hat c_\theta(z_t,t,y)
-\hat c_\theta(z_t,t,\varnothing)\|}
{\|\hat c_\theta(z_t,t,y)\|}
\right],
\end{equation}
averaged over 1,000 unguided trajectories.
A decreasing gap indicates increasing agreement between
conditional and unconditional code predictions.

\begin{table}[htbp]
\centering
\caption{\textbf{Effect of guidance intervals.}
(JiT-B, trig, 80 epochs; 1,000 samples).
Top: ResNet-50 accuracy with CFG restricted to
different intervals. Corrected: fraction of unguided
classification failures corrected by CFG.
Bottom: relative difference between conditional
and unconditional code predictions.}
\label{tab:supp_window}
\small
\setlength{\tabcolsep}{3.5pt}
\begin{adjustbox}{max width=\linewidth}
\begin{tabular}{@{}lcc@{}}
\toprule
CFG interval & Top-1 (\%) & Corrected (\%) \\
\midrule
None & 44.3 & -- \\
$[0,0.05)$ & 75.6 & 60 \\
$[0.05,0.1)$ & 60.8 & 33 \\
$[0.1,1]$ & 59.6 & 32 \\
$[0,0.1)$ & 78.4 & 65 \\
$[0,1]$ & \textbf{85.3} & \textbf{76} \\
\bottomrule
\end{tabular}
\end{adjustbox}

\vspace{4pt}
\begin{adjustbox}{max width=\linewidth}
\begin{tabular}{@{}lcccccccc@{}}
\toprule
$t$ & .01 & .02 & .032 & .048 & .068 & .09 & .16 & .5 \\
\midrule
$G(t)$ & .99 & .95 & .79 & .50 & .30 & .19 & .08 & .01 \\
\bottomrule
\end{tabular}
\end{adjustbox}
\end{table}

\begin{figure}[htbp]
\centering
\includegraphics[width=\linewidth]{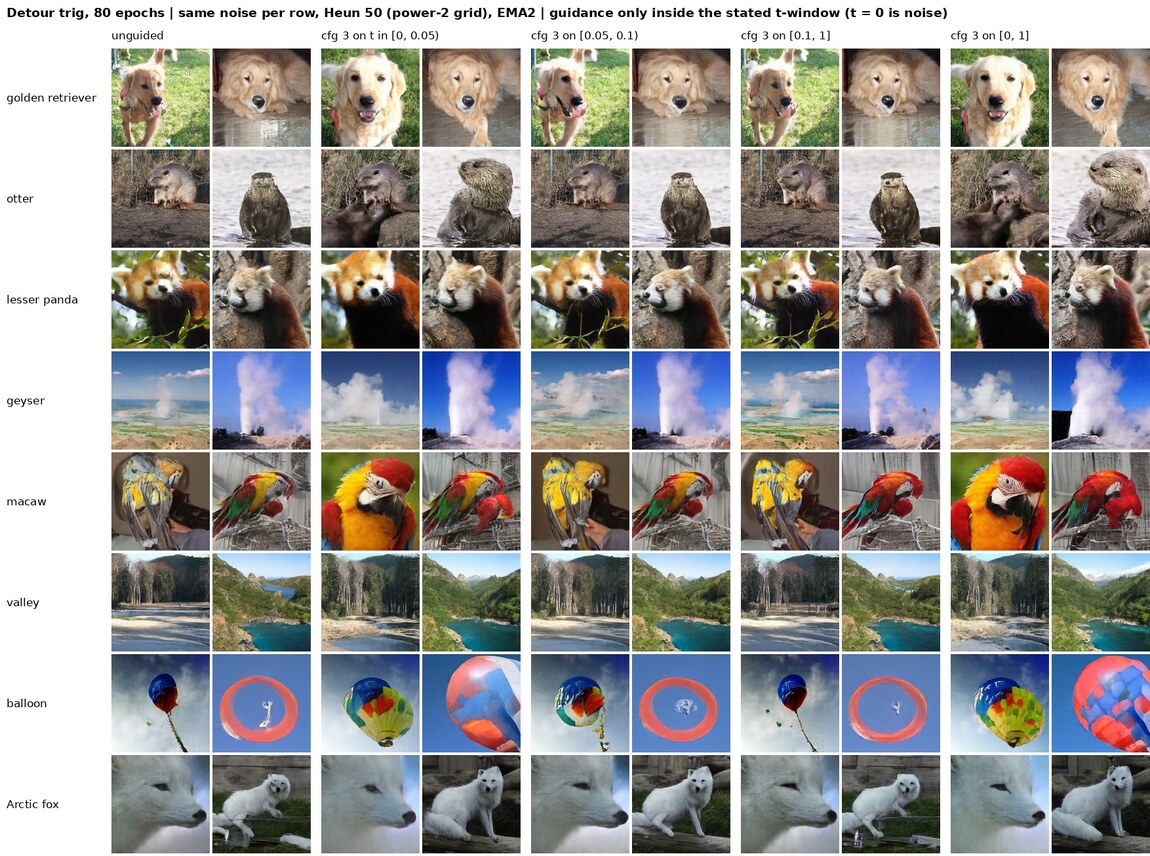}
\caption{\textbf{CFG restricted to different intervals.}
Columns: no guidance, $[0,0.05)$, $[0.05,0.1)$,
$[0.1,1]$, and $[0,1]$.}
\label{fig:supp_window}
\end{figure}

Applying CFG only within $[0,0.05)$ corrects 60\%
of unguided classification failures, compared with
76\% for full guidance and 32\% for guidance over
$[0.1,1]$ (\cref{tab:supp_window,fig:supp_window}).
The code gap also drops rapidly from 0.95 at
$t=0.02$ to 0.30 at $t=0.068$.
Together, these results suggest that class identity
is largely determined early in the trajectory.
As class information becomes encoded in the noisy
state, the code prediction becomes less dependent
on explicit class conditioning.

\subsection{Training-time sampling}
\label{sec:supp_time}

\begin{figure}[tp]
\centering
\includegraphics[width=\linewidth]{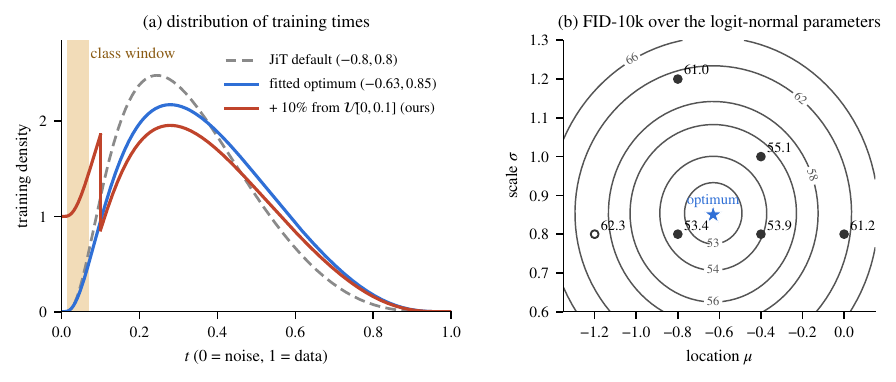}
\caption{\textbf{Training-time distribution.}
(a) JiT default, fitted logit-normal, and our sampler
with 10\% interval sampling.
(b) Quadratic FID-10k fit over logit-normal parameters,
with measured runs.}
\label{fig:supp_tsampler}
\end{figure}

\paragraph{Logit-normal distribution.}
We tune the logit-normal parameters $(\mu,\sigma)$
rather than adopting the JiT default $(-0.8,0.8)$.
A quadratic fit over five training runs identifies
$(-0.63,0.85)$, corresponding to a slightly later
and wider distribution (\cref{fig:supp_tsampler}b).
Moving the distribution substantially toward later
times or increasing its spread degrades performance.
With the norm bump and learning-rate warmup,
the fitted distribution improves FID-10k from
49.1 to 47.9 (\cref{tab:supp_tsampler}).

\begin{table}[htbp]
\centering
\caption{\textbf{Training-time sampling.}
(JiT-B, 20 epochs, unguided FID-10k), evaluated
using Heun-50 with a uniform time grid.}
\label{tab:supp_tsampler}
\small
\setlength{\tabcolsep}{4pt}
\begin{adjustbox}{max width=\linewidth}
\begin{tabular}{@{}lcc@{}}
\toprule
Variant & $(\mu,\sigma)$ & FID $\downarrow$ \\
\midrule
JiT default & $(-0.8,0.8)$ & 53.4 \\
Later & $(-0.4,0.8)$ & 53.9 \\
Much later & $(0.0,0.8)$ & 61.2 \\
Wider & $(-0.8,1.2)$ & 61.0 \\
Later and wider & $(-0.4,1.0)$ & 55.1 \\
\midrule
\multicolumn{3}{l}{\textit{With norm bump + lr warmup}} \\
JiT default & $(-0.8,0.8)$ & 49.1 \\
Fitted setting & $(-0.63,0.85)$ & 47.9 \\
+ Interval sampling & $(-0.63,0.85)$ & \textbf{45.5} \\
\bottomrule
\end{tabular}
\end{adjustbox}
\end{table}

\paragraph{Interval sampling near noise.}
Despite the importance of early semantic decisions,
the fitted logit-normal distribution places only
0.28\% of training times below $t=0.048$
(\cref{fig:supp_tsampler}a).
To improve training coverage near noise, we
introduce interval sampling:
\begin{equation}
t\sim
\begin{cases}
\mathcal U[0,0.1], & \text{with probability }0.1,\\
\mathrm{LogitNormal}(-0.63,0.85), & \text{otherwise}.
\end{cases}
\end{equation}

Interval sampling improves FID-10k from 47.9 to
45.5 at 20 epochs and FID-50k from 24.9 to 21.4
at 80 epochs. Increasing the sampling fraction
to 20\% or widening the interval to $[0,0.2]$
provides no further improvement.

\paragraph{Effect on different trajectory intervals.}
To examine where generation errors arise and whether
interval sampling addresses the early trajectory,
we place 10k real training images on their detour
paths at time $t_0$, then integrate to $t=1$ without
guidance using Heun-50 on a power-2 grid.
We measure the FID difference between two starting
times:
\begin{equation}
\Delta\mathrm{FID}_{[a,b]}
=\mathrm{FID}(a)-\mathrm{FID}(b).
\end{equation}
Starting at $t_0=0$ corresponds to ordinary generation,
while $t_0=1$ recovers real images with an FID of 3.46.
This comparison measures how generation quality
changes when different portions of the learned
trajectory are bypassed.

\begin{table}[htbp]
\centering
\caption{\textbf{FID differences across trajectory
intervals.} (JiT-B, unguided).
Interval differences sum to the total FID minus
the real-image FID of 3.46.
The final two rows differ only in interval sampling.}
\label{tab:stage}
\small
\setlength{\tabcolsep}{3.5pt}
\begin{adjustbox}{max width=\linewidth}
\begin{tabular}{@{}lccccc@{}}
\toprule
Model & $[0,.05]$ & $[.05,.3]$ & $[.3,.7]$ & $[.7,1]$
& FID $\downarrow$ \\
\midrule
Trig, epoch 20 & 6.6 & 17.0 & 18.0 & 7.4 & 52.4 \\
Trig, epoch 80 & 6.4 & 7.0 & 6.6 & 3.9 & 27.3 \\
\midrule
Full recipe, no interval, ep.\ 20
    & 9.6 & 17.7 & 11.7 & 4.1 & 46.6 \\
+ Interval sampling, ep.\ 20
    & \textbf{4.0} & 18.1 & 14.5 & 4.7 & 44.8 \\
\bottomrule
\end{tabular}
\end{adjustbox}
\end{table}

While longer training substantially improves the later
trajectory, the FID difference associated with
$[0,0.05]$ remains nearly unchanged between epochs
20 and 80. Under the matched training recipe,
interval sampling reduces this difference from
9.6 to 4.0, while the overall FID improves more
modestly (\cref{tab:stage}).
These results suggest that the standard time
distribution underrepresents the early trajectory
and that interval sampling directly addresses
this limitation.

% =====================================================================

\section{Code \& Head Design \& Objective}
\label{sec:supp_code}

\paragraph{Multi-layer code.}
Incorporating intermediate DINOv2 features
(\cref{sec:k4}) improves FID-10k from 45.5 to 38.8
at 20 epochs, with nearly unchanged IS
(39.8 vs.\ 40.5).
The improvement persists at 80 epochs, reducing
FID-50k from 20.4 to 15.8.
These results suggest that intermediate features
provide complementary spatial and object-part
information.

\paragraph{Code prediction head.}
We compare three architectures for predicting
the semantic code: a shared output layer,
separate output layers within the wide head,
and direct prediction from the transformer trunk.

\begin{table}[htbp]
\centering
\caption{\textbf{Code prediction head}
(JiT-B, interval sampling, 20 epochs,
unguided FID-10k, without multi-layer codes).}
\label{tab:supp_readout}
\small
\setlength{\tabcolsep}{3.5pt}
\begin{adjustbox}{max width=\linewidth}
\begin{tabular}{@{}lcc@{}}
\toprule
Code readout & Params & FID (IS) \\
\midrule
Shared final layer of wide head
    & 296.3M & 45.5 (39.8) \\
Separate final layer of wide head
    & 304.7M & \textbf{43.9 (42.2)} \\
Final layer on trunk output
    & 296.5M & 56.9 (30.1) \\
\bottomrule
\end{tabular}
\end{adjustbox}
\end{table}

Predicting the code directly from the trunk
substantially worsens FID (\cref{tab:supp_readout}),
suggesting that the wide head is important for
code prediction. Separate output layers provide
a small improvement at 20 epochs but no sustained
advantage with longer training.
We therefore retain the shared output layer
for efficiency.

\paragraph{Coupled versus decoupled supervision.}
Detour supervises both pixel and code predictions
through a unified velocity loss.
To examine the importance of this coupling,
we consider a generalized objective:
\begin{equation}
\begin{aligned}
\mathcal{L}_{\lambda,\gamma}
={}&w_x^2\|e_x\|^2
+\lambda w_c^2\|e_c\|^2\\
&+2\gamma w_xw_c
\langle e_x,e_c\rangle,
\end{aligned}
\label{eq:supp_loss_split}
\end{equation}
where $e_x=x-\hat{x}$,
$e_c=\Phi(c-\hat{c})$, and
\begin{equation}
\begin{aligned}
w_x &= \frac{\alpha_t}{1-t}+t\alpha'_t,\\
w_c &= \frac{\beta_t}{1-t}+t\beta'_t.
\end{aligned}
\end{equation}
Setting $\lambda=\gamma=1$ recovers our velocity
loss, while $\lambda=1,\gamma=0$ removes the
cross term and supervises the two predictions
independently.

\begin{table}[t]
\centering
\caption{\textbf{Coupled versus decoupled
supervision.} JiT-B, ImageNet
$256{\times}256$, 20 epochs, unguided FID-10k.}
\label{tab:supp_loss_split}
\small
\begin{tabular}{lcc}
\toprule
Objective & Sampling grid & FID-10k $\downarrow$\\
\midrule
Coupled ($\lambda=\gamma=1$)
    & Uniform & \textbf{45.5}\\
Decoupled ($\lambda=1,\gamma=0$)
    & Uniform & 51.7\\
Decoupled ($\lambda=1,\gamma=0$)
    & Power-2 & 51.0\\
\bottomrule
\end{tabular}
\end{table}

Removing the cross term worsens FID under both
sampling grids (\cref{tab:supp_loss_split}).
Although decoupled supervision identifies the
individual posterior means at the population
optimum when their corresponding weights are
nonzero, it yields worse generative performance
in practice.
These results support directly optimizing the
combined velocity error rather than independently
supervising pixel and code reconstruction.

% =====================================================================
\input{sec/X_codedim}

\section{Solver and time grid}
\label{sec:supp_solver}

We integrate the sampling ODE on the grid $t_i=(i/N)^p$, where $p>1$ (similar to the step schedule of EDM~\cite{karras2022elucidating}) concentrates steps near noise, when class identity is determined.

\begin{table}[htbp]
\centering
\caption{\textbf{Solver and time grid} (unguided FID-10k, epoch 80).}
\label{tab:supp_solver}
\small
\setlength{\tabcolsep}{4pt}
\begin{tabular}{@{}lcc@{}}
\toprule
Sampler & JiT-B & JiT-XL/16 \\
\midrule
Heun 50, $p{=}2$ & 18.42 & 8.73 \\
Euler 100, $p{=}2$ & 18.10 & -- \\
Euler 100, $p{=}3$ & 17.46 & 8.34 \\
Euler 100, $p{=}4$ & \textbf{16.85} & \textbf{8.06} \\
Euler 100, $p{=}5$ & 18.09 & 9.88 \\
Euler 100, $p{=}6$ & 18.57 & 10.38 \\
\bottomrule
\end{tabular}
\end{table}

Euler with 100 steps and $p=4$ achieves the lowest FID at both model sizes (\cref{tab:supp_solver}). Neither higher-order integration nor additional steps improve FID: on JiT-B, Euler with 50, 150, or 250 steps at $p=3$ performs worse than 100 steps. 

At epoch 200, switching from Heun 50 ($p=2$) to Euler 100 ($p=4$) improves JiT-XL/16 non-cfg FID-50k from 4.13 to 3.69 at the same number of function evaluations. A stochastic SiT-style sampler~\cite{ma2024sit} improves unguided JiT-B FID-50k only slightly (14.27 to 13.79), but worsens guided JiT-XL FID-50k (2.27 to 2.32). We therefore use Euler 100 with $p=4$ throughout.

\begin{figure*}[t]
\centering
\includegraphics[width=\linewidth]{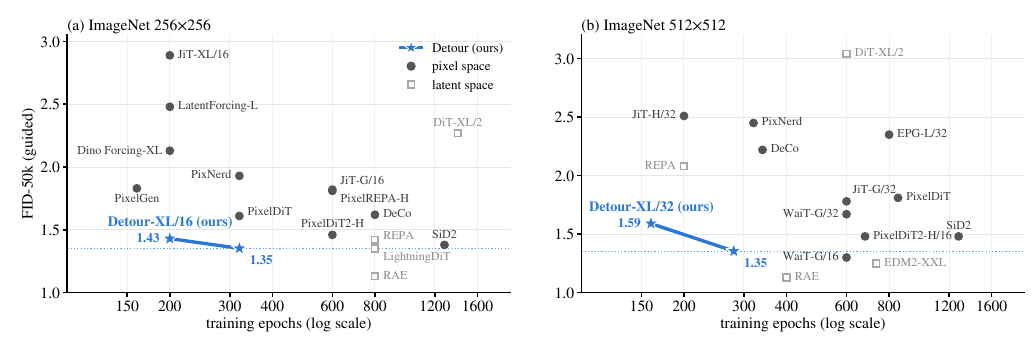}
\caption{\textbf{Guided FID-50k against training epochs} on ImageNet, from \cref{tab:system}. Blue: \method (CFG + AutoGuidance) at 200 and 320 epochs ($256^2$) and at 160 and 280 epochs ($512^2$); dark: pixel space; gray: latent space; one point per model, at its best reported FID. ADM-G (4.59 and 7.72) lies outside the axes.}
\label{fig:imagenet_conv}
\end{figure*}

\paragraph{Convergence along training.}
\Cref{tab:supp_train} reports unguided FID-10k for the XL models at different training epochs.
\Cref{fig:imagenet_conv} places the guided FID-50k of \method against the training length of the models in \cref{tab:system}. Continuing training to 320 epochs at $256^2$ and 280 epochs at $512^2$ lowers FID-50k to 1.35 at both resolutions, below every pixel-space model at $256^2$, while precision stays flat and recall rises ($0.65\to0.67$ and $0.65\to0.68$).

\begin{figure*}[t]
\centering
\includegraphics[width=0.7\textwidth]{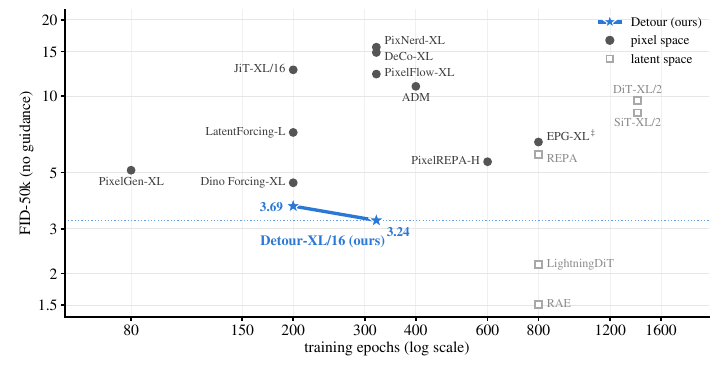}
\caption{\textbf{Unguided FID-50k against training epochs} on ImageNet $256^2$, from \cref{tab:unguided}. Blue: \method-XL/16 at 200 and 320 epochs; dark: pixel space; gray: latent space; one point per model. $^\ddagger$: epochs exclude 480 pre-training epochs.}
\label{fig:unguided_conv}
\end{figure*}

Without guidance, \cref{fig:unguided_conv} and \cref{tab:unguided} compare \method-XL/16 with published models on ImageNet $256^2$, all in FID-50k. Plain pixel flow on the same backbone family, JiT-XL/16, reaches 12.71 after 200 epochs~\cite{ghosh2026dino}; in the same 200 epochs \method-XL/16 reaches 3.69, the lowest among the pixel-space models trained for at most 200 epochs, ahead of Dino Forcing-XL (4.56) and PixelGen-XL (5.11). At 320 epochs it reaches 3.24, the lowest unguided FID among the pixel-space models in \cref{tab:unguided}.

\begin{table}[t]
\centering
\caption{\textbf{Class-conditional ImageNet $256{\times}256$ without guidance} (FID-50k). Gray: latent space. Baseline numbers as reported in~\cite{ghosh2026dino,baade2026latent,ma2026pixelgen,lei2025advancing,tan2026routerepa}. $^\ddagger$: epochs exclude 480 pre-training epochs.}
\label{tab:unguided}
\small
\setlength{\tabcolsep}{4pt}
\begin{adjustbox}{max width=\linewidth}
\begin{tabular}{@{}lrrcc@{}}
\toprule
Method & Params & Ep. & FID & IS \\
\midrule
\color{gray}DiT-XL/2 \cite{peebles2023scalable} & \color{gray}675M & \color{gray}1400 & \color{gray}9.62 & \color{gray}121.5 \\
\color{gray}SiT-XL/2 \cite{ma2024sit} & \color{gray}675M & \color{gray}1400 & \color{gray}8.61 & \color{gray}131.7 \\
\color{gray}REPA \cite{yu2024representation} & \color{gray}675M & \color{gray}800 & \color{gray}5.90 & \color{gray}157.8 \\
\color{gray}LightningDiT \cite{yao2025reconstruction} & \color{gray}675M & \color{gray}800 & \color{gray}2.17 & \color{gray}-- \\
\color{gray}RAE \cite{zheng2026diffusion} & \color{gray}839M & \color{gray}800 & \color{gray}1.51 & \color{gray}-- \\
\midrule
ADM \cite{dhariwal2021diffusion} & 554M & 400 & 10.94 & -- \\
PixelFlow-XL \cite{chen2025pixelflow} & 677M & 320 & 12.23 & 103.3 \\
PixNerd-XL \cite{wang2026pixnerd} & 700M & 320 & 15.61 & 88.9 \\
DeCo-XL/16 \cite{ma2026deco} & 682M & 320 & 14.88 & 88.2 \\
EPG-XL/16$^\ddagger$ \cite{lei2025advancing} & 583M & 800 & 6.60 & 141.1 \\
JiT-H/16 \cite{li2026back} & 953M & 600 & 7.20 & -- \\
PixelREPA-H/16 \cite{shin2026representation} & 953M & 600 & 5.52 & -- \\
\textbf{\method-XL/16} & 849M & 320 & \textbf{3.24} & 191.7 \\
\midrule
\multicolumn{5}{@{}l}{\emph{$\le$200 epochs}} \\
JiT-L/16 \cite{li2026back} & 459M & 200 & 17.19 & 85.4 \\
JiT-XL/16 \cite{li2026back} & 676M & 200 & 12.71 & 101.7 \\
LatentForcing-L \cite{baade2026latent} & 465M & 200 & 7.20 & -- \\
PixelGen-XL/16 \cite{ma2026pixelgen} & 676M & 80 & 5.11 & 159.2 \\
Dino Forcing-L \cite{ghosh2026dino} & 468M & 200 & 5.53 & 163.6 \\
Dino Forcing-XL \cite{ghosh2026dino} & 686M & 200 & 4.56 & 179.6 \\
\textbf{\method-XL/16} & 849M & 200 & \textbf{3.69} & 182.3 \\
\bottomrule
\end{tabular}
\end{adjustbox}
\end{table}

\begin{table}[htbp]
\centering
\caption{\textbf{FID along training} (unguided FID-10k, Euler 100, $p=4$; $^\dagger$Heun 50, $p=2$). Final FID-50k: 3.69 at $256^2$ and 4.19 at $512^2$.}
\label{tab:supp_train}
\small
\setlength{\tabcolsep}{2.5pt}
\begin{adjustbox}{max width=\linewidth}
\begin{tabular}{@{}lcccccccccc@{}}
\toprule
Epoch & 20 & 40 & 60 & 80 & 100 & 120 & 140 & 160 & 180 & 200 \\
\midrule
XL/16, $256^2$ & 15.69 & 10.39 & 8.89 & 8.06 & 7.49 & 7.33 & 6.76 & 6.65 & 6.58 & 6.34 \\
XL/32, $512^2$ & 21.36$^\dagger$ & 12.71$^\dagger$ & 10.16$^\dagger$ & 8.19 & 7.54 & 7.41 & 7.64 & 6.90 & -- & -- \\
\bottomrule
\end{tabular}
\end{adjustbox}
\end{table}

\section{Guidance: additional results}
\label{sec:supp_guidance}

We analyze the complementary behavior of CFG and
AutoGuidance and evaluate the robustness of their
combination.

\begin{figure*}[tp]
\centering
\includegraphics[width=\linewidth]{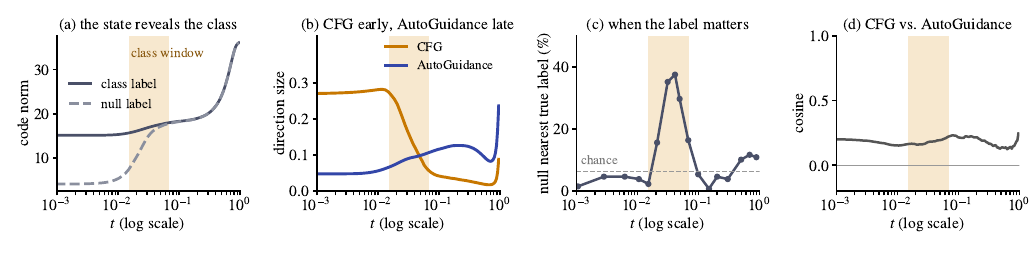}
\caption{\textbf{Guidance dynamics}
(\method-XL/16, epoch 200; shaded: class window).
(a) Conditional and unconditional code norms.
(b) RMS magnitude of the CFG and AutoGuidance
directions.
(c) Accuracy of identifying the true class among
16 candidates by matching conditional and
unconditional velocities.
(d) Cosine similarity between the two guidance
directions.}
\label{fig:supp_guidance_dirs}
\end{figure*}

\paragraph{Guidance dynamics.}
For $f\in\{x,c\}$, let
$\Delta f=\mathbb E[f\mid z_t,y]-\mathbb E[f\mid z_t]$.
As derived in the appendix,
\begin{equation}
\|\Delta f\|
\leq \bigl(1-p(y\mid z_t)\bigr)
\max_{y'}\|\mathbb E[f\mid z_t,y]
-\mathbb E[f\mid z_t,y']\|.
\label{eq:cfg_vanish}
\end{equation}
Thus, the optimal CFG direction diminishes as the
noisy state reveals the class.
We examine this behavior using 64 sampling trajectories
from \method-XL/16 at epoch 200, evaluating the
conditional, unconditional, and epoch-20 models at
the same noisy state (\cref{fig:supp_guidance_dirs}).

\textbf{(a) Code predictions.}
Near noise, the conditional-unconditional code
difference is nearly aligned with the conditional
prediction (cosine similarity 0.96), suggesting
that CFG primarily amplifies the class prototype.
Conditional and unconditional code norms become
similar around $t=0.05$.

\textbf{(b) Guidance magnitude.}
The RMS magnitude of CFG decreases from 0.27 near
noise to 0.043 at $t=0.1$, with similar behavior
across epochs 100, 160, and 200.
In contrast, AutoGuidance remains active throughout
the trajectory, increasing from 0.048 near noise
to approximately 0.12 after $t=0.1$.

\textbf{(c) Class information.}
Along 128 unguided trajectories, we identify which
of 16 candidate class labels (the true label and
15 random alternatives) produces the velocity
closest to the unconditional prediction.
The accuracy rises from chance to 38\% between
$t=0.02$ and $0.07$, indicating that class information
becomes accessible to the unconditional model.
It returns to chance after $t=0.1$, consistent
with the diminishing influence of class conditioning.

\textbf{(d) Guidance directions.}
CFG and AutoGuidance are nearly orthogonal throughout
sampling, with cosine similarities between 0.12
and 0.25 (mean 0.18).

Together, these observations suggest that CFG
primarily influences early semantic decisions,
whereas AutoGuidance provides complementary
corrections throughout the trajectory.

\begin{figure*}[tp]
\centering
\includegraphics[width=\linewidth]{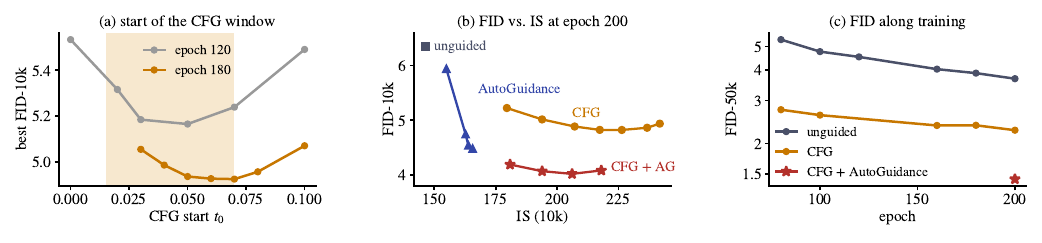}
\caption{\textbf{Guidance sweeps}
(\method-XL/16).
(a) FID-10k versus CFG start time.
(b) FID--IS trade-off at epoch 200.
(c) FID-50k across training.}
\label{fig:supp_sweeps}
\end{figure*}

\paragraph{Guidance sweeps.}
AutoGuidance improves FID with relatively little
change in IS, whereas CFG provides a larger
increase in IS.
Their combination achieves a better FID--IS
trade-off (\cref{fig:supp_sweeps}).
The preferred CFG start remains near $t_0=0.05$
across training.
CFG alone saturates around FID-50k 2.3 after
epoch 160 despite continued unguided improvement,
while the combined sampler reaches 1.43 at
epoch 200.

\begin{figure*}[tp]
\centering
\includegraphics[width=\linewidth]{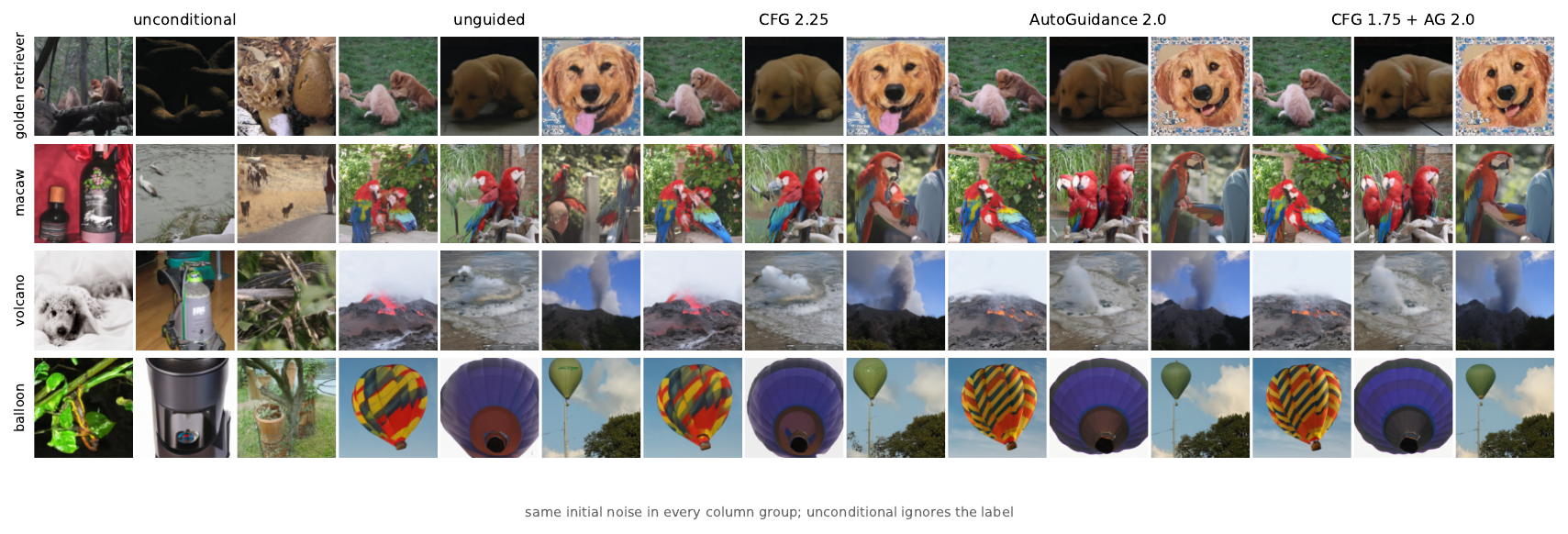}
\caption{\textbf{Samples under different guidance
settings.} Each column group uses the same
initial noise.}
\label{fig:supp_guided_samples}
\end{figure*}

CFG also reduces within-class diversity, measured
by mean pairwise DINOv2 cosine distance, from
0.466 to 0.403.
AutoGuidance largely preserves diversity (0.479),
while their combination achieves 0.433
(\cref{fig:supp_guided_samples}).
These results suggest that CFG trades diversity
for class fidelity, whereas AutoGuidance improves
sample quality with less reduction in diversity.

\paragraph{Sampler ablations.}
\Cref{tab:sampler} evaluates individual components
of our final sampler.
Although CFG has its strongest influence near noise
(\cref{sec:supp_stage}), applying it from $t=0$
substantially worsens FID.
This is consistent with early guidance amplifying
class prototypes at the expense of within-class
diversity.
Starting CFG at $t=0.05$ provides a better
FID--IS trade-off.

The epoch-20 EMA checkpoint is the most effective
AutoGuidance model, with no additional training.
Stronger checkpoints provide less effective guidance.
Moderate changes to guidance scales, time-grid
exponent, and step count have relatively little
impact.

\begin{table}[htbp]
\centering
\caption{\textbf{Sampler ablations}
(\method-XL/16, epoch 200, FID-10k).
One change at a time from the final sampler (4.02).
Guide-checkpoint comparisons use AutoGuidance alone.}
\label{tab:sampler}
\small
\setlength{\tabcolsep}{3pt}
\begin{adjustbox}{max width=\linewidth}
\begin{tabular}{@{}llp{4.55cm}@{}}
\toprule
Component & Default & Alternative $\to$ FID-10k \\
\midrule
Guidance & CFG+AG
    & CFG: 4.83; AG: 4.48; none: 6.34 \\
CFG start $t_0$ & 0.05
    & 0: 7.23; 0.02: 4.77; 0.1: 4.36 \\
CFG end $t_1$ & 0.9
    & 0.5: 4.30; 0.75: 4.10; 1.0: 4.22 \\
Guide epoch & 20
    & 40: 5.01; 120: 5.82; 160: 6.13 (20: 4.48) \\
AG scale $a$ & 2.0
    & 1.75: 4.09; 2.5: 4.29 \\
CFG scale $s$ & 1.75
    & 1.25: 4.19; 2.0: 4.09 \\
Grid exponent $p$ & 4
    & 3.5: 4.13; 4.5: 4.06 \\
Steps & 100
    & 125: 4.17 \\
\bottomrule
\end{tabular}
\end{adjustbox}
\end{table}

Alternative CFG designs, including
APG~\cite{sadat2024eliminating}, separate pixel/code
scales, and norm rescaling, provide no substantial
improvement over standard CFG (FID-10k 4.81--4.91
versus 4.83).
Stochastic sampling similarly provides no improvement
(FID-50k 2.32 versus 2.27 with CFG).
Using raw rather than EMA weights for the weak
guide worsens FID-10k from 4.48 to 4.66.
A separately trained JiT-B guide achieves 4.11
with CFG, compared with 4.02 using the epoch-20
checkpoint, while requiring additional training.

We additionally evaluate an internal guide from
block 9, achieving FID-50k of 2.33 alone and
1.96 with CFG.
Although less effective than AutoGuidance
(1.81 and 1.43, respectively), this approach
avoids an additional weak-model evaluation.

\paragraph{Results at $512{\times}512$.}
The same ordering holds for \method-XL/32:
AutoGuidance outperforms CFG alone, and their
combination achieves the best FID
(\cref{tab:supp512}).
The preferred CFG scale increases to 2.0 when
combined with AutoGuidance and 4.0 when used alone.
The combined sampler achieves FID-50k 1.59.

\begin{table}[htbp]
\centering
\caption{\textbf{Guidance at $512{\times}512$}
(\method-XL/32, epoch 160).
FID-50k: CFG 4.0 alone 2.54;
CFG 2.0 + AG 1.59;
CFG 2.5 + AG 1.63.}
\label{tab:supp512}
\small
\setlength{\tabcolsep}{4pt}
\begin{adjustbox}{max width=\linewidth}
\begin{tabular}{@{}lcc@{}}
\toprule
Setting & FID-10k $\downarrow$ & IS $\uparrow$ \\
\midrule
Unguided & 6.90 & 150 \\
CFG 2.0 / 3.0 / 4.0, $[0.05,0.9]$
    & 5.46 / 5.13 / 5.06 & 208 / 238 / 254 \\
CFG 4.0, start 0.03 / 0.1
    & 5.92 / 5.56 & 292 / 207 \\
AG (ep.\ 40), 2.0
    & 5.58 & 163 \\
AG (ep.\ 20), 1.5 / 2.0 / 2.5
    & 5.23 / 4.62 / 4.70 & 170 / 174 / 172 \\
CFG 2.0 / 2.5 / 3.0 + AG 2.0
    & \textbf{4.23} / 4.24 / 4.29 & 218 / 235 / 248 \\
\bottomrule
\end{tabular}
\end{adjustbox}
\end{table}

\section{Code feedback}
\label{sec:supp_feedback}

\begin{figure*}[tp]
\centering
\includegraphics[width=\linewidth]{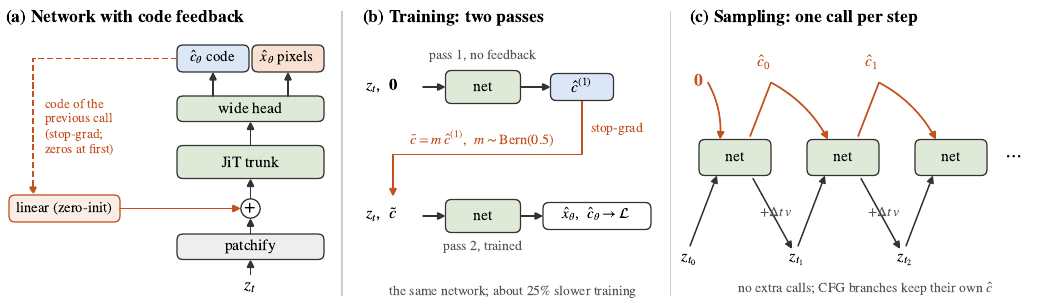}
\caption{\textbf{Code feedback.}
(a) Previous code predictions are projected into
the trunk input.
(b) Training uses a gradient-free first pass to
provide feedback for half the batch.
(c) Sampling reuses the previous prediction without
additional network evaluations.}
\label{fig:supp_feedback_arch}
\end{figure*}

\paragraph{Method.}
Following self-conditioning~\cite{ghosh2026dino,chen2023analogbitsgeneratingdiscrete},
we reuse the semantic code predicted at the previous
sampling step. A zero-initialized linear layer projects
the code to the trunk width and adds it to the input
patch tokens (\cref{fig:supp_feedback_arch}).
The initial feedback is zero.

During training, a gradient-free first pass provides
feedback for half of each batch, increasing training
time by approximately 25\%.
Sampling requires no additional network evaluations.
Under CFG, the conditional and unconditional
branches maintain separate feedback codes.

\paragraph{Results on JiT-B.}
Code feedback substantially improves convergence,
reducing FID-10k from 38.2 to 18.6 at 20 epochs
and from 18.4 to 7.98 at 80 epochs
(\cref{tab:supp_feedback}).
The detour path also outperforms plain pixel flow
under the same feedback mechanism, suggesting
that its benefits extend beyond self-conditioning.

\begin{table}[htbp]
\centering
\caption{\textbf{Code feedback on JiT-B}
(unguided FID-10k; Heun 50, $p=2$).}
\label{tab:supp_feedback}
\small
\setlength{\tabcolsep}{4pt}
\begin{adjustbox}{max width=\linewidth}
\begin{tabular}{@{}lcccc@{}}
\toprule
Model & 20 ep. & 40 ep. & 60 ep. & 80 ep. \\
\midrule
Pixel flow + feedback (Dino Forcing-style)
    & 47.9 & 24.7 & -- & -- \\
\method + feedback (last-layer code)
    & 21.6 & 13.2 & -- & -- \\
\midrule
\method, multi-layer code
    & 38.2 & 24.6 & 20.5 & 18.4 \\
\method, multi-layer code + feedback
    & \textbf{18.6} & \textbf{11.2}
    & \textbf{9.1} & \textbf{7.98} \\
\bottomrule
\end{tabular}
\end{adjustbox}
\end{table}

\begin{figure}[htbp]
\centering
\includegraphics[width=\linewidth]{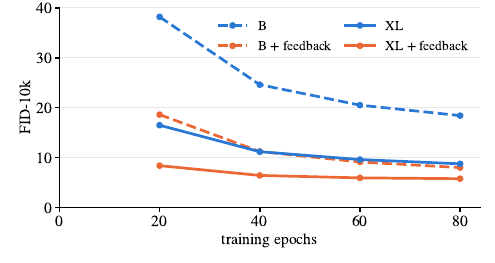}
\caption{\textbf{Code feedback accelerates unguided
convergence.} FID-10k during training
(Heun 50, $p=2$).}
\label{fig:supp_feedback_conv}
\end{figure}

\paragraph{Scaling to XL.}
Feedback provides similar gains at XL scale,
approximately halving FID-10k at epoch 20
(16.5 to 8.36) and surpassing the 200-epoch
model without feedback by epoch 40
(6.40 vs.\ 6.86; \cref{fig:supp_feedback_conv}).
At epoch 80, unguided FID-50k improves from
5.34 to 3.18.

However, these improvements do not translate into
better guided FID (\cref{tab:supp_feedback_xl}).
The feedback model achieves considerably higher
unguided IS (234 vs.\ 157), with little further
improvement from CFG.
With AutoGuidance, it reaches FID-50k 2.09,
compared with 1.87 for the model without feedback
using combined guidance.

One possible explanation is that feedback produces
a stronger epoch-20 checkpoint, making it less
effective as a weak AutoGuidance model
(\cref{sec:supp_guidance}).
Exploring earlier checkpoints may improve guidance
performance.
We therefore leave feedback out of our main XL
models despite its substantial unguided
convergence gains.

\begin{table}[htbp]
\centering
\caption{\textbf{Code feedback at XL scale}
(\method-XL/16, epoch 80; Euler 100, $p=4$).
Guidance on $[0.05,0.9]$; AutoGuidance uses each
model's epoch-20 checkpoint.
IS is measured on 50k samples.}
\label{tab:supp_feedback_xl}
\small
\setlength{\tabcolsep}{4pt}
\begin{adjustbox}{max width=\linewidth}
\begin{tabular}{@{}llccc@{}}
\toprule
Model & Guidance & FID-10k & FID-50k & IS \\
\midrule
\method & None & 8.06 & 5.34 & 157 \\
 & CFG 2.25 & -- & 2.75 & 269 \\
 & CFG 1.75 + AG 2.0
    & \textbf{4.42} & \textbf{1.87} & 248 \\
\midrule
+ feedback & None & 5.70 & 3.18 & 234 \\
 & CFG 1.25 & 5.66 & 3.12 & 239 \\
 & AG 2.5 & 4.60 & 2.09 & 250 \\
 & CFG 1.125 + AG 2.5 & 4.63 & 2.12 & 260 \\
\bottomrule
\end{tabular}
\end{adjustbox}
\end{table}

\begin{figure*}[tp]
\centering
\includegraphics[width=\linewidth]{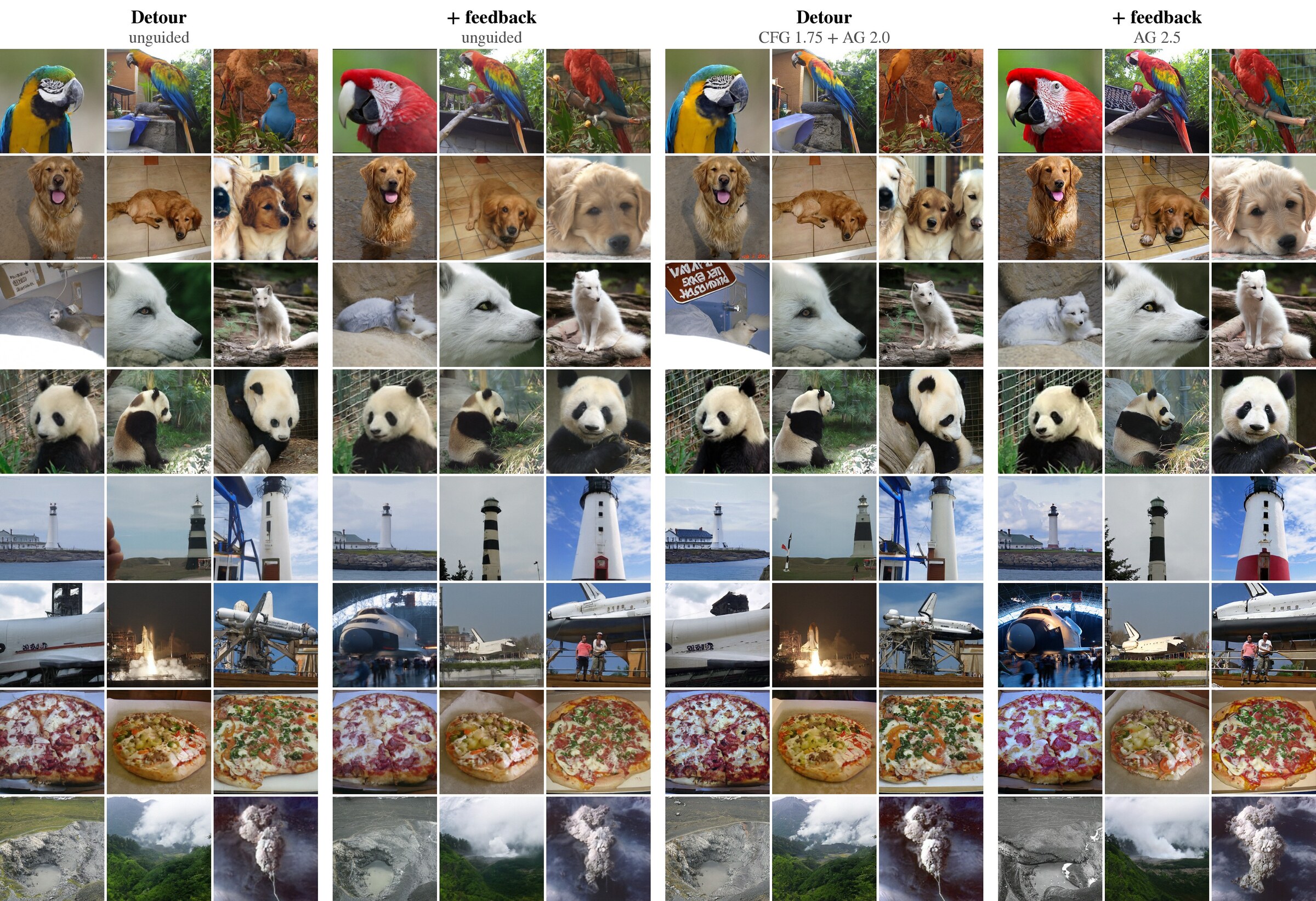}
\caption{\textbf{Qualitative comparison with code
feedback} (\method-XL/16, epoch 80).
Uncurated samples using identical noise, with each
model's best guided setting from
\cref{tab:supp_feedback_xl}.}
\label{fig:supp_feedback_qual}
\end{figure*}

% =====================================================================

\section{One-step distillation}
\label{sec:supp_dmd}
\label{sec:distill}

Since \method samples through a pixel-space ODE,
it is compatible with distribution-matching
distillation (DMD)~\cite{yin2024one,yin2024improved}.
The teacher's image and code predictions provide
the score estimate
\begin{equation}
\hat s_\theta(z_t,t)
=-\frac{\hat\varepsilon_\theta}{1-t},
\end{equation}
where $\hat\varepsilon_\theta$ is defined in
\cref{eq:param} and the score derivation is given
in \cref{sec:supp_theory}.

\paragraph{Distillation training.}
We distill the CFG-guided epoch-200
\method-XL/16 teacher (EMA2) into a one-step
generator, initializing both the generator and
fake-score network from the teacher.
Following DMD, we use a regression anchor on
25.6k teacher-generated noise-class-image
triplets, with a pseudo-Huber loss whose weight
decays from 1.0 to 0.1 over 6k steps.

For each generated image $\hat x$, we apply
a random horizontal flip and construct the
noisy state along the detour path:
\begin{equation}
z_t=t\bigl(\alpha_t\hat x+
\beta_t\Phi c(\hat x)\bigr)+(1-t)\varepsilon,
\end{equation}
where $t\sim\mathcal U[0.02,0.98]$.
Unlike standard pixel-space DMD, the generator
gradient propagates through both the pixel and
semantic components of the path, requiring
backpropagation through the frozen DINOv2 encoder.
We perform five fake-score updates using the
detour velocity objective for each generator
update. Training uses 8 GPUs.

\paragraph{Results.}
The distilled model achieves FID-50k of 2.38
with one network evaluation, approaching the
teacher's 2.27, and improves to 2.15 with four
evaluations using the same checkpoint
(\cref{tab:distill}).
These results demonstrate that \method is
compatible with distribution-matching
distillation for few-step pixel generation.

\begin{table}[htbp]
\centering
\caption{\textbf{Few-step generation}
on ImageNet $256{\times}256$.
Our distilled model uses the same checkpoint
for one and four network evaluations.}
\label{tab:distill}
\small
\begin{adjustbox}{max width=\linewidth}
\begin{tabular}{@{}lccc@{}}
\toprule
Method & NFE & FID-50k $\downarrow$ & IS $\uparrow$ \\
\midrule
pMF-H/16~\cite{lu2026one}
    & 1 & 2.22 & 269 \\
Drifting-L/16~\cite{deng2026generative}
    & 1 & 1.61 & 308 \\
\method-XL/16 teacher (CFG guided)
    & $100 \times 2$ & 2.27 & 283 \\
\method-XL/16, distilled
    & 1 & 2.38 & 263 \\
\method-XL/16, distilled
    & 4 & 2.15 & 260 \\
\bottomrule
\end{tabular}
\end{adjustbox}
\end{table}

\begin{table*}[!t]
\centering
\caption{\textbf{Text-to-image sources, filtering, and the stages that draw them.} A tick marks a source the stage samples; a blank cell means the stage does not draw it. \emph{A} is 0--50k at $256$, \emph{B} 50k--130k at $256$, \emph{B512} 130k--180k at $512$, and \emph{C} 180k--200k at $1024$. Within stage B the mix shifts toward real photographs after iteration 75k. Stage C re-filters the sources it keeps, selecting for composition, rendered text, faces, groups of people, and named visual styles.}
\label{tab:supp_t2i_data}
\small
\setlength{\tabcolsep}{5pt}
\begin{adjustbox}{max width=\textwidth}
\begin{tabular}{@{}llcccc@{}}
\toprule
Source & Content and filtering & A & B & B512 & C \\
\midrule
Flux2M~\cite{naeem_khoshnevis_2026} & FLUX.2-dev, generation prompts, 256\,px & \gcheck & \gcheck & & \\
Flux2M at 512 & Same images and prompts at 512\,px & & & \gcheck & \\
JourneyDB~\cite{sun2023journeydb} & Midjourney, recaptioned~\cite{chen2025blip3} & & \gcheck & & \\
Megalith~\cite{BoerBohan2024Megalith10m} & Real Flickr photographs, cleaned captions & & \gcheck & \gcheck & \\
SA-1B~\cite{chen2023internvl,chen2024far} & Real photographs, faces blurred at source & & \gcheck & & \\
FineT2I~\cite{Mao_2026} & Real photographs, held-out split removed & & \gcheck & \gcheck & \gcheck \\
FineT2I, enhanced & Synthetic, dense captions only & & \gcheck & \gcheck & \gcheck \\
FineT2I, original & Synthetic, short prompts only & & \gcheck & \gcheck & \gcheck \\
CommonCatalog~\cite{gokaslan2023commoncanvas} & Real CC-BY photographs, recaptioned, aspect $\leq1.6$ & & & \gcheck & \gcheck \\
Midjourney v6~\cite{brivangl_midjourney_v6_llava} & Synthetic, recaptioned & & & \gcheck & \gcheck \\
FLUX.1-dev aesthetics~\cite{fang2025flux} & Synthetic, aesthetic and detail scores & & & \gcheck & \gcheck \\
\quad Composition & Short captions, composition score $\geq8$ & & & \gcheck & \gcheck \\
TextGround, curated~\cite{Mao_2026} & Rendered text with quoted spans, aspect $\leq1.6$ & & & \gcheck & \gcheck \\
TextGround, web & Web images containing text, aspect $\leq1.6$ & & & \gcheck & \\
\bottomrule
\end{tabular}
\end{adjustbox}
\end{table*}

% =====================================================================

\section{Text-to-image architectures}
\label{sec:supp_t2i_arch}

Both text-to-image architectures (\cref{fig:t2i_arch})
follow the detour path (\cref{fig:path}) and predict
the image and semantic code from the noisy state $z_t$.
Text conditioning uses a frozen Qwen3.5-4B,
concatenating hidden states from layers 7, 15, and 27.

\paragraph{562M model.}
For the convergence study (\cref{sec:supp_t2i_conv}),
we replace JiT's class conditioning with text.
Projected text tokens are prepended to image tokens,
while adaLN receives the time embedding and projected
mean text token.
The network uses 20 transformer blocks of width 1024
on 16-pixel patches, followed by a two-block output
head of width 2048, without a patch bottleneck.
The pixel-flow baseline uses the same architecture
without code prediction.

\paragraph{3.3B model.}
Our larger model follows the two-stage JiT-DDT
architecture~\cite{chopra2026pyramidjit}.
An encoder operating on 32-pixel patches predicts
a low-resolution image at $1/8$ scale, supervised
against the average-pooled image through an
auxiliary loss.
Its tokens condition a decoder operating on
16-pixel patches, which predicts the full-resolution
image and code under the detour velocity objective.
Each stack uses two image-only and two text-only
refiner blocks, followed by 11 joint blocks of
width 2944, with 23 attention heads, gated QK-norm
attention, SwiGLU, and time-conditioned adaLN.
Both stacks share the text tokens.

\paragraph{Resolution transfer and sampling.}
We progressively train at $256$, $512$, and
approximately $1024$ resolution, interpolating
sin-cos and RoPE positions relative to the
$256{\times}256$ grid.
To adjust the signal-to-noise ratio across
resolutions, we use the time shift
\begin{equation}
t'=
\begin{cases}
t/(2-t), & 512{\times}512,\\
t/(4-3t), & 1024{\times}1024.
\end{cases}
\end{equation}
The $512$ and $1024$ stages use 17 aspect-ratio
buckets with ratios up to 2.1.
All results in \cref{tab:t2i} use the same
iteration-200k EMA checkpoint (decay 0.9995),
sampled with Heun-30 on the power-2 grid and
the corresponding time shift.
At $512$, we use CFG scale 14; at $1024$,
we use standard image-and-code CFG with scale 6.

\paragraph{Training data.}
We assemble approximately 31M image--text pairs
from public sources, comprising 18M real photographs
and 13M synthetic images.
When both dense and short captions are available,
we select the dense caption with probability 0.8.
\Cref{tab:supp_t2i_data} lists the data sources,
filtering criteria, and corresponding training
stages.
The packed corpus is released as
\url{https://huggingface.co/datasets/quandao10/t2i-pretrain-mix}.
\begin{table}[h]
\centering
\caption{\textbf{Text-to-image training cost}
on $8\times$H200 through iteration 200k.
Stage C uses 75\% of batches at $1024$ and
25\% at $512$.
The $512$ throughput is estimated from token
count; other rates are measured.}
\label{tab:supp_t2i_cost}
\small
\setlength{\tabcolsep}{4pt}
\begin{adjustbox}{max width=\linewidth}
\begin{tabular}{@{}llrrrr@{}}
\toprule
Stage & Res. & Iters & Batch & Images & GPU-days \\
\midrule
A & 256 & 50k & 1024 & 51.2M & 8 \\
B & 256 & 80k & 1024 & 81.9M & 13 \\
B512 & 512 & 50k & 1024 & 51.2M & 37 \\
C & 1024 / 512 & 20k & 256 & 3.8M / 1.3M & 19 \\
\midrule
\multicolumn{2}{@{}l}{Total}
    & 200k & & 189M & 78 \\
\bottomrule
\end{tabular}
\end{adjustbox}
\end{table}

\paragraph{Training schedule and cost.}
Training begins with 2M synthetic Flux2M images
(stage A), followed by the full corpus at $256$
resolution (stage B), with increased emphasis on
real photographs after iteration 75k.
Stage B512 introduces higher-resolution images
and rendered text.
Stage C drops lower-resolution sources and further
filters the data for prompt following, composition,
rendered text, faces, and visual styles, while
retaining 25\% of batches at $512$ resolution
for replay.

As summarized in \cref{tab:supp_t2i_cost},
the model processes approximately 189M images
throughout training, with substantially fewer
images at $1024$ resolution.
This limited high-resolution exposure may partly
explain the lower benchmark scores at $1024$
compared with $512$ in \cref{tab:t2i}.

% =====================================================================

\section{Text-to-image comparison}
\label{sec:supp_t2i_conv}

\paragraph{Experimental setup.}
We compare \method with plain pixel flow using
matched JiT-T2I architectures (approximately
562M parameters; \cref{sec:supp_t2i_arch}).
Both models are trained on the same Flux2M
subset~\cite{naeem_khoshnevis_2026} at
$256{\times}256$, with identical data order,
batch size 1024, AdamW with learning rate
$2{\times}10^{-4}$, 1k warm-up iterations,
and EMA 0.9996.
Training runs for 30k iterations
(approximately 13.5 epochs).
Both use logit-normal time sampling
$(-0.8,0.8)$, while \method additionally
uses the trigonometric hand-over with a norm bump.

We evaluate every 2.5k iterations on GenEval
(553 prompts) and DPG-Bench (1,065 prompts),
using four samples per prompt, Euler-100
sampling, and CFG 6.

\begin{figure*}[tp]
\centering
\includegraphics[width=\linewidth]
{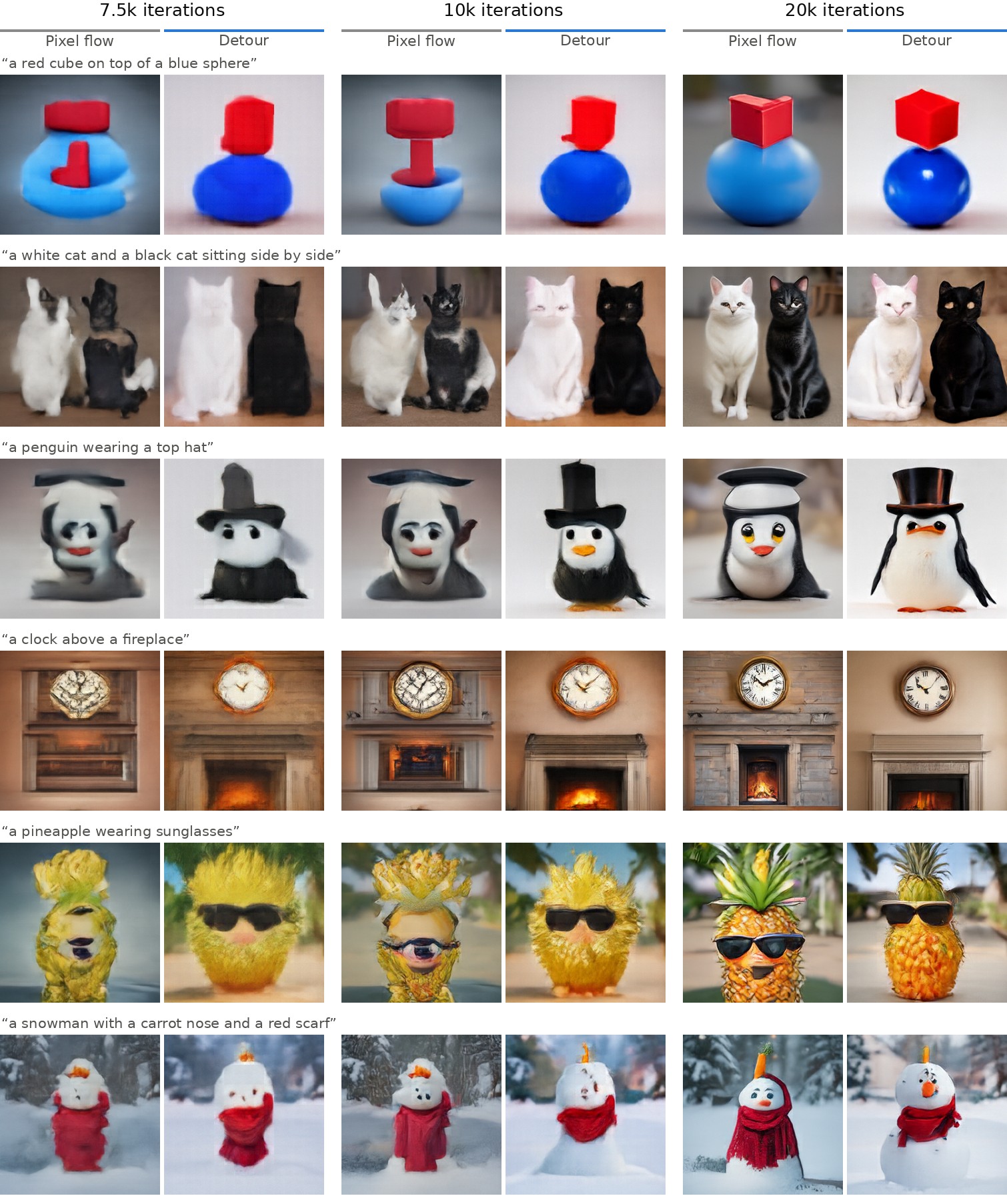}
\caption{\textbf{Text-to-image convergence.}
Samples from pixel flow and \method at
matched training iterations, prompts,
and initial noise.
Euler-100 sampling, CFG 6.}
\label{fig:supp_t2i_samples}
\end{figure*}

\paragraph{Convergence and compositional learning.}
\method consistently outperforms pixel flow
throughout training (\cref{tab:supp_t2i_conv}).
At 7.5k iterations, it doubles GenEval
(0.372 vs.\ 0.186), while at 10k iterations,
it achieves $2.7\times$ higher compositional
accuracy. The advantage persists through
30k iterations.
The qualitative comparison
(\cref{fig:supp_t2i_samples}) further shows
that \method learns object identities and
spatial relationships earlier, while pixel
flow more frequently merges or misplaces objects.

\begin{table}[t]
\centering
\caption{\textbf{Text-to-image convergence}
with matched architectures
($\approx562$M parameters).
Composition averages counting, position,
and attribute binding.
Euler-100 sampling, CFG 6.}
\label{tab:supp_t2i_conv}
\small
\setlength{\tabcolsep}{2.6pt}
\begin{adjustbox}{max width=\linewidth}
\begin{tabular}{@{}l cccc ccc ccc@{}}
\toprule
 & \multicolumn{4}{c}{GenEval}
 & \multicolumn{3}{c}{Composition}
 & \multicolumn{3}{c}{DPG-Bench} \\
\cmidrule(lr){2-5}
\cmidrule(lr){6-8}
\cmidrule(lr){9-11}
 & 7.5k & 10k & 20k & 30k
 & 10k & 20k & 30k
 & 10k & 20k & 30k \\
\midrule
Pixel flow
 & .186 & .296 & .575 & .695
 & .106 & .330 & .505
 & 65.9 & 78.7 & 81.7 \\
\method
 & \textbf{.372} & \textbf{.490}
 & \textbf{.682} & \textbf{.756}
 & \textbf{.281} & \textbf{.501}
 & \textbf{.639}
 & \textbf{72.0} & \textbf{80.8}
 & \textbf{84.3} \\
\bottomrule
\end{tabular}
\end{adjustbox}
\end{table}

\paragraph{High-frequency fidelity.}
We compare \method with RAE-based latent
generation on 5,000 held-out Flux2M images
at $256{\times}256$, using the corresponding
prompts for generation with Euler-100 sampling.
We measure the high-frequency gap between
generated and real images:
\begin{equation}
\mathrm{HF\ gap}
=
\underset{f\in\mathcal{F}_{\mathrm{high}}}
{\operatorname{mean}}
\left|
\log_{10}P_{\mathrm{gen}}(f)
-
\log_{10}P_{\mathrm{real}}(f)
\right|,
\label{eq:hfgap}
\end{equation}
where $P(f)$ denotes the dataset-averaged
radial luminance power spectrum and
$\mathcal{F}_{\mathrm{high}}$ contains the
upper half of spatial frequencies.

As a reconstruction reference, we also
evaluate the RAE decoder on the held-out
images, obtaining 18.2\,dB PSNR and an
HF gap of 0.132.

\begin{table}[t]
\centering
\caption{\textbf{High-frequency gap}
(\cref{eq:hfgap}, lower is better) on
5,000 held-out Flux2M images at
$256{\times}256$.
RAE reconstruction is a fixed reference
measured independently of generation.}
\label{tab:supp_detail}
\small
\begin{tabular}{@{}lccccc@{}}
\toprule
HF gap $\downarrow$
 & 10k & 15k & 20k & 25k & 30k \\
\midrule
\color{gray}RAE reconstruction
 & \color{gray}0.132
 & \color{gray}0.132
 & \color{gray}0.132
 & \color{gray}0.132
 & \color{gray}0.132 \\
\midrule
RAE latent
 & 0.149 & 0.155 & 0.153 & 0.153 & 0.149 \\
\method
 & 0.417 & 0.139 & 0.095 & 0.074
 & \textbf{0.058} \\
\bottomrule
\end{tabular}
\end{table}

\begin{figure}[t]
\centering
\includegraphics[width=\linewidth]
{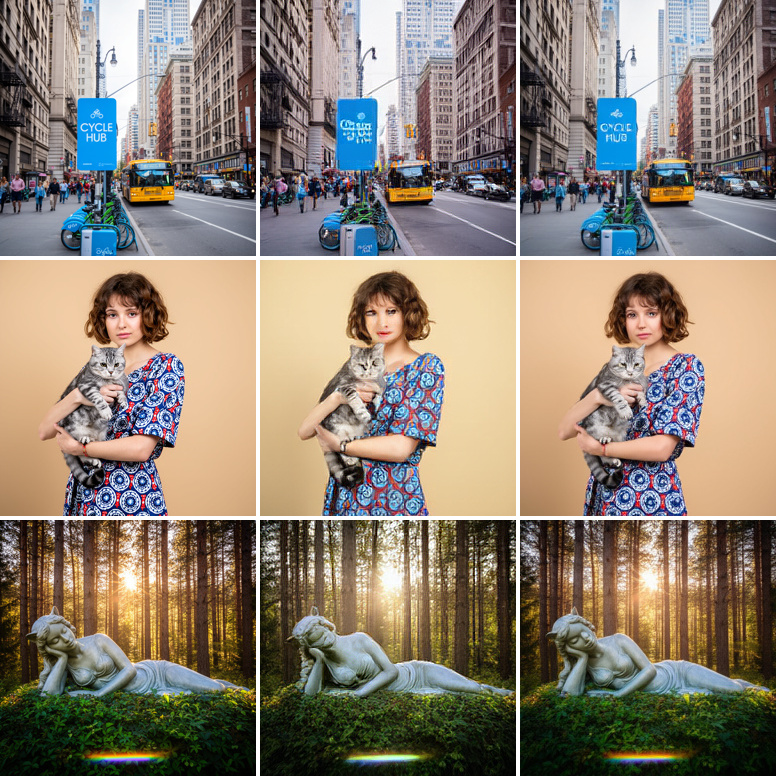}
\caption{\textbf{Reconstruction artifacts
from feature decoders.}
Ground-truth images and reconstructions
using RAE~\cite{zheng2026diffusion}
and RAEv2~\cite{singh2026improved}.
Both decoders alter fine details, including
lettering and facial structures.
On 5,000 held-out images, RAE and RAEv2
achieve 18.2 and 24.2\,dB PSNR, with HF
gaps of 0.132 and 0.096, respectively.}
\label{fig:supp_decoder_detail}
\end{figure}

The RAE-latent model maintains an HF gap near 0.15
throughout training, whereas \method improves steadily,
surpassing the RAE reconstruction reference at 20k
iterations and reaching 0.058 at 30k
(\cref{tab:supp_detail}).
These results show that direct pixel generation can
more closely match real-image high-frequency statistics
without the reconstruction constraints of a fixed
feature decoder.
\Cref{fig:supp_decoder_detail} further illustrates
the reconstruction artifacts introduced by RAE decoders,
particularly in fine details and rendered text.

% =====================================================================
\section{Additional samples}
\label{sec:supp_samples}

\begin{figure*}[p]
  \centering
  \begin{subfigure}{0.90\linewidth}
    \centering
    \includegraphics[width=\linewidth]{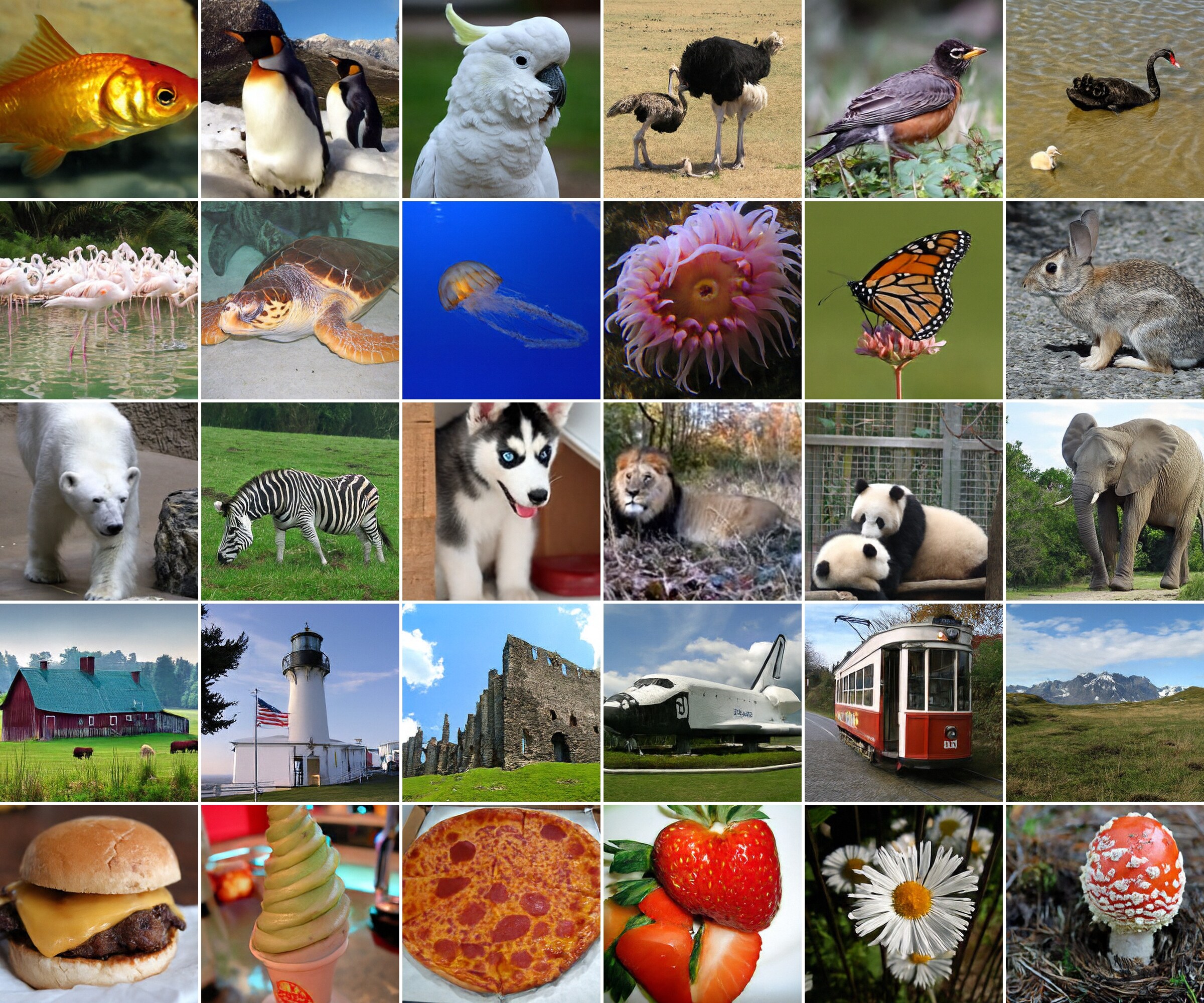}
    \caption{$512{\times}512$ from \method-XL/32 (CFG 2.5 + AutoGuidance 2.0).}
    \label{fig:supp_qual512}
  \end{subfigure}\\[4pt]
  \begin{subfigure}{0.90\linewidth}
    \centering
    \includegraphics[width=\linewidth]{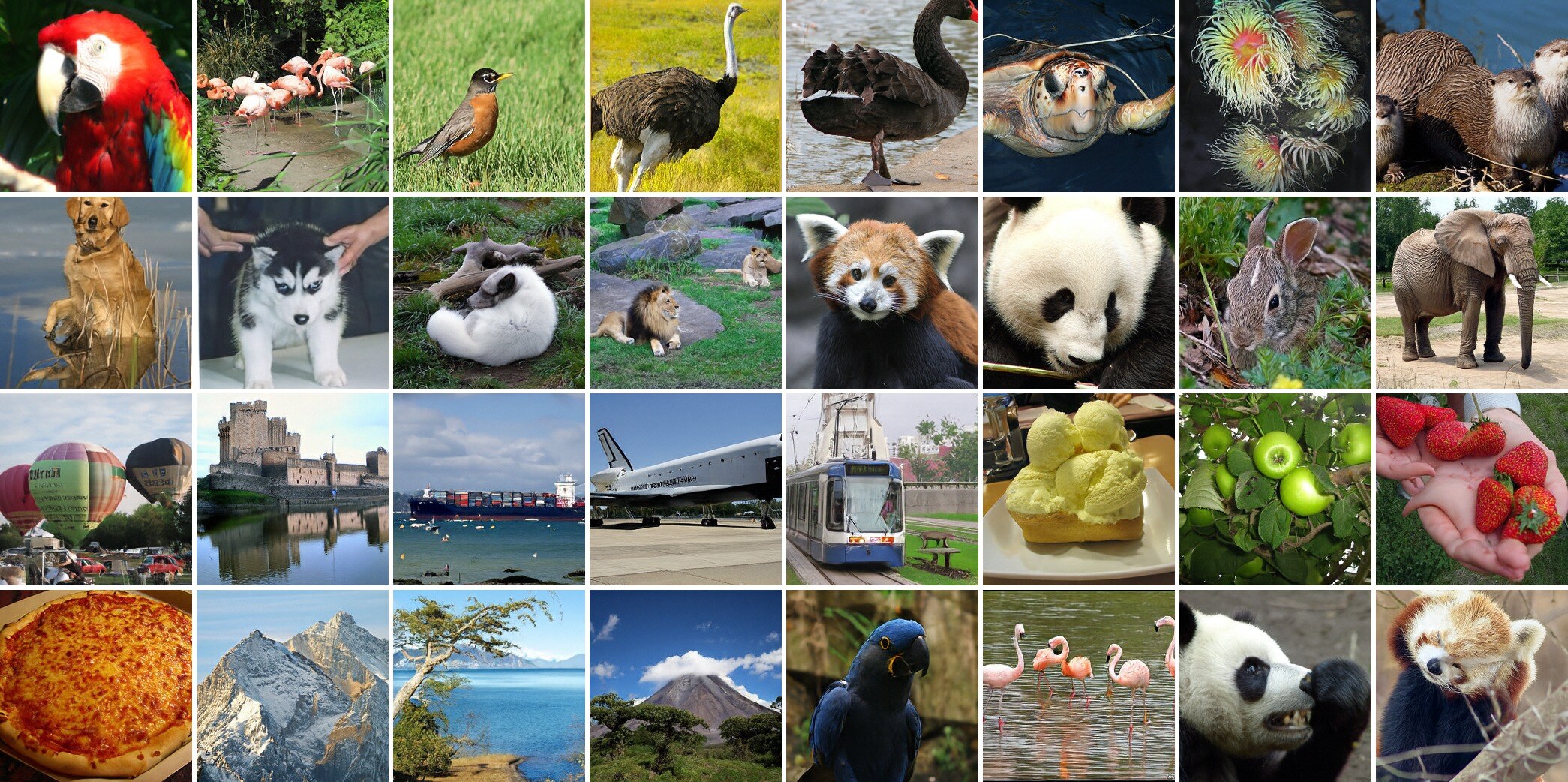}
    \caption{$256{\times}256$ from \method-XL/16 (CFG 1.75 + AutoGuidance 2.0).}
    \label{fig:supp_qual256}
  \end{subfigure}
  \caption{Samples at both resolutions, with the guidance of the reported FID-50k.}
\end{figure*}

\Cref{fig:supp_qual512,fig:supp_qual256} provide larger versions of the samples in \cref{fig:teaser}, covering diverse subjects including animals, faces, textures, and man-made objects. All samples use the same guidance settings as our FID-50k evaluation. At $512{\times}512$, fine details such as fur, feathers, text, and sharp edges illustrate the benefits of generating directly in pixel space without a latent decoder.

\begin{figure*}[p]
  \centering
  \begin{subfigure}{0.75\linewidth}
    \centering
    \includegraphics[width=\linewidth]{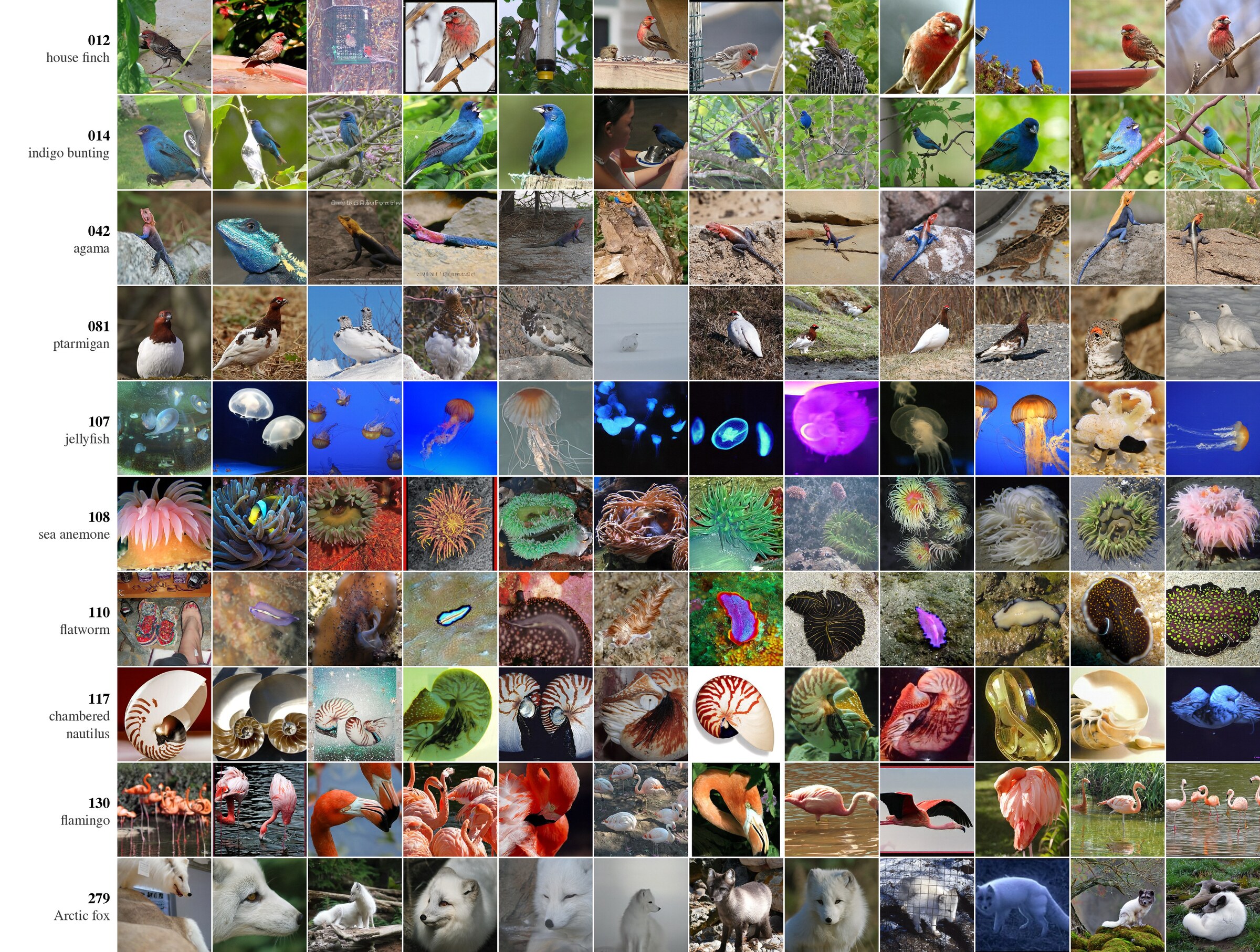}
    \caption{}
    \label{fig:supp_uncur256_1}
  \end{subfigure}\\[4pt]
  \begin{subfigure}{0.75\linewidth}
    \centering
    \includegraphics[width=\linewidth]{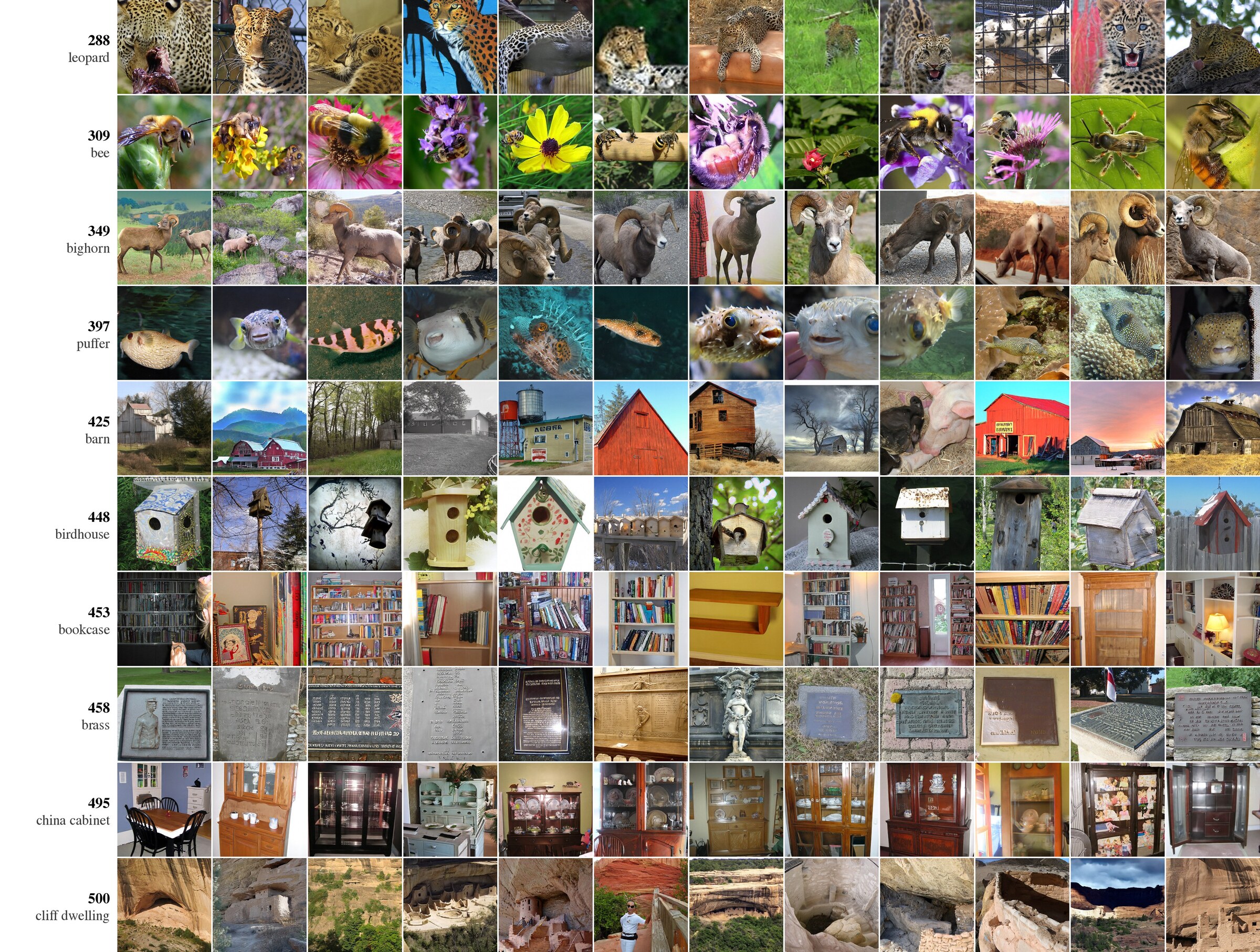}
    \caption{}
    \label{fig:supp_uncur256_2}
  \end{subfigure}
  \caption{\textbf{ImageNet $256{\times}256$ samples} from \method-XL/16, conditioned on the
  specified classes, with the guidance of the reported FID-50k of 1.43 (CFG 1.75 + AutoGuidance 2.0).}
\end{figure*}

\begin{figure*}[p]
  \centering
  \begin{subfigure}{0.75\linewidth}
    \centering
    \includegraphics[width=\linewidth]{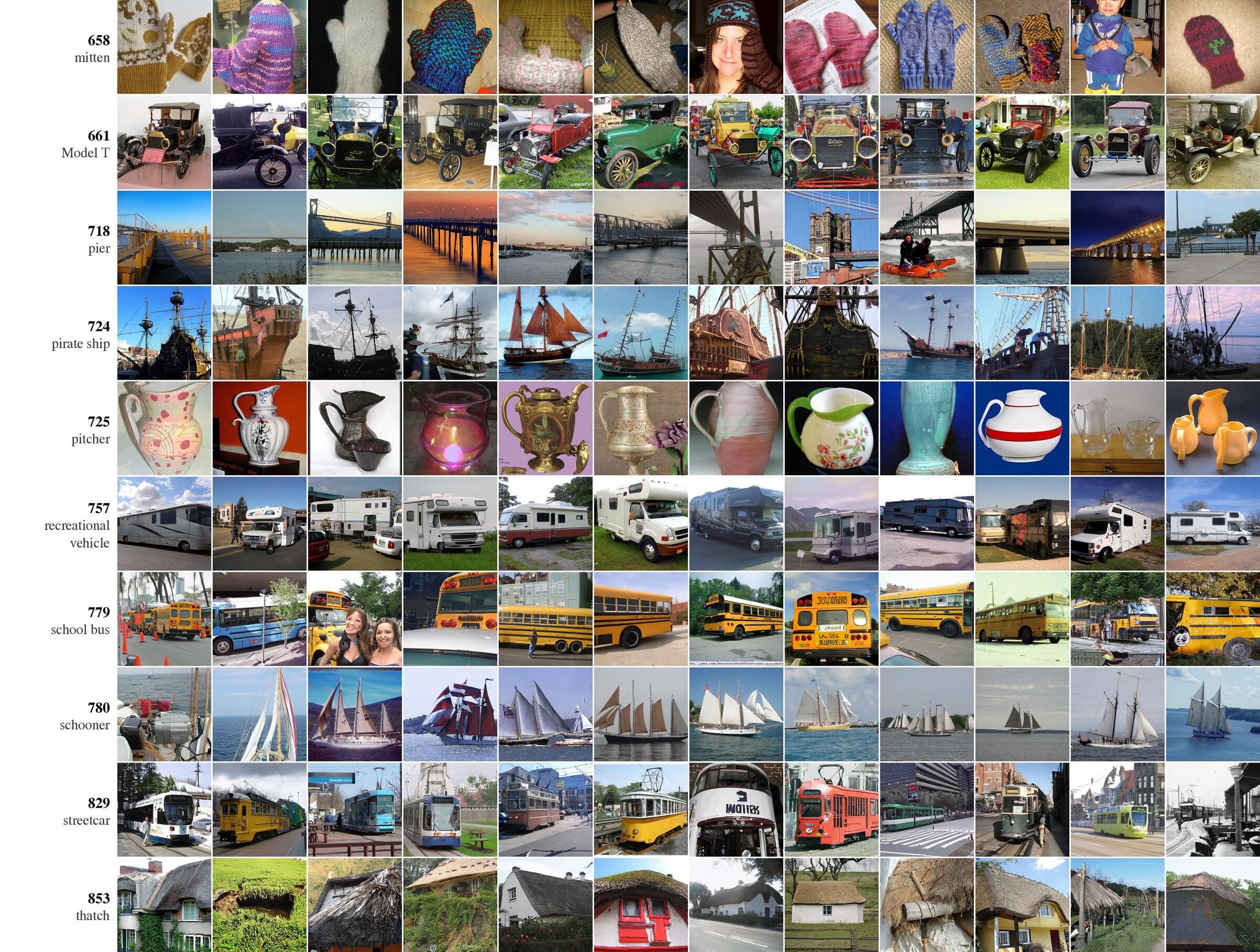}
    \caption{}
    \label{fig:supp_uncur256_3}
  \end{subfigure}\\[4pt]
  \begin{subfigure}{0.75\linewidth}
    \centering
    \includegraphics[width=\linewidth]{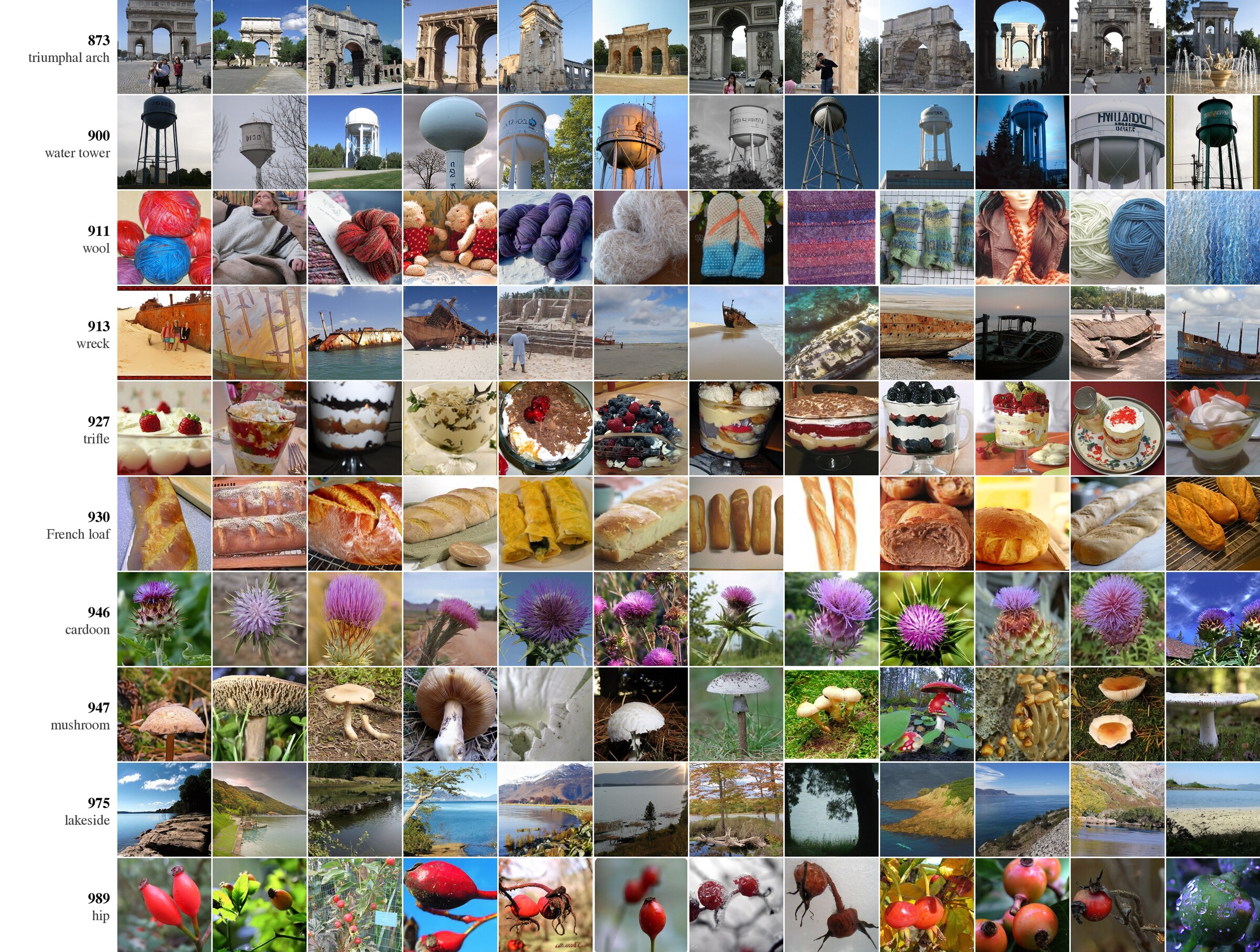}
    \caption{}
    \label{fig:supp_uncur256_4}
  \end{subfigure}
  \caption{\textbf{ImageNet $256{\times}256$ samples} from \method-XL/16, conditioned on the
  specified classes, with the guidance of the reported FID-50k of 1.43 (CFG 1.75 + AutoGuidance 2.0).}
\end{figure*}

\begin{figure*}[p]
  \centering
  \includegraphics[width=\linewidth]{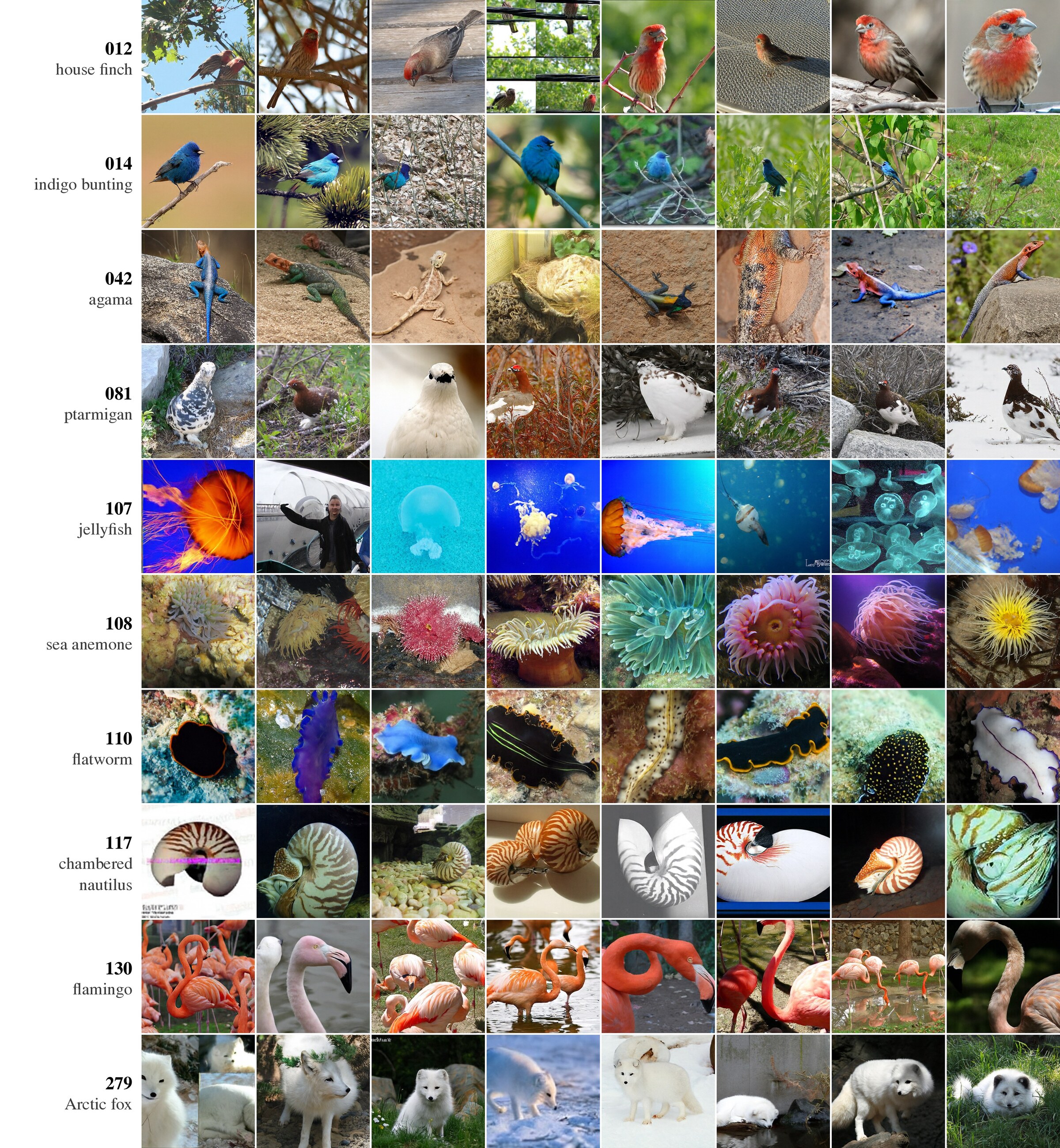}
  \caption{\textbf{ImageNet $512{\times}512$ samples} from \method-XL/32 conditioned on the specified classes, with the guidance of the reported FID-50k of 1.59 (CFG 2.0 + AutoGuidance 2.0).}
  \label{fig:supp_uncur512_1}
\end{figure*}

\begin{figure*}[p]
  \centering
  \includegraphics[width=\linewidth]{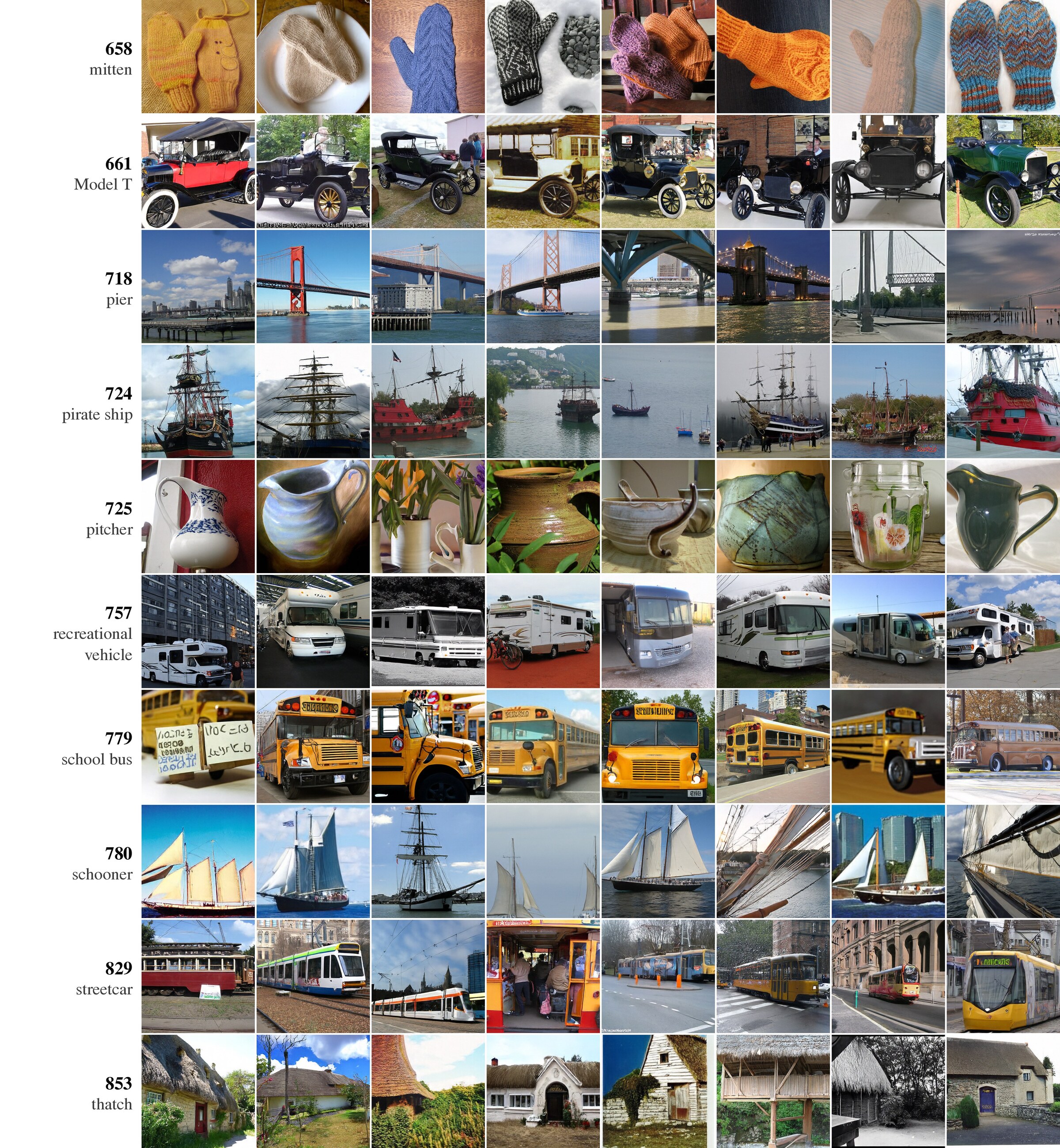}
  \caption{\textbf{ImageNet $512{\times}512$ samples} from \method-XL/32 conditioned on the specified classes (CFG 2.0 + AutoGuidance 2.0, FID-50k 1.59).}
  \label{fig:supp_uncur512_2}
\end{figure*}

\paragraph{Class-conditional gallery.}
\Cref{fig:supp_uncur256_1,fig:supp_uncur256_2,fig:supp_uncur256_3,fig:supp_uncur256_4} show 12 samples of 40 classes at $256{\times}256$, and \cref{fig:supp_uncur512_1,fig:supp_uncur512_2} 8 samples of 20 of them at $512{\times}512$. We use the guidance that gives the reported FID-50k, not a stronger one for visual appeal.

% \TODO{One-step and text-to-image samples.}

% =====================================================================
\section{Limitations}
\label{sec:supp_limits}

\method requires a pretrained vision encoder during
training, and its effectiveness depends on the quality
of the semantic representation.
Although direct pixel generation avoids the reconstruction
limitations of feature decoders, it still struggles with
rendering long or uncommon words and generating
anatomically correct hands.
Targeted fine-tuning or post-training may alleviate
these limitations, which we leave for future work.

% =====================================================================
\section{Text-to-image gallery}
\label{sec:supp_t2i_gallery}

\begin{figure*}[p]
  \centering
  \includegraphics[width=\linewidth]{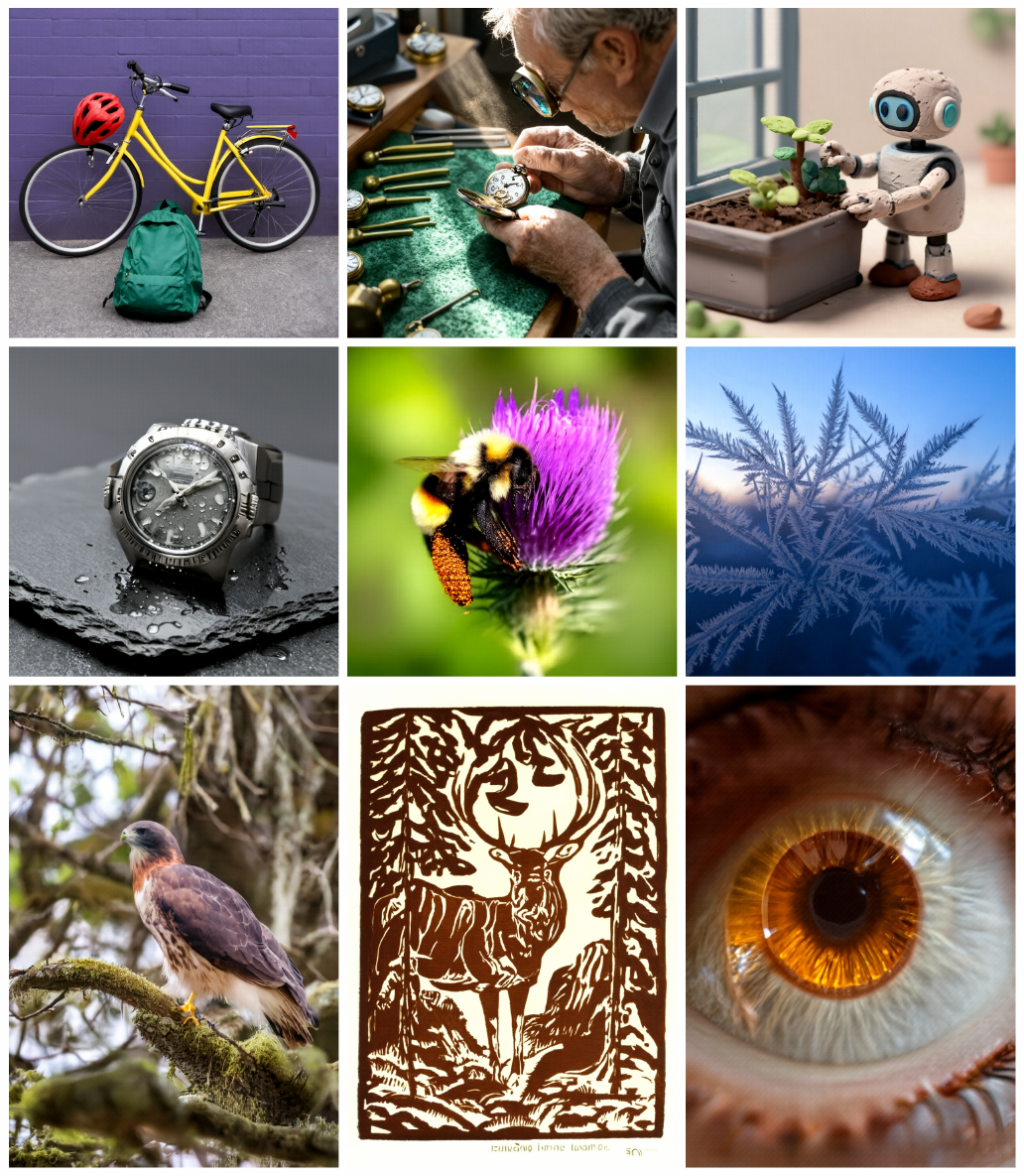}
  \caption{\textbf{Text-to-image gallery} (1 of 4), \method 3.3B at 200k iterations, about
  $1024{\times}1024$ pixels in square and non-square aspect buckets, CFG 6. Pages 1--2 hold the
  photographic, compositional and dense-scene prompts.}
  \label{fig:t2i_gal1}
\end{figure*}

\begin{figure*}[p]
  \centering
  \includegraphics[width=\linewidth]{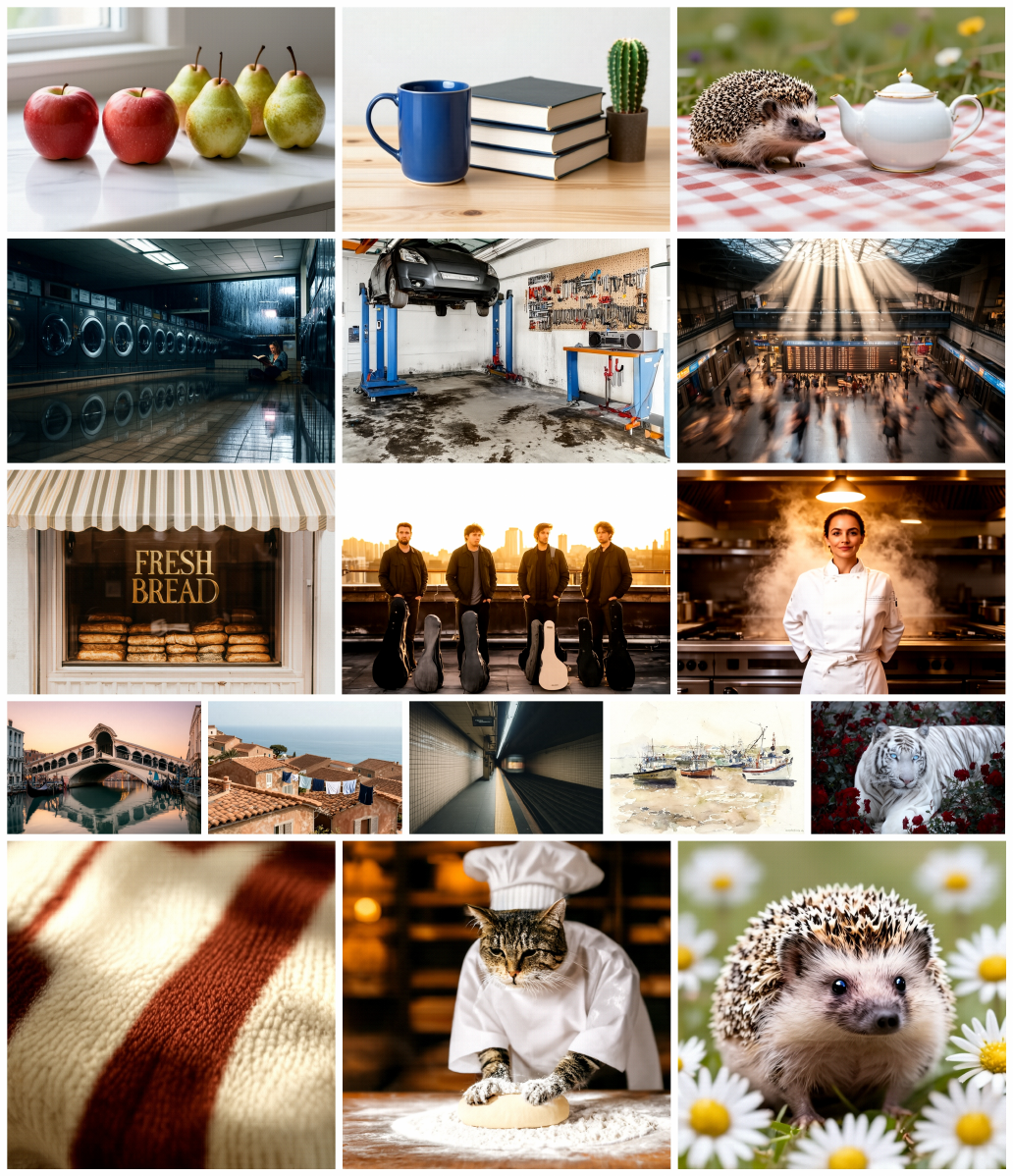}
  \caption{\textbf{Text-to-image gallery} (2 of 4), same model and settings as \cref{fig:t2i_gal1}.}
  \label{fig:t2i_gal2}
\end{figure*}

\begin{figure*}[p]
  \centering
  \includegraphics[width=\linewidth]{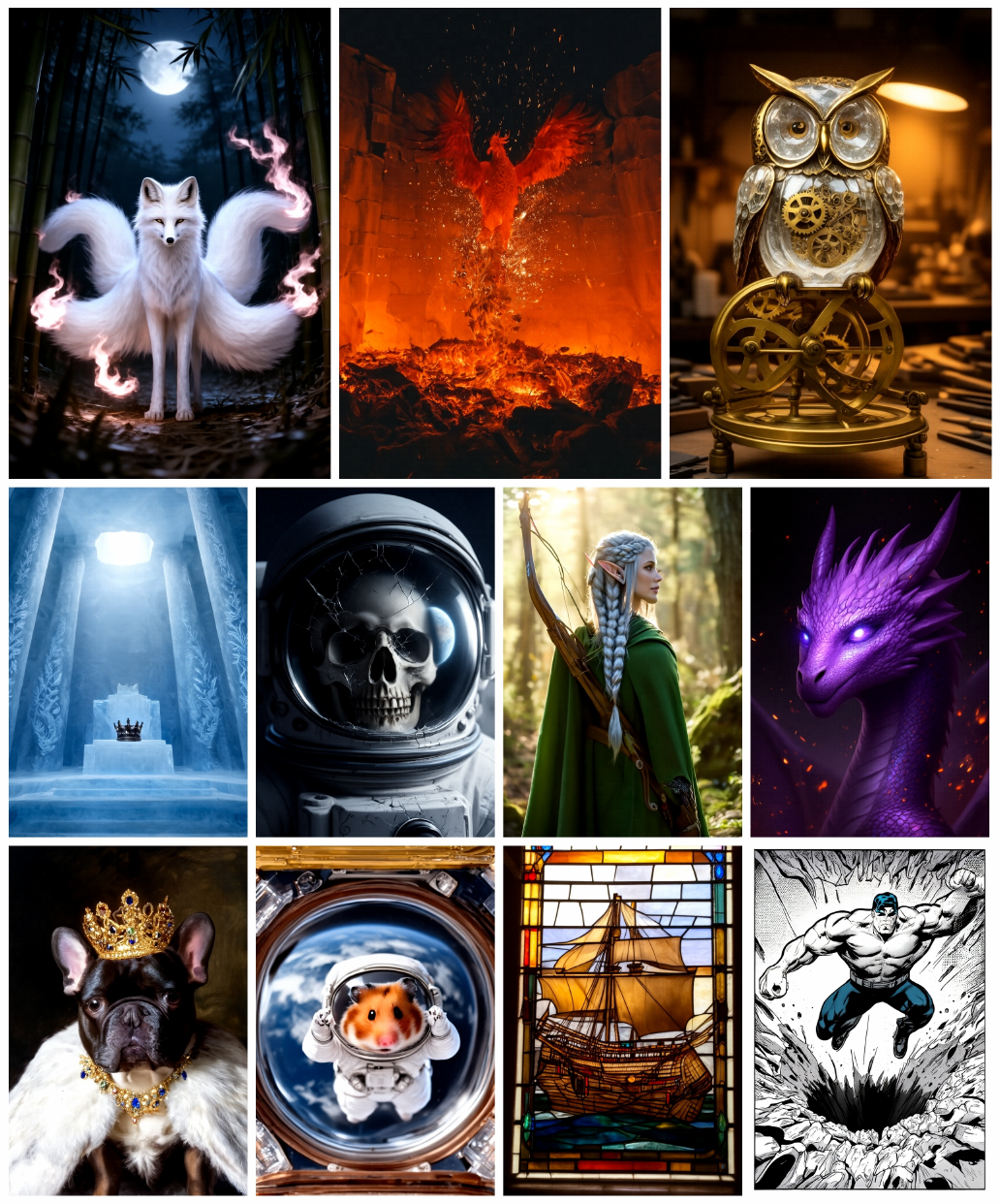}
  \caption{\textbf{Text-to-image gallery} (3 of 4). Pages 3--4 hold the prompts with no photographic
  referent -- invented creatures and places, anthropomorphic and absurd subjects, and illustration
  styles -- at the same settings as \cref{fig:t2i_gal1}.}
  \label{fig:t2i_gal3}
\end{figure*}

\begin{figure*}[p]
  \centering
  \includegraphics[width=\linewidth]{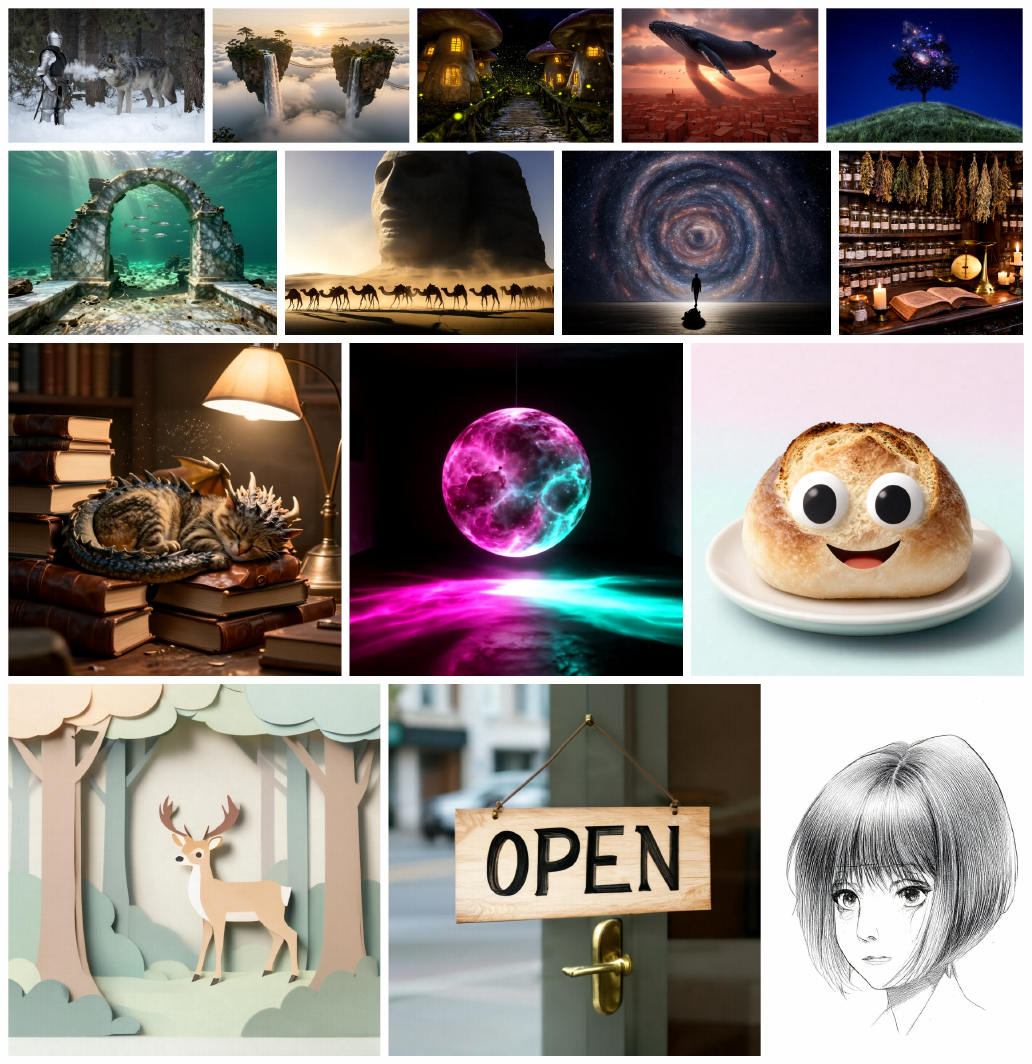}
  \caption{\textbf{Text-to-image gallery} (4 of 4), same model and settings as \cref{fig:t2i_gal1}.}
  \label{fig:t2i_gal4}
\end{figure*}

\Cref{fig:t2i_gal1,fig:t2i_gal2,fig:t2i_gal3,fig:t2i_gal4}
show 52 samples from our 3.3B \method model at 200k
iterations, generated at approximately $1024{\times}1024$
resolution across square and non-square aspect ratios.
The prompts cover diverse subjects and visual styles,
including imaginative scenes, anthropomorphic characters,
illustrations, complex photographic compositions,
landmarks, animals, and rendered text.

% Two limits are worth stating, both measured rather than impressions. Counting is the weakest category, as
% the GenEval sub-scores in \cref{tab:t2i} show: asked for four apples and three pears, none of four
% samples gave both counts. Rendered text depends on the word: ``OPEN'' is correct in all four samples
% here, whereas ``FRESH BREAD'' was correct in three of sixteen samples drawn across CFG 6, 10 and 12, and
% ``SEMANTIC'' in none of twelve. Short, common words are reliable; longer or invented ones are not.

% \subsection{Rendered text, and style range}
% \label{sec:supp_t2i_text}

% Rendered text is where the sampler choice shows most. The gallery above is drawn at the setting the
% scores use -- Heun on 30 steps of the power-2 grid at CFG 6 -- under which only short, common words come
% out reliably: \texttt{OPEN} was correct in all four samples we drew, \texttt{FRESH BREAD} in three of
% sixteen, and \texttt{SEMANTIC} in none of twelve. Drawing the same checkpoint with the sampler the
% ImageNet models use instead, Euler on 100 steps of the power-4 grid, and raising guidance to CFG 18,
% lengthens the strings that survive: the top row of \cref{fig:t2i_text} renders \texttt{DETOUR},
% \texttt{ROCK} and the four-word \texttt{SOUP OF THE DAY}, which the gallery's setting does not reach.
% The limit is still length and familiarity rather than the sampler -- \texttt{GOOD MORNING} loses a
% letter, \texttt{DRIVE ON} transposes one and \texttt{FRESH BREAD} drops several, at both settings -- but
% it sits further out than a figure drawn at CFG 6 suggests. The bottom row shows the stylistic range the
% mix covers, from pastel illustration and watercolour to 3D render and studio portraiture.

% \begin{figure*}[t]
%   \centering
%   \includegraphics[width=\linewidth]{figures/t2i_text_render.pdf}
%   \caption{\textbf{Rendered text and style range}, \method 3.3B at 200k iterations.}
%   \label{fig:t2i_text}
% \end{figure*}

%% file: sec/X_codedim.tex
\section{Vision encoder and code dimension}
\label{sec:supp_codedim}

Our default DINOv2-B code~\cite{oquab2023dinov2}
has 768 channels, matching a $16{\times}16{\times}3$
pixel patch. We examine whether Detour generalizes
to other code dimensions and vision encoders.
All experiments use JiT-B/16 on ImageNet
$256{\times}256$ for 80 epochs, following
\cref{sec:supp_impl}, with unguided EMA2 sampling
(Heun-50, power-2 grid).

\paragraph{Code dimension.}
We compare two approaches to accommodate the
1024-channel features of DINOv2-L:
\begin{enumerate}\itemsep1pt
\item \textbf{Projection:} Reduce the code to
768 channels using PCA followed by a fixed
orthogonal rotation to balance channel variances,
retaining the original architecture.
\item \textbf{State padding:} Preserve all 1024
channels by expanding the diffusion state to four
channels, with the additional clean-image channel
set to zero. The patch embedding dimension
increases accordingly.
\end{enumerate}

\begin{figure}[t]
\centering
\includegraphics[width=\linewidth]{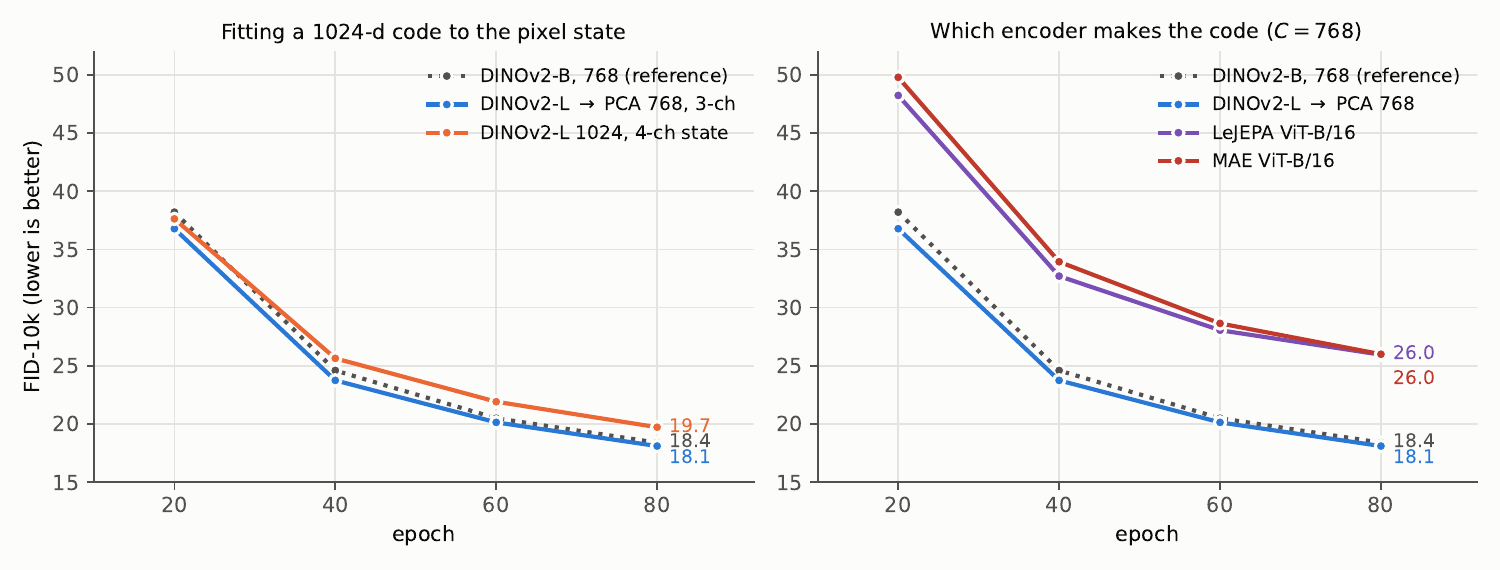}
\caption{\textbf{Code dimension and encoder ablations.}
Unguided FID-10k during training.
\textbf{Left:} Projecting versus padding DINOv2-L
features.
\textbf{Right:} Different 768-channel encoders
with the same architecture.}
\label{fig:supp_codedim}
\end{figure}

Projection retains 94.7\% of the original feature
variance and achieves better FID-50k than padding
(15.01 vs.\ 17.39), without expanding the diffusion
state (\cref{fig:supp_codedim,tab:supp_codedim}).
This demonstrates that Detour can accommodate
wider semantic representations while retaining
the original pixel-space interface.

\paragraph{Vision encoder comparison.}
We compare DINOv2 with MAE~\cite{he2022masked}
and LeJEPA~\cite{balestriero2025lejepa} using the
same 768-channel interface and architecture.
Both alternatives converge more slowly and achieve
worse FID than DINOv2 (\cref{fig:supp_codedim}).

To examine whether this difference is related to
pixel predictability, we fit a linear ridge
regressor to reconstruct image patches from their
corresponding code tokens and report $R^2$.
Although MAE and LeJEPA achieve higher reconstruction
scores than projected DINOv2-L, their generative
performance is worse (\cref{tab:supp_codedim}).
This suggests that the effectiveness of the detour
depends on the semantic structure of the
representation rather than pixel predictability
alone.

\begin{table}[t]
\centering
\caption{\textbf{Code dimension and encoder comparison}
(JiT-B/16, 80 epochs, unguided).
$R^2$ measures linear patch reconstruction accuracy,
estimated from 1.03M tokens.}
\label{tab:supp_codedim}
\small
\setlength{\tabcolsep}{3.5pt}
\begin{adjustbox}{max width=\linewidth}
\begin{tabular}{@{}llcccc@{}}
\toprule
Encoder & Adaptation & $C$ & State
& $R^2$ & FID-50k $\downarrow$ \\
\midrule
DINOv2-B & -- & 768 & 3
    & -- & 16.85$^\dagger$ \\
DINOv2-L & Project & 768 & 3
    & 0.682 & \textbf{15.01} \\
DINOv2-L & Pad & 1024 & 4
    & -- & 17.39 \\
\midrule
MAE & -- & 768 & 3 & 0.916 & 22.77 \\
LeJEPA & -- & 768 & 3 & 0.793 & 23.05 \\
\bottomrule
\end{tabular}
\end{adjustbox}
\end{table}